\documentclass[12pt]{article}
\usepackage[margin=1in]{geometry} 
\usepackage{lipsum} 
\usepackage{graphicx} 
\usepackage{wrapfig} 
\usepackage{tabularx} 
\usepackage{multirow} 
\usepackage{colortbl} 
\usepackage{makecell} 
\usepackage{mdframed} 
\usepackage{times} 
\usepackage{titlesec} 
\usepackage{colortbl} 
\usepackage{caption} 
\usepackage{xcolor} 
\usepackage{framed} 
\usepackage{changepage} 
\usepackage{bigfoot} 
\usepackage{needspace} 
\usepackage[hang, flushmargin]{footmisc} 
\usepackage{titlesec} 
\usepackage{longtable} 
\usepackage{array} 
\usepackage{geometry} 
\usepackage{cellspace} 
\usepackage{float} 
\usepackage{hanging} 
\usepackage{placeins} 
\usepackage{xurl} 
\usepackage[most]{tcolorbox}

\titleformat{\section}
  {\normalfont\Large\bfseries}{\thesection}{1em}{}
\titleformat{\subsection}
  {\normalfont\large\bfseries}{\thesubsection}{1em}{}
\titleformat{\subsubsection}
  {\normalfont\normalsize\bfseries}{\thesubsubsection}{1em}{}

\DeclareNewFootnote[para]{default}
\MakeSortedPerPage{default}

\titlespacing*{\section}{0pt}{\baselineskip}{\baselineskip}
\titlespacing*{\subsection}{0pt}{\baselineskip}{\baselineskip}
\titlespacing*{\subsubsection}{0pt}{\baselineskip}{\baselineskip}

\newtcolorbox{infobox}[1][]{
    colback=white,
    colframe=black,
    fonttitle=\bfseries,
    coltitle=white,
    colbacktitle=black,
    title={#1},
    nofloat,
    breakable,
}

\newcommand{\infopar}{\\[5pt]}
\newcommand{\infoparthree}{\\[4pt]} 

\newcommand{\indicator}[2]{\textbf{#1}: #2}
\definecolor{lightblack}{rgb}{0.7, 0.7, 0.7}
\definecolor{lightgray}{gray}{0.9}

\newcommand{\whitemultirow}[3]{%
  \multirow{#1}{#2}{%
    \cellcolor{white}#3%
  }%
}

\definecolor{lightgray1}{gray}{0.9}
\definecolor{lightgray2}{gray}{0.8}
\definecolor{white}{rgb}{1,1,1}
\definecolor{darkest}{HTML}{0F6291}
\definecolor{medium}{HTML}{A2B8D1}
\definecolor{lightest}{HTML}{C8DDEB}

\newenvironment{abstractpage}
  {
   \begin{center}
   \hrule height 3pt
   \vspace{0.5em}
   \LARGE \textbf{Consciousness in Artificial Intelligence:} \\ \LARGE \textbf{Insights from the Science of Consciousness} 
   \vspace{0.5em}
   \hrule height 1pt
   \vspace{1.5em}

   {
   \begin{tabular*}{\textwidth}{@{\extracolsep{\fill}}ccc}
   \large\textbf{Patrick Butlin\footnotemark[1]} & \large\textbf{Robert Long\footnotemark[1]} & \large\textbf{Eric Elmoznino} \\
   \large\textbf{Yoshua Bengio} & \large\textbf{Jonathan Birch} & \large\textbf{Axel Constant} \\
   \large\textbf{George Deane} & \large\textbf{Stephen M. Fleming} & \large\textbf{Chris Frith} \\
   \large\textbf{Xu Ji} & \large\textbf{Ryota Kanai} & \large\textbf{Colin Klein} \\
   \large\textbf{Grace Lindsay} & \large\textbf{Matthias Michel} & \large\textbf{Liad Mudrik} \\
   \large\textbf{Megan A. K. Peters} & \large\textbf{Eric Schwitzgebel} & \large\textbf{Jonathan Simon} \\
   \multicolumn{3}{c}{\large\textbf{Rufin VanRullen}} \\
   \end{tabular*}
   } 

   \vspace{0.5em} 
   \footnotetext[1]{\small Joint first authors and corresponding authors (patrickbutlin@gmail.com, rgblong@gmail.com)}

   \vspace{0em}
   \large \textbf{Abstract}
   \vspace{0.5em}
   \end{center}
  }
  {\newpage}

\newenvironment{executivesummarypage}
  {
   \noindent\LARGE \textbf{Executive Summary} \normalsize
   \vspace{1em}
   \par\noindent
  }
  {\newpage}

\newcommand{\myindicator}[1]{
    \par\vspace{1ex} 
    \noindent\hangindent=1em\hspace{1em}\textbf{\parbox{\dimexpr\linewidth-2em}{#1}}
    \vspace{1ex} 
    \par 
}

\newenvironment{myquote}
  {\begin{quote}\leftskip=1cm\rightskip=1cm}
  {\end{quote}}

\titleclass{\subsubsubsection}{straight}[\subsubsection]
\newcounter{subsubsubsection}[subsubsection]
\renewcommand\thesubsubsubsection{\thesubsubsection(\alph{subsubsubsection})}
\titleformat{\subsubsubsection}
  {\normalfont\normalsize\itshape}{\thesubsubsubsection\ }{0em}{}
\titlespacing*{\subsubsubsection}{0pt}{3.25ex plus 1ex minus .2ex}{1.5ex plus .2ex}

\makeatletter
\newcommand*\l@subsubsubsection{\@dottedtocline{4}{7em}{4.5em}}
\makeatother

\usepackage[bookmarks=false]{hyperref}
\usepackage{bookmark}
\hypersetup{
  bookmarksdepth=4, 
}

\makeatletter
\def\subsubsubsection{\@startsection{subsubsubsection}{4}{\z@}%
                                     {-3.25ex\@plus -1ex \@minus -.2ex}%
                                     {1.5ex \@plus .2ex}%
                                     {\normalfont\normalsize\itshape}}
\makeatother

\bookmarksetup{
  numbered,
  open,
}

\makeatletter
\def\toclevel@subsubsubsection{4}
\makeatother

\definecolor{blueclr}{HTML}{0000FF}
\definecolor{greenclr}{HTML}{66CC66}
\definecolor{orangecolor}{HTML}{FF9933}
\definecolor{greyclr}{HTML}{999999}

\begin{document}

\thispagestyle{empty}  

\begin{abstractpage}
Whether current or near-term AI systems could be conscious is a topic of scientific interest and increasing public concern. This report argues for, and exemplifies, a rigorous and empirically grounded approach to AI consciousness: assessing existing AI systems in detail, in light of our best-supported neuroscientific theories of consciousness. We survey several prominent scientific theories of consciousness, including recurrent processing theory, global workspace theory, higher-order theories, predictive processing, and attention schema theory. From these theories we derive "indicator properties" of consciousness, elucidated in computational terms that allow us to assess AI systems for these properties. We use these indicator properties to assess several recent AI systems, and we discuss how future systems might implement them. Our analysis suggests that no current AI systems are conscious, but also suggests that there are no obvious technical barriers to building AI systems which satisfy these indicators.\footnote[1]{A previous version of this sentence read "...but also shows that there are no obvious barriers to building conscious AI systems." We have amended it to better reflect the messaging of the report: that satisfying these indicators may be feasible. But satisfying the indicators would not mean that such an AI system would definitely be conscious.}
\end{abstractpage}

\begin{center}
    \textbf{Authors}
\end{center}

\noindent\textbf{Patrick Butlin*}, Future of Humanity Institute, University of Oxford \vfill
\noindent\textbf{Robert Long*}, Center for AI Safety \vfill
\noindent\textbf{Eric Elmoznino}, University of Montreal and MILA - Quebec AI Institute \vfill
\noindent\textbf{Yoshua Bengio}, University of Montreal and MILA - Quebec AI Institute \vfill
\noindent\textbf{Jonathan Birch}, Centre for Philosophy of Natural and Social Science, London School of Economics and Political Science \vfill
\noindent\textbf{Axel Constant}, School of Engineering and Informatics, The University of Sussex and Centre de Recherche en Éthique, University of Montreal \vfill
\noindent\textbf{George Deane}, Department of Philosophy, University of Montreal \vfill
\noindent\textbf{Stephen M. Fleming}, Department of Experimental Psychology and Wellcome Centre for Human Neuroimaging, University College London \vfill
\noindent\textbf{Chris Frith}, Wellcome Centre for Human Neuroimaging, University College London and Institute of Philosophy, University of London \vfill
\noindent\textbf{Xu Ji}, University of Montreal and MILA - Quebec AI Institute \vfill
\noindent\textbf{Ryota Kanai}, Araya, Inc. \vfill
\noindent\textbf{Colin Klein}, School of Philosophy, The Australian National University \vfill
\noindent\textbf{Grace Lindsay}, Department of Psychology and Center for Data Science, New York University \vfill
\noindent\textbf{Matthias Michel}, Center for Mind, Brain and Consciousness, New York University \vfill
\noindent\textbf{Liad Mudrik}, School of Psychological Sciences and Sagol School of Neuroscience, Tel-Aviv University and CIFAR Program in Brain, Mind and Consciousness \vfill
\noindent\textbf{Megan A. K. Peters}, Department of Cognitive Sciences, University of California, Irvine and CIFAR Program in Brain, Mind and Consciousness \vfill
\noindent\textbf{Eric Schwitzgebel}, Department of Philosophy, University of California, Riverside \vfill
\noindent\textbf{Jonathan Simon}, Department of Philosophy, University of Montreal \vfill
\noindent\textbf{Rufin VanRullen}, Centre de Recherche Cerveau et Cognition, CNRS, Université de Toulouse

\newpage

\begin{center}
    \textbf{Details}
\end{center}

\noindent\textbf{Authorship statement:} PB and RL are joint first authors. PB and RL planned and coordinated the project and formulated the core ideas. PB drafted the majority of the report with substantial contributions from RL. EE wrote the first drafts of sections 3.1.2 and 3.1.3, GD wrote the first draft of section 4.1.2, and GL wrote the box on attention. All authors participated in workshops where we planned the report, developed the ideas and reviewed drafts. Authors other than PB, RL and EE are listed in alphabetical order.

\smallskip

\noindent\textbf{Acknowledgements:} The authors would like to thank: Nick Bostrom, who proposed the project to PB and RL and contributed in the early stages; Tim Bayne, Matt Botvinick, David Chalmers, Hakwan Lau, Matt McGill and Murray Shanahan, who attended workshops or took part in other discussions contributing to the preparation of the report; Xander Balwit for her help as a research assistant in the final stages of the project; and Charlie Thompson for his help with graphics and formatting.

\smallskip

\noindent\textbf{Funding:} 
\begin{itemize}
\item RL and PB ran workshops for the report that were supported by Effective Ventures and the EA Long-Term Future Fund.
\item EE was supported by the FRQNT Strategic Clusters Program (Centre UNIQUE) and a Vanier Doctoral Canada Graduate Scholarship.
\item AC was supported by European Research Council grant (XSCAPE) ERC-2020-SyG 951631.
\item YB, AC, GD and JS were supported by a grant from Open Philanthropy.
\item JS was supported by a grant from FRQ and a grant from SSHRC
\item MM was supported by the Templeton World Charity Foundation, as part of the grant ‘Analyzing and Merging Theories of Consciousness’, at the Center for Mind, Brain, and Consciousness (NYU).
\item SMF was supported by a Wellcome/Royal Society Sir Henry Dale Fellowship (206648/Z/17/Z).
\item SMF, LM and MAKP were supported by Fellowships from the CIFAR Program in Brain, Mind and Consciousness.
\item XJ was supported by IVADO.
\item CK was supported by Templeton World Charity Foundation grant TWCF-2020-20539.
\end{itemize}

\smallskip

\noindent{The authors have no conflicts of interest to report.}

\newpage

\begin{executivesummarypage}\noindent The question of whether AI systems could be conscious is increasingly pressing. Progress in AI has been startlingly rapid, and leading researchers are taking inspiration from functions associated with consciousness in human brains in efforts to further enhance AI capabilities. Meanwhile, the rise of AI systems that can convincingly imitate human conversation will likely cause many people to believe that the systems they interact with are conscious.  In this report, we argue that consciousness in AI is best assessed by drawing on neuroscientific theories of consciousness. We describe prominent theories of this kind and investigate their implications for AI.

\begin{adjustwidth}{0.5cm}{0.5cm}
We take our principal contributions in this report to be:

\begin{enumerate}
    \item Showing that the assessment of consciousness in AI is scientifically tractable because consciousness can be studied scientifically and findings from this research are applicable to AI;
    \item Proposing a rubric for assessing consciousness in AI in the form of a list of indicator properties derived from scientific theories;
    \item Providing initial evidence that many of the indicator properties can be implemented in AI systems using current techniques, although no current system appears to be a strong candidate for consciousness.
\end{enumerate}
\end{adjustwidth}

\noindent The rubric we propose is provisional, in that we expect the list of indicator properties we would include to change as research continues. 

Our method for studying consciousness in AI has three main tenets. First, we adopt \textit{computational functionalism}, the thesis that performing computations of the right kind is necessary and sufficient for consciousness, as a working hypothesis. This thesis is a mainstream—although disputed—position in philosophy of mind. We adopt this hypothesis for pragmatic reasons: unlike rival views, it entails that consciousness in AI is possible in principle and that studying the workings of AI systems is relevant to determining whether they are likely to be conscious. This means that it is productive to consider what the implications for AI consciousness would be if computational functionalism were true. Second, we claim that neuroscientific theories of consciousness enjoy meaningful empirical support and can help us to assess consciousness in AI. These theories aim to identify functions that are necessary and sufficient for consciousness in humans, and computational functionalism implies that similar functions would be sufficient for consciousness in AI. Third, we argue that a \textit{theory-heavy approach} is most suitable for investigating consciousness in AI. This involves investigating whether AI systems perform functions similar to those that scientific theories associate with consciousness, then assigning credences based on (a) the similarity of the functions, (b) the strength of the evidence for the theories in question, and (c) one’s credence in computational functionalism. The main alternative to this approach is to use behavioural tests for consciousness, but this method is unreliable because AI systems can be trained to mimic human behaviours while working in very different ways.

Various theories are currently live candidates in the science of consciousness, so we do not endorse any one theory here. Instead, we derive a list of \textit{indicator properties} from a survey of theories of consciousness. Each of these indicator properties is said to be necessary for consciousness by one or more theories, and some subsets are said to be jointly sufficient. Our claim, however, is that AI systems which possess more of the indicator properties are more likely to be conscious. To judge whether an existing or proposed AI system is a serious candidate for consciousness, one should assess whether it has or would have these properties.

The scientific theories we discuss include recurrent processing theory, global workspace theory, computational higher-order theories, and others. We do not consider integrated information theory, because it is not compatible with computational functionalism. We also consider the possibility that agency and embodiment are indicator properties, although these must be understood in terms of the computational features that they imply. This yields the following list of indicator properties:

\begin{table}[h!]
    \centering
    \arrayrulecolor{lightblack}
    \begin{tabularx}{0.9\textwidth}{|>{\raggedright\arraybackslash}X|}
        \hline
        \rowcolor{darkest} \multicolumn{1}{|c|}{\textcolor{white}{\textbf{Recurrent processing theory}}} \\
        \arrayrulecolor{white} \hline
        \rowcolor{medium} \indicator{RPT-1} Input modules using algorithmic recurrence \\
        \arrayrulecolor{white} \hline
        \rowcolor{lightest} \indicator{RPT-2} Input modules generating organised, integrated perceptual representations \\
        \arrayrulecolor{lightblack} \hline
        
        \rowcolor{darkest} \multicolumn{1}{|c|}{\textcolor{white}{\textbf{Global workspace theory}}} \\
        \arrayrulecolor{white} \hline
        \rowcolor{medium} \indicator{GWT-1} Multiple specialised systems capable of operating in parallel (modules) \\
        \arrayrulecolor{white} \hline
        \rowcolor{lightest} \indicator{GWT-2} Limited capacity workspace, entailing a bottleneck in information flow and a selective attention mechanism \\
        \arrayrulecolor{white} \hline
        \rowcolor{medium} \indicator{GWT-3} Global broadcast: availability of information in the workspace to all modules \\
        \arrayrulecolor{white} \hline
        \rowcolor{lightest} \indicator{GWT-4} State-dependent attention, giving rise to the capacity to use the workspace to query modules in succession to perform complex tasks \\
        \arrayrulecolor{lightblack} \hline

        \rowcolor{darkest} \multicolumn{1}{|c|}{\textcolor{white}{\textbf{Computational higher-order theories}}} \\
        \arrayrulecolor{white} \hline
        \rowcolor{medium} \indicator{HOT-1} Generative, top-down or noisy perception modules \\
        \arrayrulecolor{white} \hline
        \rowcolor{lightest} \indicator{HOT-2} Metacognitive monitoring distinguishing reliable perceptual representations from noise \\
        \arrayrulecolor{white} \hline
        \rowcolor{medium} \indicator{HOT-3} Agency guided by a general belief-formation and action selection system, and a strong disposition to update beliefs in accordance with the outputs of metacognitive monitoring \\
        \arrayrulecolor{white} \hline
        \rowcolor{lightest} \indicator{HOT-4} Sparse and smooth coding generating a ``quality space" \\
        \arrayrulecolor{lightblack} \hline

        \rowcolor{darkest} \multicolumn{1}{|c|}{\textcolor{white}{\textbf{Attention schema theory}}} \\
        \arrayrulecolor{white} \hline
        \rowcolor{medium} \indicator{AST-1} A predictive model representing and enabling control over the current state of attention \\
        \arrayrulecolor{lightblack} \hline

        \rowcolor{darkest} \multicolumn{1}{|c|}{\textcolor{white}{\textbf{Predictive processing}}} \\
        \arrayrulecolor{white} \hline
        \rowcolor{medium} \indicator{PP-1} Input modules using predictive coding \\
        \arrayrulecolor{lightblack} \hline

        \rowcolor{darkest} \multicolumn{1}{|c|}{\textcolor{white}{\textbf{Agency and embodiment}}} \\
        \arrayrulecolor{white} \hline
        \rowcolor{medium} \indicator{AE-1} Agency: Learning from feedback and selecting outputs so as to pursue goals, especially where this involves flexible responsiveness to competing goals \\
        \arrayrulecolor{white} \hline
        \rowcolor{lightest} \indicator{AE-2} Embodiment: Modeling output-input contingencies, including some systematic effects, and using this model in perception or control \\
        \arrayrulecolor{lightblack} \hline
    \end{tabularx}
    \caption{Indicator Properties}
\end{table}

We outline the theories on which these properties are based and describe the evidence and arguments that support them in section 2 of the report, as well as explain the formulations used in the table.

Having formulated this list of indicator properties, in section 3.1 we discuss how AI systems could be constructed, or have been constructed, with each of the indicator properties. In most cases, standard machine learning methods could be used to build systems that possess individual properties from this list, although experimentation would be needed to learn how to build and train functional systems which combine multiple properties. There are some properties in the list which are already clearly met by existing AI systems (such as RPT-1, algorithmic recurrence), and others where this is arguably the case (such as the first part of AE-1, agency). Researchers have also experimented with systems designed to implement particular theories of consciousness, including global workspace theory and attention schema theory.

In section 3.2, we consider whether some specific existing AI systems possess the indicator properties. These include Transformer-based large language models and the Perceiver architecture, which we analyse with respect to the global workspace theory. We also analyse DeepMind’s Adaptive Agent, which is a reinforcement learning agent operating in a 3D virtual environment; a system trained to perform tasks by controlling a virtual rodent body; and PaLM-E, which has been described as an ``embodied multimodal language model". We use these three systems as case studies to illustrate the indicator properties concerning agency and embodiment. This work does not suggest that any existing AI system is a strong candidate for consciousness.

This report is far from the final word on these topics. We strongly recommend support for further research on the science of consciousness and its application to AI. We also recommend urgent consideration of the moral and social risks of building conscious AI systems, a topic which we do not address in this report. The evidence we consider suggests that, if computational functionalism is true, conscious AI systems could realistically be built in the near term.

\end{executivesummarypage}

\newpage

\tableofcontents
\newpage

\pdfbookmark[0]{Introduction}{introduction}
\section{Introduction}

In the last decade, striking progress in artificial intelligence (AI) has revived interest in deep and long-standing questions about AI, including the question of whether AI systems could be conscious. This report is about what we take to be the best scientific evidence for and against consciousness in current and near-term AI systems.

Because consciousness is philosophically puzzling, difficult to define and difficult to study empirically, expert opinions about consciousness—in general, and regarding AI systems—are highly divergent. However, we believe that it is possible to make progress on the topic of AI consciousness despite this divergence. There are scientific theories of consciousness that enjoy significant empirical support and are compatible with a range of views about the metaphysics of consciousness. Although these theories are based largely on research on humans, they make claims about properties and functions associated with consciousness that are applicable to AI systems. We claim that using the tools these theories offer us is the best method currently available for assessing whether AI systems are likely to be conscious. In this report, we explain this method in detail, identify the tools offered by leading scientific theories and show how they can be used.

We are publishing this report in part because we take seriously the possibility that conscious AI systems could be built in the relatively near term—within the next few decades. Furthermore, whether or not conscious AI is a realistic prospect in the near term, the rise of large language model-based systems which are capable of imitating human conversation is likely to cause many people to believe that some AI systems are conscious. These prospects raise profound moral and social questions, for society as a whole, for those who interact with AI systems, and for the companies and individuals developing and deploying AI systems. Humanity will be better equipped to navigate these changes if we are better informed about the science of consciousness and its implications for AI. Our aim is to promote understanding of these topics by providing a mainstream, interdisciplinary perspective, which illustrates the degree to which questions about AI consciousness are scientifically tractable, and which may be a basis for future research.

In the remainder of this section, we outline the terminology, methods and assumptions which underlie this report.

\pdfbookmark[1]{Terminology}{terminology}
\subsection{Terminology}

What do we mean by ``conscious" in this report? To say that a person, animal or AI system is conscious is to say either that they are currently having a conscious experience or that they are capable of having conscious experiences. We use ``consciousness" and cognate terms to refer to what is sometimes called ``phenomenal consciousness" (Block 1995). Another synonym for ``consciousness", in our terminology, is ``subjective experience". This report is, therefore, about whether AI systems might be phenomenally conscious, or in other words, whether they might be capable of having conscious or subjective experiences.

What does it mean to say that a person, animal or AI system is having (phenomenally) conscious experiences? One helpful way of putting things is that a system is having a conscious experience when there is ``something it is like'' for the system to be the subject of that experience (Nagel 1974). Beyond this, however, it is difficult to define ``conscious experience" or ``consciousness" by giving a synonymous phrase or expression, so we prefer to use examples to explain how we use these terms. Following Schwitzgebel (2016), we will mention both positive and negative examples—that is, both examples of cognitive processes that are conscious experiences, and examples that are not. By ``consciousness", we mean the phenomenon which most obviously distinguishes between the positive and negative examples.

Many of the clearest positive examples of conscious experience involve our capacities to sense our bodies and the world around us. If you are reading this report on a screen, you are having a conscious visual experience of the screen. We also have conscious auditory experiences, such as hearing birdsong, as well as conscious experiences in other sensory modalities. Bodily sensations which can be conscious include pains and itches. Alongside these experiences of real, current events, we also have conscious experiences of imagery, such as the experience of visualising a loved one’s face.

In addition, we have conscious emotions such as fear and excitement. But there is disagreement about whether emotional experiences are simply bodily experiences, like the feeling of having goosebumps. There is also disagreement about experiences of thought and desire (Bayne \& Montague 2011). It is possible to think consciously about what to watch on TV, but some philosophers claim that the conscious experiences involved are exclusively sensory or imagistic, such as the experience of imagining what it would be like to watch a game show, while others believe that we have ``cognitive" conscious experiences, with a distinctive phenomenology\footnote{The ``phenomenology" or ``phenomenal character" of a conscious experience is what it is like for the subject. In our terminology, all and only conscious experiences have phenomenal characters.} associated specifically with thought.

As for negative examples, there are many processes in the brain, including very sophisticated information-processing that are wholly non-conscious. One example is the regulation of hormone release, which the brain handles without any conscious awareness. Another example is memory storage: you may remember the address of the house where you grew up, but most of the time this has no impact on your consciousness. And, perception in all modalities involves extensive unconscious processing, such as the processing necessary to derive the conscious experience you have when someone speaks to you from the flow of auditory stimulation. Finally, most vision scientists agree that subjects unconsciously process visual stimuli rendered invisible by a variety of psychophysical techniques. For example, in ``masking", a stimulus is briefly flashed on a screen then quickly followed by a second stimulus, called the ``mask" (Breitmeyer \& Ogmen 2006). There is no conscious experience of the first stimulus, but its properties can affect performance on subsequent tasks, such as by ``priming" the subject to identify something more quickly (e.g., Vorberg et al. 2003).

In using the term ``phenomenal consciousness", we mean to distinguish our topic from ``access consciousness", following Block (1995, 2002). Block writes that ``a state is [access conscious] if it is broadcast for free use in reasoning and for direct `rational' control of action (including reporting)" (2002, p. 208). There seems to be a close connection between a mental state’s being conscious, in our sense, and its contents being available to us to report to others or to use in making rational choices. For example, we would expect to be able to report seeing a briefly-presented visual stimulus if we had a conscious experience of seeing it and to be unable to report seeing it if we did not. However, these two properties of mental states are conceptually distinct. How phenomenal consciousness and access consciousness relate to each other is an open question.

Finally, the word ``sentient" is sometimes used synonymously with (phenomenally) ``conscious", but we prefer ``conscious". ``Sentient" is sometimes used to mean having senses, such as vision or olfaction. However, being conscious is not the same as having senses. It is possible for a system to sense its body or environment without having any conscious experiences, and it may be possible for a system to be conscious without sensing its body or environment. ``Sentient" is also sometimes used to mean capable of having conscious experiences such as pleasure or pain, which feel good or bad, and we do not want to imply that conscious systems must have these capacities. A system could be conscious in our sense even if it only had ``neutral" conscious experiences. Pleasure and pain are important but they are not our focus here.\footnote{For the sake of further illustration, here are some other definitions of phenomenal consciousness: 

Chalmers (1996): ``When we think and perceive, there is a whir of information-processing, but there is also a subjective aspect. As Nagel (1974) has put it, there is \textit{something it is like} to be a conscious organism. This subjective aspect is experience. When we see, for example, we \textit{experience} visual sensations: the felt quality of redness, the experience of dark and light, the quality of depth in a visual field. Other experiences go along with perception in different modalities: the sound of a clarinet, the smell of mothballs. Then there are bodily sensations, from pains to orgasms; mental images that are conjured up internally; the felt quality of emotion, and the experience of a stream of conscious thought. What unites all of these states is that there is something it is like to be in them."

Graziano (2017): ``You can connect a computer to a camera and program it to process visual information—color, shape, size, and so on. The human brain does the same, but in addition, we report a subjective experience of those visual properties. This subjective experience is not always present. A great deal of visual information enters the eyes, is processed by the brain and even influences our behavior through priming effects, without ever arriving in awareness. Flash something green in the corner of vision and ask people to name the first color that comes to mind, and they may be more likely to say `green' without even knowing why. But some proportion of the time we also claim, `I have a subjective visual experience. I \textit{see} that thing with my conscious mind. Seeing \textit{feels} like something.'"
}

\pdfbookmark[1]{Methods and Assumptions}{methods-assumptions}
\subsection{Methods and Assumptions}
Our method for investigating whether current or near-future AI systems might be conscious is based on three assumptions. These are:

\begin{adjustwidth}{0.5cm}{0.5cm}
\begin{enumerate}
    \item \textit{Computational functionalism:} Implementing computations of a certain kind is necessary and sufficient for consciousness, so it is possible in principle for non-organic artificial systems to be conscious.
    \item \textit{Scientific theories:} Neuroscientific research has made progress in characterising functions that are associated with, and may be necessary or sufficient for, consciousness; these are described by scientific theories of consciousness.
    \item \textit{Theory-heavy approach:} A particularly promising method for investigating whether AI systems are likely to be conscious is assessing whether they meet functional or architectural conditions drawn from scientific theories, as opposed to looking for theory-neutral behavioural signatures.
\end{enumerate}
\end{adjustwidth}

These ideas inform our investigation in different ways. We adopt computational functionalism as a working hypothesis because this assumption makes it relatively straightforward to draw inferences from neuroscientific theories of consciousness to claims about AI. Some researchers in this area reject computational functionalism (e.g. Searle 1980, Tononi \& Koch 2015) but our view is that it is worth exploring its implications. We accept the relevance and value of some scientific theories of consciousness because they describe functions that could be implemented in AI and we judge that they are supported by good experimental evidence. And, our view is that, although this may not be so in other cases, a theory-heavy approach is necessary for AI. A theory-heavy approach is one that focuses on how systems work, rather than on whether they display forms of outward behaviour that might be taken to be characteristic of conscious beings (Birch 2022b). We explain these three ideas in more detail in this section.

Two further points about our methods and assumptions are worth noting before we go on. The first is that, for convenience, we will generally write as though whether a system is conscious is an all-or-nothing matter, and there is always a determinate fact about this (although in many cases this fact may be difficult to learn). However, we are open to the possibility that this may not be the case: that it may be possible for a system to be partly conscious, conscious to some degree, or neither determinately conscious nor determinately non-conscious (see Box 1). 

\begin{infobox} [Box 1: Determinacy, degrees, dimensions]
In this report, we generally write as though consciousness is an all-or-nothing matter: a system either is conscious, or it isn’t. However, there are various other possibilities. There seem to be many properties that have ``blurry'' boundaries, in the sense that whether some object has that property may be indeterminate. For example, a shirt may be a colour somewhere on the borderline between yellow and green, such that there is no fact of the matter about whether it is yellow or not. In principle, consciousness could be like this: there could be creatures that are neither determinately conscious nor determinately non-conscious (Simon 2017, Schwitzgebel forthcoming). If this is the case, some AI systems could be in this ``blurry'' zone. This kind of indeterminacy arguably follows from materialism about consciousness (Birch 2022a). 
\infopar
Another possibility is that there could be degrees of consciousness so that it is possible for one system to be more conscious than another (Lee 2022). In this case, it might be possible to build AI systems that are conscious but only to a very slight degree, or even systems which are conscious to a much greater degree than humans (Shulman \& Bostrom 2021). Alternatively, rather than a single scale, it could be that consciousness varies along multiple dimensions (Birch et al. 2020).
\infopar
Lastly, it could be that there are multiple elements of consciousness. These would not be necessary conditions for some further property of consciousness, but rather constituents which make up consciousness. These elements may be typically found together in humans, but separable in other animals or AI systems. In this case, it would be possible for a system to be partly conscious, in the sense of having some of these elements.

\end{infobox}

The second is that we recommend thinking about consciousness in AI in terms of confidence or credence. Uncertainty about this topic is currently unavoidable, but there can, nonetheless, be good reasons to think that one system is much more likely than another to be conscious, and this can be relevant to how we should act. So it is useful to think about one’s credence in claims in this area. For instance, one might think it justified to have a credence of about 0.5 in the conjunction of a set of theoretical claims which imply that a given AI system is conscious; if so, one should have a similar credence that the system is conscious.

\pdfbookmark[2]{Computational functionalism}{comp-functionalism}
\subsubsection{Computational functionalism}

Computational functionalism about consciousness is a claim about the kinds of properties of systems with which consciousness is correlated. According to \textit{functionalism} about consciousness, it is necessary and sufficient for a system to be conscious that it has a certain functional organisation: that is, that it can enter a certain range of states, which stand in certain causal relations to each other and to the environment. Computational functionalism is a version of functionalism that further claims that the relevant functional organisation is computational.\footnote{Computational functionalism is compatible with a range of views about the relationship between consciousness and the physical states which implement computations. In particular, it is compatible with both (i) the view that there is nothing more to a state’s being conscious than its playing a certain role in implementing a computation; and (ii) the view that a state’s being conscious is a matter of its having \textit{sui generis} phenomenal properties, for which its role in implementing a computation is sufficient.}

Systems that perform computations process information by implementing algorithms; computational functionalism claims that it is sufficient for a state to be conscious that it plays a role of the right kind in the implementation of the right kind of algorithm. For a system to implement a particular algorithm is for it to have a set of features at a certain level of abstraction: specifically, a range of possible information-carrying states, and particular dispositions to make transitions between these states. The algorithm implemented by a system is an abstract specification of the transitions between states, including inputs and outputs, which it is disposed to make. For example, a pocket calculator implements a particular algorithm for arithmetic because it generates transitions from key-presses to results on screen by going through particular sequences of internal states.

An important upshot of computational functionalism, then, is that whether a system is conscious or not depends on features that are more abstract than the lowest-level details of its physical make-up. The material substrate of a system does not matter for consciousness except insofar as the substrate affects which algorithms the system can implement. This means that consciousness is, in principle, multiply realisable: it can exist in multiple substrates, not just in biological brains. That said, computational functionalism does not entail that  \textit{any} substrate can be used to construct a conscious system (Block 1996). As Michel and Lau (2021) put it, ``Swiss cheese cannot implement the relevant computations.'' We tentatively assume that computers as we know them are in principle capable of implementing algorithms sufficient for consciousness, but we do not claim that this is certain.

It is also important to note that systems that compute the same mathematical function may do so by implementing different algorithms, so computational functionalism does not imply that systems that ``do the same thing'' in the sense that they compute the same input-output function are necessarily alike in consciousness (Sprevak 2007). Furthermore, it is consistent with computational functionalism that consciousness may depend on performing operations on states with specific representational formats, such as analogue representation (Block 2023). In terms of Marr’s (1982) levels of analysis, the idea is that consciousness depends on what is going on in a system at the algorithmic and representational level, as opposed to the implementation level, or the more abstract ``computational'' (input-output) level.

We adopt computational functionalism as a working hypothesis primarily for pragmatic reasons. The majority of leading scientific theories of consciousness can be interpreted computationally—that is, as making claims about computational features which are necessary or sufficient for consciousness in humans. If computational functionalism is true, and if these theories are correct, these features would also be necessary or sufficient for consciousness in AI systems. Non-computational differences between humans and AI systems would not matter. The assumption of computational functionalism, therefore, allows us to draw inferences from computational scientific theories to claims about the likely conditions for consciousness in AI. On the other hand, if computational functionalism is false, there is no guarantee that computational features which are correlated with consciousness in humans will be good indicators of consciousness in AI. It could be, for instance, that some non-computational feature of living organisms is necessary for consciousness (Searle 1980, Seth 2021), in which case consciousness would be impossible in non-organic artificial systems.

Having said that, it would not be worthwhile to investigate artificial consciousness on the assumption of computational functionalism if this thesis were not sufficiently plausible. Although we have different levels of confidence in computational functionalism, we agree that it is plausible.\footnote{One influential argument is by Chalmers (1995): if a person’s neurons were gradually replaced by functionally-equivalent artificial prostheses, their behaviour would stay the same, so it is implausible that they would undergo any radical change in conscious experience (if they did, they would act as though they hadn’t noticed).} These different levels of confidence feed into our personal assessments of the likelihood that particular AI systems are conscious, and of the likelihood that conscious AI is possible at all.

\pdfbookmark[2]{Scientific theories of consciousness}{sci-theories}
\subsubsection{Scientific theories of consciousness}

The second idea which informs our approach is that some scientific theories of consciousness are well-supported by empirical evidence and make claims which can help us assess AI systems for consciousness. These theories have been developed, tested and refined through decades of high-quality neuroscientific research (for recent reviews, see Seth \& Bayne 2022, Yaron et al. 2022). Positing that computational functions are sufficient for consciousness would not get us far if we had no idea which functions matter; but these theories give us valuable indications.

Scientific theories of consciousness are different from metaphysical theories of consciousness. Metaphysical theories of consciousness make claims about how consciousness relates to the material world in the most general sense. Positions in the metaphysics of consciousness include property dualism (Chalmers 1996, 2002), panpsychism (Strawson 2006, Goff 2017), materialism (Tye 1995, Papineau 2002) and illusionism (Frankish 2016). For example, materialism claims that phenomenal properties are physical properties while property dualism denies this. In contrast, scientific theories of consciousness make claims about which specific material phenomena—usually brain processes—are associated with consciousness. Some explicitly aim to identify the neural correlates of conscious states (NCCs), defined as the minimal sets of neural events which are jointly sufficient for those states (Crick \& Koch 1990, Chalmers 2000). The central question for scientific theories of consciousness is what distinguishes cases in which conscious experience arises from those in which it does not, and while this is not the only question such theories might address, it is the focus of this report.

We discuss several specific scientific theories in detail in section 2. Here, we provide a brief overview of the methods of consciousness science to show that consciousness can be studied scientifically.

The scientific study of consciousness relies on assumptions about links between consciousness and behaviour (Irvine 2013). For instance, in a study on vision, experimenters might manipulate a visual stimulus (e.g. a red triangle) in a certain way—say, by flashing it at two different speeds. If they find that subjects report seeing the stimulus in one condition but not in the other, they might argue that subjects have a conscious visual experience of the stimulus in one condition but not in the other. They could then measure differences in brain activity between the two conditions, and draw inferences about the relationships between brain activity and consciousness—a method called ``contrastive analysis'' (Baars 1988). This method relies on the assumption that subjects’ reports are a good guide to their conscious experiences.

As a method for studying consciousness in humans and other animals, relying on subjects’ reports has two main problems. The first problem is uncertainty about the relationship between conscious experience, reports and cognitive processes which may be involved in making reports, such as attention and memory. Inasmuch as reports or reportability require more processing than conscious experience, studies that rely on reports may be misleading: brain processes which are involved in processing the stimulus and making reports, but are not necessary for consciousness, could be misidentified as among the neural correlates of consciousness (Aru et al. 2012). Another possibility is that phenomenal consciousness may have relatively rich contents, of which only a proportion are selected by attention for further processing yielding cognitive access, which is, in turn, necessary for report. In this case, relying on reports may lead us to misidentify the neural basis of access as that of phenomenal consciousness (Block 1995, 2007). The methodological problem here is arguably more severe because it is an open question whether phenomenal consciousness ``overflows'' cognitive access in this way—researchers have conflicting views (Phillips 2018a).

A partial solution to this problem may be the use of ``no-report paradigms'', in which indicators of consciousness other than reports are used, having been calibrated for correlation with consciousness in separate experiments, which do use reports (Tsuchiya et al. 2015). The advantage of this paradigm is that subjects are not required to make reports in the main experiments, which may mitigate the problem of report confounds. No-report paradigms are not a ``magic bullet'' for this problem (Block 2019, Michel \& Morales 2020), but they may be an important step in addressing it.

Another possible method for measuring consciousness is the use of metacognitive judgments such as confidence ratings (e.g. Peters \& Lau 2015). For example, subjects might be asked how confident they are in an answer about a stimulus, e.g. about whether a briefly-presented stimulus was oriented vertically or horizontally. The underlying thought here is that subjects’ ability to track the accuracy of their responses using confidence ratings (known as metacognitive sensitivity) depends on their being conscious of the relevant stimuli. Again, this method is imperfect, but it has some advantages over asking subjects to report their conscious experiences (Morales \& Lau 2021; Michel 2022). There are various potential confounds in consciousness science, but researchers can combine evidence from studies of different kinds to reduce the force of methodological objections (Lau 2022).

The second problem with the report approach is that there are presumably some subjects of conscious experience who cannot make reports, including non-human animals, infants and people with certain kinds of cognitive disability. This problem is perhaps most pressing in the case of non-human animals, because if we knew more about consciousness in animals—especially those which are relatively unlike us—we might have a far better picture of the range of brain processes that are correlated with consciousness. This difficult problem has recently received increased attention (e.g. Birch 2022b). However, although current scientific theories of consciousness are primarily based on data from healthy adult humans, it can still be highly instructive to examine whether AI systems use processes similar to those described by these theories.

\begin{infobox} [Box 2: Metaphysical theories and the science of consciousness]

Major positions in the metaphysics of consciousness include materialism, property dualism, panpsychism and illusionism (for a detailed and influential overview, see Chalmers 2002). 
\infopar
\textbf{\textit{Materialism}} claims that consciousness is a wholly physical phenomenon. Conscious experiences are states of the physical world—typically brain states—and the properties that make up the phenomenal character of our experiences, known as phenomenal properties, are physical properties of these states. For example, a materialist might claim that the experience of seeing a red tulip is a particular brain state and that the ``redness'' of the experience is a feature of that state.
\infopar
\textbf{\textit{Property dualism}} denies materialism, claiming that phenomenal properties are non-physical properties. Unlike \textit{substance} dualism, this view claims that there is just one sort of substance or entity while asserting that it has both physical and phenomenal properties. The ``redness'' of the experience of seeing the tulip may be a property of the brain state involved, but it is distinct from any physical property of this state. \infopar
\textbf{\textit{Panpsychism}} claims that phenomenal properties, or simpler but related ``proto-phenomenal'' properties, are present in all fundamental physical entities. A panpsychist might claim that an electron, as a fundamental particle, has either a property like the ``redness'' of the tulip experience or a special precursor of this property. Panpsychists do not generally claim that everything has conscious experiences—instead, the phenomenal aspects of fundamental entities only combine to give rise to conscious experiences in a few macro-scale entities, such as humans. 
\infopar
\textbf{\textit{Illusionism}} claims that we are subject to an illusion in our thinking about consciousness and that either consciousness does not exist (\textit{strong} illusionism), or we are pervasively mistaken about some of its features (\textit{weak} illusionism). However, even strong illusionism acknowledges the existence of ``quasi-phenomenal'' properties, which are properties that are misrepresented by introspection as phenomenal. For example, an illusionist might say that when one seems to have the conscious experience of seeing a red tulip, some brain state is misrepresented by introspection as having a property of phenomenal ``redness''. 
\infopar
Importantly, there is work for the science of consciousness to do on all four of these metaphysical positions. If \textbf{materialism} is true, then some brain states are conscious experiences and others are not, and the role of neuroscience is to find out what distinguishes them. Similarly, \textbf{property dualism} and \textbf{panpsychism} both claim that some brain states but not others are associated with conscious experiences, and are compatible with the claim that this difference can be investigated scientifically. According to \textbf{illusionism}, neuroscience can explain why the illusion of consciousness arises, and in particular why it arises in connection with some brain states but not others.

\end{infobox}

\pdfbookmark[2]{Theory-heavy approach}{theory-heavy}
\subsubsection{Theory-heavy approach}

In section 1.2.1 we adopted computational functionalism, the thesis that implementing certain computational processes is necessary and sufficient for consciousness, as a working hypothesis, and in section 1.2.2 we noted that there are scientific theories that aim to describe correlations between computational processes and consciousness. Combining these two points yields a promising method for investigating consciousness in AI systems: we can observe whether they use computational processes which are similar to those described in scientific theories of consciousness, and adjust our assessment accordingly. To a first approximation, our confidence that a given system is conscious can be determined by (a) the similarity of its computational processes to those posited by a given scientific theory of consciousness, (b) our confidence in this theory, (c) and our confidence in computational functionalism.\footnote{The theory may entail computational functionalism, in which case (c) would be unnecessary. But we find it helpful to emphasise that if computational functionalism is a background assumption in one’s construal of a theory, one should take into account both uncertainty about this assumption, and uncertainty about the specifics of the theory.} Considering multiple theories can then give a fuller picture. This method represents a ``theory-heavy'' approach to investigating consciousness in AI.

The term ``theory-heavy'' comes from Birch (2022b), who considers how we can  scientifically investigate consciousness in non-human animals, specifically invertebrates.

Birch argues against using the theory-heavy approach in this case. One of Birch’s objections is that the evidence from humans that supports scientific theories does not tell us how much their conditions can be relaxed while still being sufficient for consciousness (see also Carruthers 2019). That is, while we might have good evidence that some process is sufficient for consciousness in humans, this evidence will not tell us whether a process in another animal, which is similar in some respects but not others, is also sufficient for consciousness. To establish this we would need antecedent evidence about which non-human animals or systems are conscious—unfortunately, the very question we are uncertain about.

Another way of thinking about this problem is in terms of how we should \textit{interpret} theories of consciousness. As we will see throughout this report, it is possible to interpret theories either in relatively restrictive ways, as claiming only that very specific features found in humans are sufficient for consciousness, or as giving much more liberal, abstract conditions, which may be met by surprisingly simple artificial systems (Shevlin 2021). Moderate interpretations which strike a balance between appealing generality (consciousness is not just \textit{this} very specific process in the human brain) and unintuitive liberality (consciousness is not a property satisfied by extremely simple systems) are attractive, but it is not clear that these have empirical support over the alternatives.

While this objection does point to an important limitation of theory-heavy approaches, it does not show that a theory-heavy approach cannot give us useful information about consciousness in AI. Some AI systems will use processes that are much more similar to those identified by theories of consciousness than others, and this objection does not count against the claim that those using more similar processes are correspondingly better candidates for consciousness. Drawing on theories of consciousness is necessary for our investigation because they are the best available guide to the features we should look for. Investigating animal consciousness is different because we already have reasons to believe that animals that are more closely related to humans and display more complex behaviours are better candidates for consciousness. Similarities in cognitive architecture can be expected to be substantially correlated with phylogenetic relatedness, so while it will be somewhat informative to look for these similarities, this will be less informative than in the case of AI. \footnote{Birch (2022b) advocates a ``theory-light'' approach, which has two aspects: (1) rejecting the idea that we should assess consciousness in non-human animals by looking for processes that particular theories associate with consciousness; and (2) not committing to any particular theory now, but aiming to develop better theories in the future when we have more evidence about animal (and perhaps AI) consciousness. Our approach is ``theory-heavy'' in the sense that, in contrast with the first aspect of the theory-light approach, we do assess AI systems by looking for processes that scientific theories associate with consciousness. However, like Birch, we do not commit to any one theory at this time. More generally, Birch’s approach makes recommendations about how the science of consciousness should be developed, whereas we are only concerned with what kind of evidence should be used to make assessments of consciousness in AI systems now, given our current knowledge.}

The main alternative to the theory-heavy approach for AI is to use behavioural tests that purport to be neutral between scientific theories. Behavioural tests have been proposed specifically for consciousness in AI (Elamrani \& Yampolskiy 2019). One interesting example is Schneider’s (2019) Artificial Consciousness Test, which requires the AI system to show a ready grasp of consciousness-related concepts and ideas in conversation, perhaps exhibiting ``problem intuitions'' like the judgement that spectrum inversion is possible (Chalmers 2018). The Turing test has also been proposed as a test for consciousness (Harnad 2003).

In general, we are sceptical about whether behavioural approaches to consciousness in AI can avoid the problem that AI systems may be trained to mimic human behaviour while working in very different ways, thus ``gaming'' behavioural tests (Andrews \& Birch 2023). Large language model-based conversational agents, such as ChatGPT, produce outputs that are remarkably human-like in some ways but are arguably very unlike humans in the way they work. They exemplify both the possibility of cases of this kind and the fact that companies are incentivised to build systems that can mimic humans.\footnote{See section 4.1.2 for discussion of the risk that there may soon be many non-conscious AI systems that seem conscious to users.} Schneider (2019) proposes to avoid gaming by restricting the access of systems to be tested to human literature on consciousness so that they cannot learn to mimic the way we talk about this subject. However, it is not clear either whether this measure would be sufficient, or whether it is possible to give the system enough access to data that it can engage with the test, without giving it so much as to enable gaming (Udell \& Schwitzgebel 2021).

\newpage

\pdfbookmark[0]{Scientific Theories of Consciousness}{sci-toc}
\section{Scientific Theories of Consciousness}

In this section, we survey a selection of scientific theories of consciousness, scientific proposals which are not exactly theories of consciousness but which bear on our project, and other claims from scientists and philosophers about putatively necessary conditions for consciousness. From these theories and proposals, we aim to extract a list of indicators of consciousness that can be applied to particular AI systems to assess how likely it is that they are conscious.\footnote{A similar approach to the question of AI consciousness is found in Chalmers (2023) which considers several features of LLMs which give us reason to think they are conscious, and several commonly-expressed ``defeaters'' for LLM consciousness. Many of these considerations are, like our indicators, drawn from scientific theories of consciousness.} Because we are looking for indicators that are relevant to AI, we discuss possible artificial implementations of theories and conditions for consciousness at points in this section. However, we address this topic in more detail in section 3, which is about what it takes for AI systems to have the features that we identify as indicators of consciousness.

Sections 2.1-2.3 cover recurrent processing theory, global workspace theory and higher-order theories of consciousness—with a particular focus in 2.3 on perceptual reality monitoring theory. These are established scientific theories of consciousness that are compatible with our computational functionalist framework. Section 2.4 discusses several other scientific theories, along with other proposed conditions for consciousness, and section 2.5 gives our list of indicators.

We do not aim to adjudicate between the theories which we consider in this section, although we do indicate some of their strengths and weaknesses. We do not adopt any one theory, claim that any particular condition is definitively necessary for consciousness, or claim that any combination of conditions is jointly sufficient. This is why we describe the list we offer in section 2.5 as a list of \textit{indicators} of consciousness, rather than a list of conditions. The features in the list are there because theories or theorists claim that they are necessary or sufficient, but our claim is merely that it is \textit{credible} that they are necessary or (in combination) sufficient because this is implied by credible theories. Their presence in a system makes it more probable that the system is conscious. We claim that assessing whether a system has these features is the best way to judge whether it is likely to be conscious given the current state of scientific knowledge of the subject.

\pdfbookmark[1]{Recurrent Processing Theory}{rpt}
\subsection{Recurrent Processing Theory}

\pdfbookmark[2]{Introduction to recurrent processing theory}{intro-rpt}
\subsubsection{Introduction to recurrent processing theory}

The recurrent processing theory (RPT; Lamme 2006, 2010, 2020) is a prominent member of a group of neuroscientific theories of consciousness that focus on processing in perceptual areas in the brain (for others, see Zeki \& Bartels 1998, Malach 2021). These are sometimes referred to as ``local'' (as opposed to ``global'') theories of consciousness because they claim that activity of the right form in relatively circumscribed brain regions is sufficient for consciousness, perhaps provided that certain background conditions are met. RPT is primarily a theory of visual consciousness: it seeks to explain what distinguishes states in which stimuli are consciously seen from those in which they are merely unconsciously represented by visual system activity. The theory claims that unconscious vs. conscious states correspond to distinct stages in visual processing. An initial feedforward sweep of activity through the hierarchy of visual areas is sufficient for some visual operations like extracting features from the scene, but not sufficient for conscious experience. When the stimulus is sufficiently strong or salient, however, recurrent processing occurs, in which signals are sent back from higher areas in the visual hierarchy to lower ones. This recurrent processing generates a conscious representation of an organised scene, which is influenced by perceptual inference—processing in which some features of the scene or percept are inferred from other features. On this view, conscious visual experience does not require the involvement of non-visual areas like the prefrontal cortex, or attention—in contrast with ``global'' theories like global workspace theory and higher-order theories, which we will consider shortly.

\pdfbookmark[2]{Evidence for recurrent processing theory}{evidence-rpt}
\subsubsection{Evidence for recurrent processing theory}

The evidence for RPT is of two kinds: the first is evidence that recurrent processing is necessary for conscious vision, and the second is evidence against rival theories which claim that additional processing for functions beyond perceptual organisation is required.

Evidence of the first kind comes from experiments involving backward masking and transcranial magnetic stimulation, which indicate that feedforward activity in the primary visual cortex (the first stage of processing mentioned above) is not sufficient for consciousness (Lamme 2006). Lamme also argues that, although feedforward processing is sufficient for basic visual functions like categorising features, important functions like feature grouping and binding and figure-ground segregation require recurrence. He, therefore, claims that recurrent processing is necessary for the generation of an organised, integrated visual scene—the kind of scene that we seem to encounter in conscious vision (Lamme 2010, 2020).

Evidence against more demanding rival theories includes results from lesion and brain stimulation studies suggesting that additional processing in the prefrontal cortex is not necessary for conscious visual perception. This counts against non-``local'' views insofar as they claim that functions in the prefrontal cortex are necessary for consciousness (Malach 2022; for a countervailing analysis see Michel 2022). Proponents of RPT also argue that the evidence used to support rival views is confounded by experimental requirements for downstream cognitive processes associated with making reports. The idea is that when participants produce the reports (and other behavioural responses) that are used to indicate conscious perception, this requires cognitive processes that are not themselves necessary for consciousness. So where rival theories claim that downstream processes are necessary for consciousness, advocates of RPT and similar theories respond that the relevant evidence is explained by confounding factors (see the methodological issues discussed in section 1.2.2).

\newpage

\pdfbookmark[2]{Indicators from recurrent processing theory}{indicators-rpt}
\subsubsection{Indicators from recurrent processing theory}

There are various possible interpretations of RPT that have different implications for AI consciousness. For our purposes, a crucial issue is that the claim that recurrent processing is necessary for

\parshape=4 0pt \linewidth 0pt \linewidth 0pt \linewidth 0pt \linewidth 
\begin{wrapfigure}{r}{0.6\textwidth} 
\centering
\includegraphics[width=0.6\textwidth,keepaspectratio]{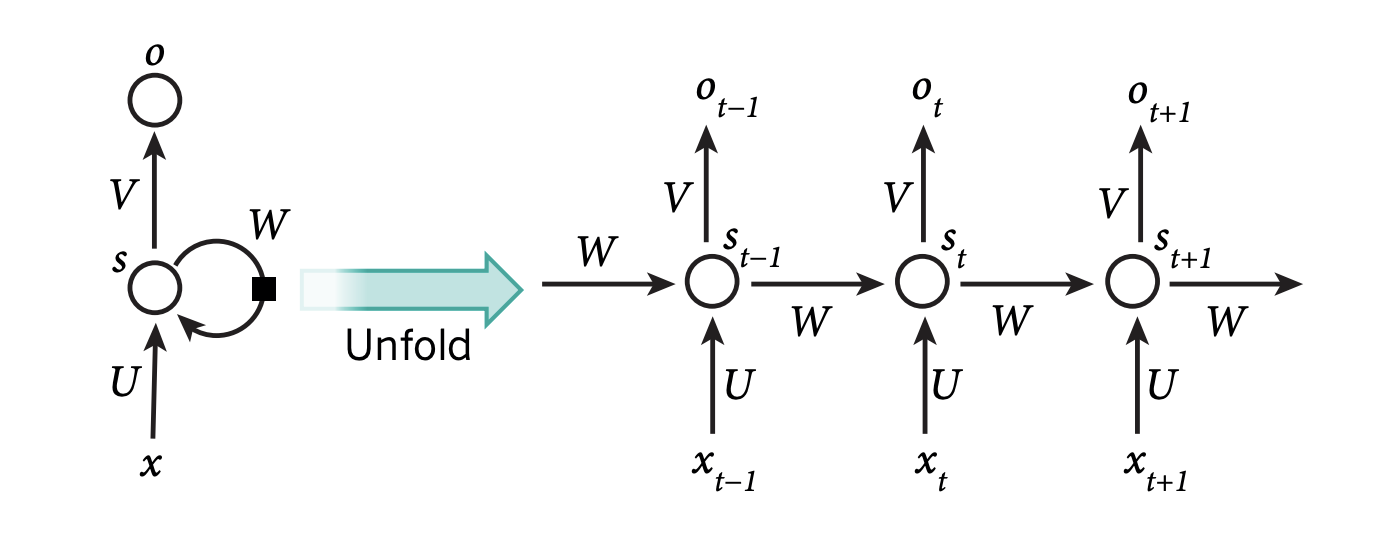} 
\caption{\textbf{An unfolded recurrent neural network} as depicted in LeCun, Bengio, \& Hinton (2015). Attribution-Share Alike 4.0 International.}
\end{wrapfigure}

\vspace{-\parskip}
\noindent consciousness can be interpreted in two different ways. In the brain, it is common for individual neurons to receive inputs that are influenced by their own earlier outputs, as a result of feedback loops from connected regions. However, a form of recurrence can be achieved without this structure: any finite sequence of operations by a network with feedback loops can be mimicked by a suitable feedforward network with enough layers. To achieve this, the feedforward network would have multiple layers with shared weights, so that the same operations would be performed repeatedly—thus mimicking the effect of repeated processing of information by a single set of neurons, which would be produced by a network with feedback loops (Savage 1972, LeCun et al. 2015). In current AI, recurrent neural networks are implemented indistinguishably from deep feedforward networks in which layers share weights, with different groups of input nodes for successive inputs feeding into the network at successive layers.

We will say that networks with feedback loops such as those in the brain, which allow individual physically-realised neurons to process information repeatedly, display \textit{implementational recurrence}. However, deep feedforward networks with weight-sharing display only \textit{algorithmic recurrence}—they are algorithmically similar to implementationally recurrent networks but have a different underlying structure. So there are two possible interpretations of RPT available here: it could be interpreted either as claiming that consciousness requires implementational recurrence, or as making only the weaker claim that algorithmic recurrence is required. Doerig et al. (2019) interpret RPT as claiming that implementational recurrence is required for consciousness and criticise it for this claim. However, in personal communication, Lamme has suggested to us that RPT can also be given the weaker algorithmic interpretation.

Implementational and algorithmic recurrence are both possible indicators of consciousness in AI, but we focus on algorithmic recurrence. It is possible to build an artificial system that displays implementational recurrence, but this would involve ensuring that individual neurons were physically realised by specific components in the hardware. This would be a very different approach from standard methods in current AI, in which neural networks are simulated without using specific hardware components to realise each component of the network. An implementational recurrence indicator would therefore be less relevant to our project, so we do not adopt this indicator. 

Using algorithmic recurrence, in contrast, is a weak condition that many AI systems already meet. However, it is non-trivial, and we argue below that there are other reasons, besides the evidence for RPT, to believe that algorithmic recurrence is necessary for consciousness. So we adopt this as our first indicator:

\myindicator{RPT-1: Input modules using algorithmic recurrence}

This is an important indicator because systems that lack this feature are significantly worse candidates for consciousness.

RPT also suggests a second indicator, because it may be interpreted as claiming that it is sufficient for consciousness that algorithmic recurrence is used to generate integrated perceptual representations of organised, coherent scenes, with figure-ground segregation and the representation of objects in spatial relations. This second indicator is:

\myindicator{RPT-2: Input modules generating organised, integrated perceptual representations}

An important contrast for RPT is between the functions of feature extraction and perceptual organisation. Features in visual scenes can be extracted in unconscious processing in humans, but operations of perceptual organisation such as figure-ground segregation may require conscious vision; this is why RPT-2 stresses organised, integrated perceptual representations.

There are also two further possible interpretations of RPT, which we set aside for different reasons. First, according to the biological interpretation of RPT, recurrent processing in the brain is necessary and sufficient for consciousness because it is associated with certain specific biological phenomena, such as recruiting particular kinds of neurotransmitters and receptors which facilitate synaptic plasticity. This biological interpretation is suggested by some of Lamme’s arguments (and was suggested to us by Lamme in personal communication): Lamme (2010) argues that there could be a ``fundamental neural difference" between feedforward and recurrent processing in the brain and that we should expect consciousness to be associated with a ``basic neural mechanism". We set this interpretation aside because if some particular, biologically-characterised neural mechanism is necessary for consciousness, artificial systems cannot be conscious.

Second, RPT may be understood as a theory only of visual consciousness, which makes no commitments about what is necessary or sufficient for consciousness more generally. On this interpretation, RPT would leave open both: (i) whether non-visual conscious experiences require similar processes to visual ones, and (ii) whether some further background conditions, typically met in humans but not specified by the theory, must be met even for visual consciousness. This interpretation of the theory is reasonable given that the theory has not been extended beyond vision and that it is doubtful whether activity in visual brain areas sustained \textit{in vitro} would be sufficient for consciousness (Block 2005). But on this interpretation, RPT would have very limited implications for AI.

\pdfbookmark[1]{Global Workspace Theory}{gwt}
\subsection{Global Workspace Theory}
\pdfbookmark[2]{Introduction to global workspace theory}{intro-gwt}
\subsubsection{Introduction to global workspace theory}

The global workspace theory of consciousness (GWT) is founded on the idea that humans and other animals use many specialised systems, often called modules, to perform cognitive tasks of particular kinds. These specialised systems can perform tasks efficiently, independently and in parallel. However, they are also integrated to form a single system by features of the mind which allow them to share information. This integration makes it possible for modules to operate together in co-ordinated and flexible ways, enhancing the capabilities of the system as a whole. GWT claims that one way in which modules are integrated is by their common access to a ``global workspace''—a further ``space'' in the system where information can be represented. Information represented in the global workspace can influence activity in any of the modules. The workspace has a limited capacity, so an ongoing process of competition and selection is needed to determine what is represented there.

GWT claims that what it is for a state to be conscious is for it to be a representation in the global workspace. Another way to express this claim is that states are conscious when they are ``globally broadcast'' to many modules, through the workspace. GWT was introduced by Baars (1988) and has been elaborated and defended by Dehaene and colleagues, who have developed a neural version of the theory (Dehaene et al. 1998, 2003, Dehaene \& Naccache 2001, Dehaene \& Changeux 2011, Mashour et al. 2020). Proponents of GWT argue that the global workspace explains why some privileged subset of perceptual (and other) representations are available at any given time for functions such as reasoning, decision-making and storage in episodic memory. Perceptual representations get stronger due to the strength of the stimulus or are amplified by attention because they are relevant to ongoing tasks; as a result, these representations ``win the contest'' for entry to the global workspace. This allows them to influence processing in modules other than those that produced them.

The neural version of GWT claims there is a widely distributed network of ``workspace neurons'', originating in frontoparietal areas, with activity in this network, which is sustained by recurrent processing, constituting conscious representations. When perceptual representations become sufficiently strong, a process called ``ignition'' takes place in which activity in the workspace neurons comes to code for their content. Ignition is a step-function, so whether a given representation is broadcast, and, therefore, conscious, is not a matter of degree.

\setlength{\intextsep}{20pt} 

GWT is typically presented as a theory of access consciousness—that is, of the phenomenon that some information represented in the brain, but not all, is available for rational decision-making. However, it can also be interpreted as a theory of phenomenal consciousness, motivated by the thought that access consciousness and phenomenal consciousness may coincide, or even be the same property, despite being conceptually distinct (Carruthers 2019). Since our topic is phenomenal consciousness, we interpret the theory in this way. It is notable that although GWT does not explicitly require agency, it can only explain access consciousness if the system is a rational agent since access consciousness is defined as availability for rational control of action (we discuss agency in section 2.4.5)


\newpage

\begin{figure}[H] 
\centering
\includegraphics[height=0.525\textheight,keepaspectratio]{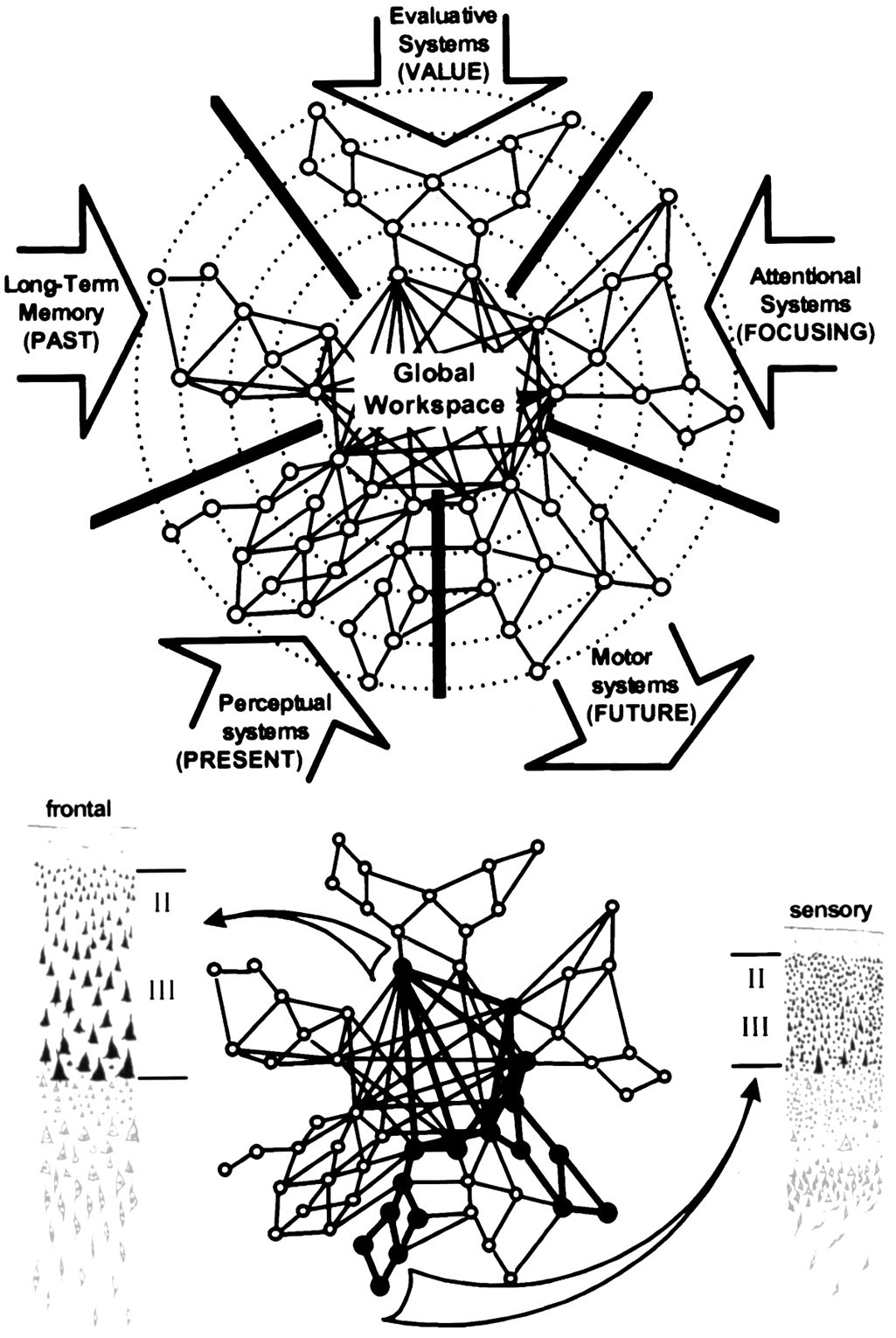} 
\caption{\textbf{Global Workspace.} The figure used by Dehaene et al. (1998) to illustrate the basic idea of a global workspace. Note that broadcast to a wide range of consumer systems such as planning, reasoning and verbal report does not feature in the figure. 1998. National Academy of Sciences. Reprinted with permission.}
\end{figure}


\pdfbookmark[2]{Evidence for global workspace theory}{evidence-gwt}
\subsubsection{Evidence for global workspace theory}

There is extensive evidence for global workspace theory, drawn from many studies, of which we can mention only a few representative examples (see Dehaene 2014 and Mashour et al. 2020 for reviews). These studies generally employ the method of contrastive analysis, in which brain activity is measured and a comparison is made between conscious and unconscious conditions, with efforts made to control for other differences. Various stimuli and tasks are used to generate the conscious and unconscious conditions, and activity is measured using fMRI, MEG, EEG, or single-cell recordings. According to GWT advocates, these studies show that conscious perception is associated with reverberant activity in widespread networks which include the prefrontal cortex (PFC)—this claim contrasts with the ``local'' character of RPT discussed above—whereas unconscious states involve more limited activity confined to particular areas. This widespread activity seems to arise late in perceptual processing, around 250-300ms after stimulus onset, supporting the claim that global broadcast requires sustained perceptual representations (Mashour et al. 2020).

Examples of recording studies in monkeys that support a role for PFC in consciousness include experiments by Panagiotaropoulos et al. (2012) and van Vugt et al. (2018). In the former study, researchers were able to decode the presumed content of conscious experience during binocular rivalry from activity in PFC (Panagiotaropoulos et al. 2012). In this study the monkeys viewed stimuli passively—in contrast with many studies supporting GWT—so the results are not confounded by behavioural requirements (this was a no-report paradigm; see sections 1.2.2 and 2.1.2). In the latter, activity was recorded from the visual areas V1 and V4 and dorsolateral PFC, while monkeys performed a task requiring them to respond to weak visual stimuli with eye movements. The monkeys were trained to move their gaze to a default location if they did not see the stimulus and to a different location if they did. Seen stimuli were associated with stronger activity in V1 and V4 and late, substantial activity in PFC. Importantly, while early visual activity registered the objective presence of the stimulus irrespective of the animal’s response, PFC activity seemed to encode the conscious percept, as this activity was also present in false alarms—cases in which monkeys acted as though they had seen a stimulus even though no stimulus was present. Activity associated with unseen stimuli tended to be lost in transmission from V1 through V4, to PFC. Studies on humans using different measurement techniques have similarly found that conscious experience is associated with ignition-like activity patterns and decodability from PFC (e.g. Salti et al. 2015).

\pdfbookmark[2]{Indicators from global workspace theory}{indicators-gwt}
\subsubsection{Indicators from global workspace theory}

We want to identify the conditions which must be met for a system to be conscious, according to GWT, because these conditions will be indicators of consciousness in artificial systems. This means that a crucial issue is exactly what it takes for a system to implement a global workspace. Several authors have noted that it is not obvious how similar a system must be to the human mind, in respect of its workspace-like features, to have the kind of global workspace that is sufficient, in context, for consciousness (Bayne 2010, Carruthers 2019, Birch 2022b, Seth \& Bayne 2022). There are perhaps four aspects to this problem. First, workspace-like architectures could be used with a variety of different combinations of modules with different capabilities; as Carruthers (2019) points out, humans have a rich and specific set of capabilities that seem to be facilitated by the workspace and may not be shared with other systems. So one question is whether some specific set of modules accessing the workspace is required for workspace activity to be conscious. Second, it’s unclear what degree of similarity a process must bear to selection, ignition and broadcasting in the human brain to support consciousness. Third, it is difficult to know what to make of possible systems which use workspace-like mechanisms but in which there are multiple workspaces—perhaps integrating overlapping sets of modules—or in which the workspaces are not global, in the sense that they do not integrate all modules. And fourth, there are arguably two stages involved in global broadcast—selection for representation in the workspace, and uptake by consumer modules—in which case there is a question about which of these makes particular states conscious.

Although these questions are difficult, it is possible that empirical evidence could be brought to bear on them. For example, studies on non-human animals could help to identify a natural kind that includes the human global workspace and facilitates consciousness-linked abilities (Birch 2020). Reflection on AI can also be useful here because we can recognise functional similarities and dissimilarities between actual or possible systems and the hypothesised global workspace, separately from the range of modules in the system or the details of neurobiological implementation, and thus develop a clearer sense of the possible functional kinds in this area.

Advocates of GWT have argued that the global workspace facilitates a range of functions in humans and other animals (Baars 1988, Shanahan 2010). These include making it possible for modules to exert ongoing control over others for the duration of a task (e.g. in the case of searching for a face in a crowd), and dealing with novel stimuli by broadcasting information about them, thus putting the system in a position to learn the most effective response. Global broadcast and the capacity to sustain a representation over time, while using it to process incoming stimuli, are necessary for these functions. Because the global workspace requires that information from different modules is represented in a common ``language'', it also makes it possible to learn and generate crossmodal analogies (VanRullen \& Kanai 2021, Goyal et al. 2022). A particularly sophisticated and notable possible function of the global workspace is ``System 2 thought'', which involves executing strategies for complex tasks in which the workspace facilitates extended and controlled interactions between modules (Kahneman 2011, VanRullen \& Kanai 2021, Goyal \& Bengio 2022). For example, planning a dinner party may involve engaging in an extended process, controlled by this objective, of investigative actions (looking to see what is in the fridge), calls to episodic memory, imagination in various modalities (how the food will taste, how difficult it will be to cook, how the guests will interact), evaluation and decision-making. In this case, according to the theory, the workspace would maintain a representation of the goal, and perhaps compressed summaries of interim conclusions, and would pass queries and responses between modules.

We argue that GWT can be expressed in four conditions of progressively increasing strength. Systems that meet more of these conditions possess more aspects of the full global workspace architecture and are, therefore, better candidates for consciousness.

The first condition is possessing specialised systems which can perform tasks in parallel. We call these systems ``modules'', but they need not be modules in the demanding sense set out by Fodor (1983); they need not be informationally encapsulated or use dedicated components of the architecture with functions assigned prior to training. Mashour et al.’s recent statement of the global neuronal workspace hypothesis claims only that modules in which unconscious processing takes place are localised and specialised, and that they process ``specific perceptual, motor, memory and evaluative information'' (2020, p. 777). It may be that having more independent and differentiated modules makes a system a better candidate for consciousness, but GWT is most plausibly interpreted as claiming that what matters for consciousness is the process that integrates the modules, rather than their exact characteristics. The first indicator we draw from this theory is, therefore:

\myindicator{GWT-1: Multiple specialised systems capable of operating in parallel (modules)}

Building on this, a core condition of GWT is the existence of a bottleneck in information flow through the system: the capacity of the workspace must be smaller than the collective capacity of the modules which feed into it. Having a limited capacity workspace enables modules to share information efficiently, in contrast to schemes involving pairwise interactions such as Transformers, which become expensive with scale (Goyal et al. 2022, Jaegle et al. 2021a). The bottleneck also forces the system to learn useful, low-dimensional, multimodal representations (Bengio 2017, Goyal \& Bengio 2022). With the bottleneck comes a requirement for an attention mechanism that selects information from the modules for representation in the workspace. This yields our second indicator:

\myindicator{GWT-2: Limited capacity workspace, entailing a bottleneck in information flow and a selective attention mechanism}

A further core condition is that information in the workspace is globally broadcast, meaning that it is available to all modules. The two conditions we have seen so far are not enough to ensure that ongoing interaction between modules is possible, or that information in the workspace is available to multiple output modules which can use it for different tasks. Our third indicator is, therefore:

\myindicator{GWT-3: Global broadcast: availability of information in the workspace to all modules}

This entails that all modules must be able to take inputs from the global workspace, including those modules which process inputs to the system as a whole. The first two conditions can be satisfied by wholly feedforward systems which have multiple input modules, feeding into a limited-capacity workspace, from which information then flows on to one or more output modules. But this new condition entails that information must also flow back from the workspace to the input modules, influencing their processing. In turn, this means that the input modules must be (algorithmically) recurrent—and thus provides further justification for indicator RPT-1—although output modules, which map workspace states to behaviour, need not be recurrent.

Finally, for the workspace to facilitate ongoing, controlled interactions between modules it must have one further feature. This is that the selection mechanism that determines information uptake from the modules must be sensitive to the state of the system, as well as to new inputs. That is, the system must implement a form of ``top-down attention'' as well as ``bottom-up attention''. This allows representations in the workspace itself or in other modules to affect which information is selected from each module. State-dependent selection can be readily implemented by systems that meet GWT-3 because global broadcast entails that information flows from the workspace to the modules. Generating controlled, functional interactions between modules, however, will require that the system as a whole is suitably trained. Our fourth indicator is:

\myindicator{GWT-4: State-dependent attention, giving rise to the capacity to use the workspace to query modules in succession to perform complex tasks}

Compared to other scientific theories of consciousness, many more proposals have been made for the implementation of GWT in artificial systems (e.g. Franklin \& Graesser 1999, Shanahan 2006, Bao et al. 2020). We discuss implementations of GWT, together with other theories, in section 3.1.

\newpage


\begin{infobox} [Box 3: Attention in neuroscience and in AI]

The fields of neuroscience and machine learning each have their own distinct concepts of attention (Lindsay 2020). In machine learning, several different forms of attention have been developed, but at present, the most common is ``self-attention'' (Vaswani et al. 2023). This is the mechanism at the heart of Transformer networks, which power large language models.

\begin{wrapfigure}{L}{0.4\textwidth}
\centering
\includegraphics[width=0.36\textwidth]{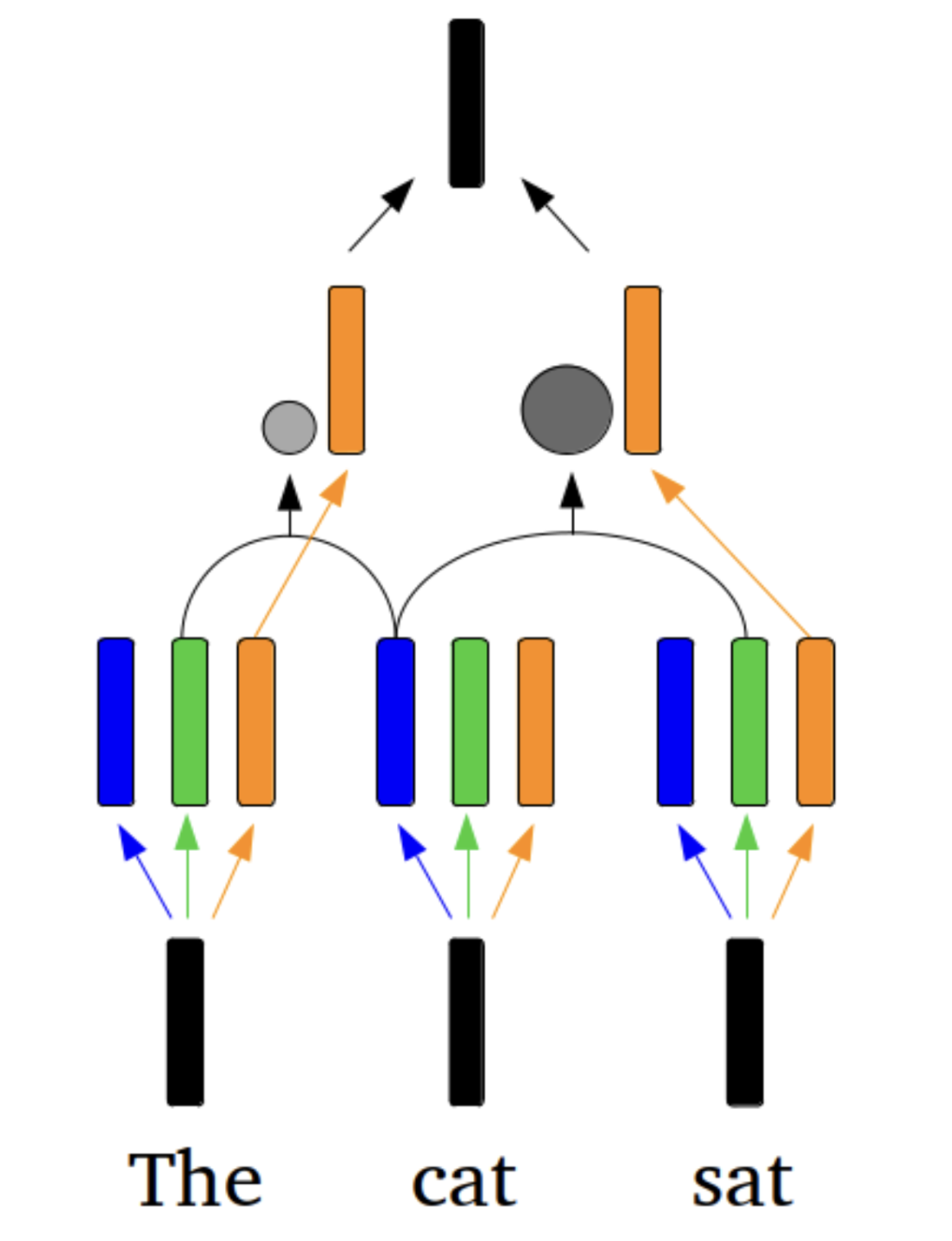}
\end{wrapfigure}
\vspace{2pt}
In self-attention, representations of elements of an input sequence (for example, words in a sentence) are allowed to interact multiplicatively. Specifically, each \textbf{word representation}, that is given as a vector is transformed into three new vectors: a \textcolor{blueclr}{query}, \textcolor{greenclr}{key}, and \textcolor{orangecolor}{value}. The \textcolor{blueclr}{query} vector of one word is multiplied by the \textcolor{greenclr}{key} vectors of all other words to determine a \textcolor{greyclr}{weighting} for each of these words. This \textcolor{greyclr}{weighting} is applied to the \textcolor{orangecolor}{value} vectors of these words; the sum of these \textcolor{greyclr}{weighted} \textcolor{orangecolor}{value} vectors forms the new \textbf{representation} of the word. This process is done in parallel for all words.
\infoparthree
``Cross-attention'' follows a similar formula but allows the query to be generated from one set of representations and the key and values to come from another (self-attention and cross-attention are both forms of ``key-query attention''). This can be helpful, for example, in translation networks that use the words of the sentence being generated in the target language to guide attention toward the appropriate words of the sentence in the original language. 
\infoparthree
Key-query attention has only loose connections to how attention is conceptualised in neuroscience. Similar to self-attention, gain modulation (wherein attention multiplicatively scales neural activity) has been found in many neural systems (Treue \& Trujillo 1999, Reynolds \& Heeger 2009). However, this attentional modulation is frequently thought to arise from recurrent top-down connections, not from the parallel processing of concurrent inputs (Noudoost et al. 2010, Bichot et al. 2015). Previous versions of attention in machine learning have relied on recurrent processing, and in this way could be considered more similar to biological attention (Mnih et al. 2014, Bahdanau et al. 2014). However, it should be noted that there are many different flavors of attention within neuroscience and the underlying neural mechanisms may vary across them. Therefore, saying definitively which forms of artificial attention are closest to biological attention in general, is not straightforward.  
\infoparthree
Insofar as different theories of consciousness depend on recurrent processing or other specific components of the attention mechanism, self-attention may not be sufficient to form the basis of artificial consciousness. For example, there is nothing akin to the binary ignition process in global workspace theory in self-attention, as attention is implemented as a graded weighting of inputs. There is also no built-in model of the attention process on top of attention itself, as required in attention schema theory.
\end{infobox}

\pdfbookmark[1]{Higher-Order Theories}{hot}
\subsection{Higher-Order Theories}

\pdfbookmark[2]{Introduction to higher-order theories}{intro-hot}
\subsubsection{Introduction to higher-order theories}
The core claim of higher-order theories of consciousness is helpfully distilled by Brown et al. (2019):

\begin{myquote}
The basic idea … is that conscious experiences entail some kind of minimal inner awareness of one’s ongoing mental functioning, and this is due to the first-order state being in some ways monitored or meta-represented by a relevant higher-order representation. (p. 755)
\end{myquote}

Higher-order theories are distinguished from others by the emphasis that they place on the idea that for a mental state to be conscious the subject must be aware of being in that mental state, and the way in which they propose to account for this awareness. This is accounted for by an appeal to higher-order representation, a concept with a very specific meaning. Higher-order representations are ones that represent something about \textit{other representations}, whereas first-order representations are ones that represent something about the (non-representational) world. This distinction can be applied to mental states. For example, a visual representation of a red apple is a first-order mental state, and a belief that one has a representation of a red apple is a higher-order mental state.

Higher-order theories have long been advocated by philosophers (Carruthers \& Gennaro 2020, Rosenthal 2005). One of the main motivations for the view is the so-called ``simple argument'' (Lycan 2001): if a mental state is conscious, the subject is aware that they are in that state; being aware of something involves representing it; so consciousness requires higher-order representation of one’s own mental states. The substantive commitment of this argument is that there is a single sense of ``awareness'' of mental states on which both premises are true—which is both weak enough that consciousness entails awareness of mental states, and strong enough that this awareness entails higher-order representation. In the last two decades, higher-order theories have been elaborated, refined and tested by neuroscientists, and influenced by new experimental methods and ideas from the study of metacognition, signal detection theory, and the theory of predictive processing.

A variety of higher-order theories have been proposed, which describe distinct forms of monitoring or meta-representation, and imply different conditions for consciousness (Brown et al. 2019).  They include: several philosophical theories, including higher-order thought theory (Rosenthal 2005) and higher-order representation of a representation theory (Brown 2015); the self-organising meta-representational account (Cleeremans et al. 2020); higher-order state space theory (Fleming 2020); and perceptual reality monitoring theory (Lau 2019, 2022, Michel forthcoming). We will concentrate on perceptual reality monitoring theory (PRM) and to some degree also the closely-related higher-order state space theory (HOSS). These are both recent computational theories based on extensive assessments of neuroscientific evidence.

The core claim of PRM is that consciousness depends on a mechanism for distinguishing meaningful activity in perceptual systems from noise. There are multiple possible sources of neural activity in perceptual systems. This activity could be caused by perceptible stimuli in the environment; it could be sustained after these stimuli have passed; it could be generated top-down through expectations, imagination, dreaming or episodic memory; or it could be due to random noise. PRM claims that a ``reality monitoring'' mechanism, which operates automatically, is used to discriminate between these different kinds of activity and assess the reliability of first-order representations. Perceptual representations are conscious when they are identified as reliable, or in other words, as being sufficiently different from noise.

Meanwhile, HOSS makes the similar claim that ``awareness is a higher-order state in a generative model of perceptual contents'' (Fleming 2020, p. 2). This higher-order state, which is the product of a metacognitive inference, signals the probability that some particular content is represented in the perceptual system. This is presented as a theory of the basis of awareness reports (i.e. reports of the form ``I am/not aware of X''), but Fleming suggests that higher-order awareness states are necessary for consciousness.

\pdfbookmark[2]{Computational HOTs and GWT}{comp-hot}
\subsubsection{Computational HOTs and GWT}

Computational higher-order theories are sometimes grouped together with GWT as ``global'' theories, in opposition to ``local'' theories such as RPT (Michel \& Doerig 2022). Like GWT, higher-order theories such as PRM claim that cognitive functions supported by the prefrontal cortex play an important role in consciousness. As such, the evidence reviewed above in favour of the involvement of the PFC in consciousness supports PRM as well as GWT. PRM also claims, again like GWT, that ``consciousness is the \textit{gating mechanism} by which perception impacts cognition; it selects what perceptual information should directly influence our rational thinking'' (Lau 2022, p. 159). Lau (2022) endorses the existence of global broadcast as a phenomenon in the brain, and also affirms that it is \textit{related} to consciousness: when representations are conscious, ``global broadcast and access'' of that representation ``are \textit{likely} to happen'' (p. 159).

However, higher-order theorists reject the claim that broadcast in the global workspace is necessary and sufficient for consciousness. Notably, according to higher-order theories, unconscious representations can be encoded in the global workspace, which implies that a representation might be unconscious and yet available for high-level cognitive processes, such as reasoning. Higher-order theories and GWT make distinct predictions, and advocates of computational HOTs appeal to experiments testing these predictions as providing important evidence in favour of their view.

In one such experiment, Lau and Passingham (2006) conducted a visual discrimination task under a range of different masking conditions and asked participants to press a key to indicate whether they had seen or merely guessed the shape of the stimulus in each trial. They identified two different masking conditions in which participants’ ability to discriminate between stimuli was at the same level, but differed in how likely they were to report having seen the stimulus. Higher-order theorists interpret this result as showing that there can be a difference in conscious perception of a stimulus without a corresponding difference in task performance, and claim that this result is inconsistent with a prediction of GWT (Lau \& Rosenthal 2011). This purported prediction of GWT is that differences in consciousness should entail differences in task performance, because—according to GWT—consciousness makes information available to a wide range of cognitive functions, useful across a wide range of tasks. Furthermore, according to GWT, ignition leading to global broadcast is necessary and sufficient for consciousness, and ignition depends on the same factors which affect visual task performance, such as signal strength and attention.

The broader argument here is that evidence for GWT is confounded by differences in performance between conscious and unconscious conditions (Morales et al. 2022). Ignition leading to global broadcast is correlated with better performance on many tasks, and in the absence of controls, better performance is also correlated with consciousness. But since experiments seem to show that performance and consciousness are dissociable, it is possible that ignition is neither necessary nor sufficient for consciousness (Fleming 2020, Lau 2022).

\pdfbookmark[2]{Indicators from computational HOTs}{indicators-hot}
\subsubsection{Indicators from computational HOTs}

As we have seen, PRM claims that perceptual states are conscious when they are identified as reliable by a metacognitive monitoring mechanism. This mechanism outputs higher-order representations which label first-order states as accurate representations of reality. Similarly, HOSS claims that consciousness depends on higher-order awareness states directed at perceptual representations. Therefore, these two theories both suggest that metacognitive monitoring of perceptual systems with relevant properties is necessary for consciousness in AI.

We propose two indicators based on this claim:

\myindicator{HOT-1: Generative, top-down or noisy perception modules}

\vspace{3pt}
    
\myindicator{HOT-2: Metacognitive monitoring distinguishing reliable perceptual representations from noise}

HOT-1 is an indicator of consciousness because, according to computational HOTs, the function of the monitoring mechanisms which are responsible for consciousness is to discriminate between different sources of activity in perceptual systems. This means that consciousness is more likely in systems in which there are multiple possible sources of such activity. Examples of such systems include ones in which perceptual representations can be produced top-down in imagination, as well as ones affected by random noise. HOT-2 is a statement of the main necessary condition for consciousness according to computational HOTs.

Lau (2022) suggests that generative adversarial networks (GANs) may possess these two indicators, a possibility which we discuss further in section 3.1.3. However, PRM advocates claim that current AI systems do not meet the conditions of their theory (Michel \& Lau 2021, Lau 2022). They emphasise that perceptual reality monitoring systems must have a further feature: in addition to discriminating between perceptual states, a perceptual reality monitoring mechanism must output to a system for ``general belief-formation and rational decision-making'' (Michel \& Lau 2021). This condition is justified on the grounds that conscious experience has a certain ``assertoric force''. Some of our conscious perceptual experiences present themselves to us as accurate impressions of the outside world, and it is difficult for us to resist believing that things are as these experiences represent them.\footnote{Imaginative experiences are an exception, but PRM can claim that they are classified differently by the reality monitoring mechanism, and, therefore, have a different phenomenal character. Alternatively, higher-order theories like HOSS and PRM can hold that imaginative experiences have some minimal amount of assertoric force, thus explaining results in which participants are more likely to report a target as visible if it is congruent with their mental imagery (Dijkstra et al. 2021, 2022; Dijkstra \& Fleming 2023).} Even if we believe that we are subject to an illusion, an impression that our conscious experience is representing the world as it is remains—knowing about the Müller-Lyer illusion does not prevent the two lines from looking unequal. Such experiences are persistent inputs to cognition, which are not under direct cognitive control.

Another aspect of this idea is that what makes it the case that the monitoring mechanism labels some perceptual contents as ``real'' is that the person or system as a whole tends to take them to be real when they are so labelled. This entails that the system which includes the monitoring mechanism must be an agent which relies on perceptual representations tagged as ``real'' when selecting actions. The function of the reality monitoring mechanism, then, is to identify which perceptual states are accurate enough to be relied on in this way. Advocates of PRM propose that to rely on perceptual content is a matter of believing that content—given their picture of belief, this implies a system for reasoning and action selection with a holistic character, in which any belief can in principle be called on in examining any other or in reasoning about what to do. These claims give us our third indicator arising from HOTs:

\myindicator{HOT-3: Agency guided by a general belief-formation and action selection system, and a strong disposition to update beliefs in accordance with the outputs of metacognitive monitoring}

Computational HOTs also make a further claim which yields a fourth indicator. Like most scientific theories of consciousness, PRM aims to answer the question ``what makes a state conscious, rather than unconscious?'' However, it also attempts to answer the further question ``why do conscious mental states feel the way they do?'' Its answer to this second question appeals to quality space theory, a view that claims that phenomenal qualities can be reduced to the discriminations they allow for the system (Clark 2000, Rosenthal 2010, Lau et al. 2022). For instance, two features feel the same in virtue of the fact that, from the perspective of the system, they are indiscriminable (Rosenthal 2010). According to this proposal, the subjective similarity of two experiences is the inverse of their discriminability, and the experience of subjective qualities depends on implicit knowledge of the similarity space. As such, quality space theory provides a functional account of qualities (see Lau et al. 2022). For example, to have a conscious experience of the red colour of a tulip one must have an implicit grasp of its similarity to the colour of a red apple and its discriminability from the green of a new leaf. One hypothesis is that this implicit knowledge depends on sparse and smooth coding in perceptual systems—that is, on qualities being represented by relatively few neurons, and represented according to a continuous coding scheme rather than one which divides stimuli into absolute categories (Lau et al. 2022).

Importantly, PRM claims that consciousness is not possible without qualities, so although quality space theory is not a theory of what makes a state conscious, the posits of this theory are putative necessary conditions for consciousness. Our final indicator arising from HOTs is, therefore:

\myindicator{HOT-4 Sparse and smooth coding generating a ``quality space''}

This condition may be relatively readily met in AI: all deep neural networks use smooth representation spaces, and sparseness can also be achieved by familiar machine learning techniques (see section 3.1.3).

\pdfbookmark[1]{Other Theories and Conditions}{other-theories}
\subsection{Other Theories and Conditions}
Many scientific theories of consciousness have been proposed (see Seth \& Bayne 2022 for a list). In addition, there are influential theoretical proposals that are not exactly theories of consciousness, but which bear on our investigation. Furthermore, there may be necessary conditions for consciousness that are not explicitly emphasised in scientific theories because all humans meet them (such as having a body), but which need to be considered in the context of AI. In this section, we survey several relevant theories, proposals and conditions, before presenting our indicators in section 2.5.

One theory that we do not discuss below is integrated information theory (IIT; Oizumi et al. 2014, Tononi \& Koch 2015). The standard construal of IIT is incompatible with our working assumption of computational functionalism; Tononi \& Koch (2015) hold that a system that implemented the same algorithm as the human brain would not be conscious if its components were of the wrong kind. Relatedly, proponents of IIT claim that the theory implies that digital computers are unlikely to be conscious, whatever programs they run (Albantakis \& Tononi 2021). As a result, in contrast to other scientific theories, IIT does not imply that some AI systems built on conventional hardware would be better candidates for consciousness than others; this makes it less relevant to our project. It has recently been proposed that measures of information integration may be correlated with ``global states of consciousness'' such as wakefulness, sleep and coma—a paradigm called ``weak IIT'' (Michel \& Lau 2020, Mediano et al. 2022). But the implications of weak IIT for our project are limited: it suggests that measurable properties of integration and differentiation matter for consciousness, but does not (yet) tell us which measures to rely on or how to interpret their results when applied to artificial systems.

\pdfbookmark[2]{Attention Schema Theory}{ast}
\subsubsection{Attention Schema Theory}

The attention schema theory of consciousness (AST) claims that the human brain constructs a model of attention, which represents—and may misrepresent—facts about the current objects of attention. This model helps the brain to control attention, in a similar way to how the body schema helps with control of bodily movements. Conscious experience depends on the contents of the attention schema. For example, I will have the conscious experience of seeing an apple if the schema represents that I am currently attending to an apple (Webb \& Graziano 2015, Graziano 2019a). Attention schema theory claims that the workings of the attention schema explain our intuitions about our experiences: because the attention schema does not represent the details of the mechanism of attention, the theory claims, it seems to us that we are related to the stimulus (e.g. an apple) in an immediate and seemingly mysterious way.

AST can be thought of as a higher-order theory of consciousness because it claims that consciousness depends on higher-order representations of a particular kind (in this case, representations of our attention). Like other higher-order theories, it places special emphasis on our awareness of our own mental states. Unlike most higher-order theories, however, AST has been developed with the specific aim of explaining what we believe and say about consciousness, such as that we are conscious and that consciousness seems difficult to square with physical descriptions of the world (that is, AST aims to solve the meta-problem of consciousness—see Chalmers 2018, Graziano 2019b). Because it focuses on explaining why we (potentially mistakenly) believe certain things about consciousness, AST could be construed as an attempt to explain \textit{away} consciousness. But it is also open to interpretation as an account of the conditions for consciousness.

AST gives us a further indicator of consciousness in AI:

\myindicator{AST-1: A predictive model representing and enabling control over the current state of attention}

Representing the current state of attention allows the mind to learn about both the effects of attention and how attention is affected by events in the mind and the environment. Having a model, therefore, makes it easier for the mind to learn to take attention-affecting actions because they will have beneficial effects on other cognitive processes. A predictive model is especially valuable because it allows the mind to anticipate how the objects of attention might change, conditional on changes in the mind or the environment, and make adjustments accordingly. These could include preemptive adjustments, for instance when distraction from an important task is anticipated. Attention schema-like models which enable the effective control of attention could be valuable for the increasing number of AI systems which employ attention—here understood as active control of information flow (Liu et al. 2023).

\pdfbookmark[2]{Predictive Processing}{pp}
\subsubsection{Predictive Processing}

Predictive processing (PP) is presented as a comprehensive, unifying theory of human cognition, and has been used as a framework for addressing many questions about the mind. Because it is a general framework, some PP theorists describe it as a theory \textit{for} consciousness, not a theory \textit{of} consciousness (Seth \& Hohwy 2021, Seth \& Bayne 2022)—a paradigm within which theories of consciousness should be developed. However, advocates of PP have used it to explain many specific features of conscious experience, such as the puzzling nature of ``qualia'' (Clark 2019) and the phenomenology of emotion and embodiment (Seth 2021). Consciousness has been discussed extensively by PP theorists, but relatively little direct attention has been given to the questions which are most important for our purposes: what distinguishes conscious from non-conscious systems, and what distinguishes conscious from non-conscious states within conscious systems (Deane 2021, Hohwy 2022, Nave et al. 2022).

PP claims that the essence of human and animal cognition is minimisation
of errors made by a hierarchical generative model in predicting sensory
stimulation. In perception, this model is continually generating
predictions at multiple levels, each influenced by predictions at
neighbouring levels and in the immediate past, and by prediction error
signals which ultimately arise from sensory stimulation itself. This
process is modulated by attention, and—according to the thesis of
`active inference'—it can also control action because acting can be a
means of reducing prediction error. Adaptive actions will be selected if
organisms predict their own success (Friston 2010).

Although PP is not a theory of consciousness, its popularity means that
many researchers regard predictive processing as a plausible necessary
condition for consciousness. We, therefore, include the use of
predictive coding among our indicators:

\myindicator{PP-1 Input modules using predictive coding}

In keeping with the ``theory \textit{for} consciousness'' idea, the PP
framework has been employed in developments of GWT and HOT. Hohwy (2013)
and Whyte (2019) propose that global broadcast (via ignition) takes
place when the process of perceptual inference settles on some
representation of the state of the environment as most probable, then
makes this representation available as the basis for active inference.
Meanwhile, the higher-order state space theory adopts the PP framework
(Fleming 2020).

\pdfbookmark[2]{Midbrain Theory}{midbrain}
\subsubsection{Midbrain Theory}

While the neuroscientific theories of consciousness we have discussed so far focus primarily on cortical processes, Merker (2007) argues that the cortex is not necessary for consciousness. This view has been particularly influential in recent discussions of consciousness in non-human animals. Merker’s proposal is that activity in parts of the midbrain and basal ganglia constitute a ``unified multimodal neural model of the agent within its environment, which is weighted by the current needs and state of the agent'' (Klein \& Barron 2016), and that this activity is sufficient for subjective experience. In Merker’s account, one midbrain region, the superior colliculus, integrates information from spatial senses and the vestibular system to construct a model of the organism’s position and movement in space. Other regions including the hypothalamus, periaqueductal gray, and parts of the basal ganglia bring in information about the organism’s physiological state and contribute to identifying opportunities and selecting actions. In functional terms, the midbrain theory claims that consciousness depends on ``integrated spatiotemporal modeling'' for action selection (Klein \& Barron 2016).

Birch (2022b) summarises and criticises evidence for the midbrain theory. For our purposes, it is notable because it offers a particular perspective on the significance of cognitive integration for consciousness. The midbrain theory proposes that the integration which is necessary for consciousness arose to solve the biologically ancient problem of decision-making in complex mobile animals, particularly the need to distinguish the effects of self-caused motion on perceptual input (Merker 2005). Hence the midbrain theory emphasises the need to integrate specific kinds of information, such as spatial, affective, and homeostatic information, into a single common model. This theory, therefore, gives us additional reason to believe that systems for purposeful navigation of a body through space are necessary for consciousness, and thus contribute to the case for the indicators we present in section 2.4.5.

\pdfbookmark[2]{Unlimited Associative Learning}{ual}
\subsubsection{Unlimited Associative Learning}

Another influential theory in animal consciousness literature is Ginsburg and Jablonka’s (2019) unlimited associative learning framework (Birch et al. 2020). The proposal here is that the capacity for unlimited associative learning (UAL) is an evolutionary ``transition marker'' for consciousness: a single feature that indicates that an evolutionary transition to consciousness has taken place in a given lineage. Ginsburg and Jabolanka identify UAL as a marker on the grounds that it requires a combination of several ``hallmarks'' that are argued to be jointly sufficient for consciousness in living organisms. The UAL framework, therefore, brings together a list of features that seem to be related to consciousness and argues that they are unified by facilitating UAL.

Somewhat like this report, the UAL project aims to identify a set of jointly sufficient conditions for consciousness. This set is summarised in the table below (which follows the presentation of these conditions in Birch et al. 2020).

\begin{center} 
\begin{tabular}{|p{0.8\textwidth}|} 
\hline
\multicolumn{1}{|c|}{\textbf{Hallmarks of consciousness according to UAL}} \\ 
\hline
\rowcolor{lightgray1} Global accessibility integrating sensory, evaluative and mnemonic information  \\
\hline
\rowcolor{lightgray2} Selective attention \\
\hline
\rowcolor{lightgray1} Integration over time through forms of short-term memory \\
\hline
\rowcolor{lightgray2} Embodiment and agency \\
\hline
\rowcolor{lightgray1} Self-other registration, used in constructing a representation of the moving body in space \\
\hline
\rowcolor{lightgray2} Flexible value system capable of revaluation and weighing needs \\
\hline
\rowcolor{lightgray1} Binding/unification of features to form compound stimuli, enabling discrimination of complex patterns \\
\hline
\rowcolor{lightgray2} Intentionality, i.e. representation of the body and environment \\
\hline
\end{tabular}
\end{center}

This list of conditions is similar to the conditions which are emerging from our work in this section (like us, Ginsburg and Jablonka examined scientific theories of consciousness in developing their list). The global accessibility and selective attention conditions are met by systems that implement global workspace architectures. The integration over time condition may be as well; we discuss integration over time in section 2.4.6. Several of the other conditions are related to agency and embodiment, which we discuss in section 2.4.5. These include not only the embodiment and agency condition itself but also self-other registration (which also aligns with the midbrain theory) and having a flexible value system. We take it that the binding/unification condition will be met by any neural network-based AI system because these systems are designed to learn to discriminate complex patterns. The final condition, that the system should represent its body and the environment, raises philosophical questions about intentionality which we will not go into here. But any naturalistic theory of intentionality is likely to entail that systems that meet the other conditions will also meet this one because the other conditions entail the performance of the functions which might ground intentionality.

What about the capacity for unlimited associative learning itself? This could be argued to be an indicator of consciousness in artificial systems. In the UAL framework, it is characterised as an open-ended capacity for associative learning. In particular, to have this capacity an organism must be capable of conditioning with compound stimuli and novel stimuli and of quickly and flexibly updating the values it associates with stimuli, actions and outcomes. It must also be capable of second-order conditioning, meaning that it can link together chains of associations, and trace conditioning, which is learning an association when there is a time gap between stimuli. Having the capacity for unlimited associated learning, therefore, implies that an organism or system can integrate information across modalities and over time, and that can evaluate stimuli and change these evaluations in response to new information.

This capacity is also, therefore, linked to many of our indicators. Integration of information from different modalities and flexible learning are emphasised by GWT. PRM also suggests similar capacities by requiring a general-purpose belief-formation and decision-making system, which receives input from any sensory modality subject to metacognitive monitoring. Evaluation and evaluative learning are closely connected with agency. AI systems that combine agency with indicators implying integration and flexibility may be particularly good candidates for consciousness, because they will share notable cognitive capacities with the animals which are, according to UAL, most likely to be conscious. Furthermore, having an architecture of the kind described by either GWT or PRM is perhaps particularly good evidence for consciousness if it facilitates flexible learning.

However, it is possible that the capacity for unlimited associative learning could be achieved in AI systems in different ways—using architectures that are both unlike those described by theories like GWT, and unlike those belonging to animals that share this capacity. These systems might have this capacity while lacking some of the hallmarks of consciousness identified by Ginsburg and Jablonka. This would undermine the argument for consciousness in such systems, which is a reason to doubt whether UAL itself is a good indicator for consciousness in artificial systems. Another reason is that the status of the UAL hypothesis is very different from that of the other theories we have considered: rather than claiming to identify the mechanism underlying conscious experience, it claims to identify a behavioural marker for consciousness in living organisms. For these reasons, we do not include the capacity for UAL in our list of indicators.

\pdfbookmark[2]{Agency and Embodiment}{agency-embodiment}
\subsubsection{Agency and Embodiment}

Current AI systems often relate to their environments in very different ways from humans and other animals, and it can be argued that these differences are evidence against consciousness in such systems. For example, consider the well-known image classifier AlexNet (Krizhevsky et al. 2012), a relatively small and simple DNN trained by supervised learning. The beings which we usually take to be conscious are very different from AlexNet: they are agents which pursue goals and make choices; they are alive and have bodies; and they continually interact with the environment in a way that involves storing and integrating information over short periods of time. AlexNet, in contrast, has the function of classifying images but does not take actions or pursue any goal. It is physically realised in the sense that the weights and other features which define it are stored in physical memory devices, but it does not have a body. And it processes inputs that are separated in time and independent of each other and its outputs in feedforward passes that do not change its state. In this and the following section, we discuss whether further indicators of consciousness can be identified among these ``big-picture'' differences between humans and some AI systems.

As we will see in this section, scientists and philosophers have argued that various properties related to agency and embodiment are necessary for consciousness. We survey some of these arguments, identifying possible indicators, then discuss whether they can be formulated in ways that are consistent with computational functionalism. We close the subsection by adding two more indicators to our list.

\phantomsection
\pdfbookmark[3]{Agency}{agency}
\subsubsubsection{Agency}

One argument for the claim that agency is necessary for consciousness is that this is implied by many scientific theories. Most of the theories of consciousness we have discussed make some reference to agency.

PRM is particularly clear that agency is a necessary condition for consciousness. PRM claims that the subsystem that discriminates between sensory signals and noise must output to a ``general belief-formation and rational decision-making system'' (Lau \& Michel 2021). One of the foundational ideas on which the theory is built is that conscious perceptual experiences have an ``assertoric force'' which may be explained either as a certain kind of persistence of the signal as an input to this decision-making system or as a strong disposition to form corresponding beliefs. What distinguishes beliefs from other representations, on most philosophical accounts, is in part their use in decisions about how to act (Stich 1978, Dretske 1988, Schwitzgebel 2021). We have already formulated an indicator that expresses this aspect of PRM, but the point remains that agency is a prerequisite for the functions cited in this theory.

The midbrain theory explicitly requires agency, since it claims that the function of the midbrain is to integrate information for action selection, and GWT also emphasises agency, without being quite so clear that it is necessary. Dehaene and Naccache (2001) claim that representation in the global workspace is needed for information to be available for intentional action, in keeping with their presentation of GWT as a theory of access consciousness, and, therefore, of the availability of information for rational agency. The UAL hypothesis claims that ``agency and embodiment'' and a ``flexible value system'' (Birch et al. 2020) are hallmarks of consciousness.

There are also independent arguments in philosophy that agency is necessary for consciousness (Evans 1982, Hurley 1998, Clark \& Kiverstein 2008). Hurley (1998) claims that consciousness requires intentional agency, which she spells out as agency in which:

\begin{myquote}
…[the system’s] actions depend holistically on relationships between what it perceives and intends, or between what it believes and desires. Relations between stimuli and responses are not invariant but reflect the rational relations between what it perceives and intends and various possibilities of mistake or misrepresentation. (p. 137)
\end{myquote}

Hurley’s argument for this claim is related to the sensorimotor theory of consciousness and theories of embodied and enactive cognition, which we explore further below. A key idea is that consciousness requires a perspective or point of view on the environment, and embodied agency can give rise to such a perspective. Hurley also argues that part of what it is to be conscious is to have access to the contents of one’s conscious experiences and that this requires the ability to act intentionally in the light of these contents. However, other philosophers argue that it is at least conceptually possible that there could be conscious experience in entities that are entirely incapable of action (Strawson 1994, Bayne et al. 2020).

There are three possible indicators of consciousness suggested by these considerations (in addition to indicator HOT-3, arising from PRM). These are: being an agent of any kind; having flexible goals or values, as suggested by advocates of UAL; and being an intentional agent, as suggested by Hurley. The latter two are strictly stronger conditions than the first, but that does not in itself imply that any should be excluded. It could be that agents in general are significantly stronger candidates for consciousness than non-agents, and intentional agents (say) are also significantly stronger candidates than other agents, in which case agency and intentional agency would both be useful indicators.

In their classic AI textbook, Russell and Norvig write that ``an agent is anything that can be viewed as perceiving its environment through sensors and acting upon that environment through activators'' (2010, p. 34). This is a very liberal definition of agency. AlexNet meets these conditions—in fact, \textit{all} AI systems meet them—and so do many simple artifacts, such as thermostats. So it is too liberal a notion of agency for our purposes. 

A more substantive notion of agency can be defined by adding three conditions to Russell and Norvig’s account. First, it is a plausible condition for agency that the system’s outputs affect its subsequent inputs. Without this, a system can only respond to inputs individually, as opposed to interacting with an environment. AlexNet does not meet this condition because, in general, the labels it produces as output do not affect which images it is subsequently given as input. This relates to the second condition, which is that agents pursue goals. We typically think of this as involving ongoing interaction with an environment, in which outputs are selected because they will bring the system closer to the goal—that is, they will change the environment state, affecting their own input, so that the goal can be more readily achieved by future outputs.

The third condition is that the system must \textit{learn} to produce goal-conducive outputs. This point is emphasised by Dretske (1988, 1999), who argues that a system’s output is only an action if it is explained by the system’s own sensitivity to the benefit of producing that output when receiving a given input. A system’s being sensitive to benefits in this way is manifested in its learning to produce beneficial outputs. Dretske’s idea is that this distinguishes some of the behaviour of animals, which he thinks of as exhibiting agency, from the behaviour of plants, which have evolved to respond to stimuli in particular ways, rather than learning to do so, and artefacts like thermostats, which have been designed to produce particular responses. In a similar vein, Russell and Norvig say that an agent lacks autonomy if its success depends on its designer’s prior knowledge of the task (2010, p. 39).

Reinforcement learning (RL) research explicitly aims to build artificial agents which pursue goals (Sutton \& Barto 2018), and typical RL systems meet all three of these conditions. In RL, the task is to maximise cumulative reward over an episode of interaction with an environment in which the system’s outputs affect its subsequent inputs. So there is a strong case that typical RL systems meet substantive criteria for agency (Butlin 2022, 2023).\footnote{A possible exception is so-called ``bandit'' systems, which learn from reward signals in environments in which outputs do not affect subsequent inputs.} However, this is not to say that RL is necessary for agency—there are other methods by which systems can learn from feedback to more effectively pursue goals. It is also important to note that the criteria for agency suggested here are relatively minimal.

According to UAL advocates, there are two senses in which agents’ goals or values can be flexible (Bronfman et al. 2016, Birch et al. 2020). One way is that the agent can be capable of learning new goals, such as through classical conditioning, in which a novel stimulus can come to be valued through association with one that the agent already values. The other is that the agent’s goals and values can be sensitive to its changing needs, as in cases in which animals’ preferences change depending on their homeostatic condition. A requirement for flexibility of either kind would be somewhat stronger than the requirement for agency, although even very simple agents tend to be capable of either classical conditioning or habit learning, in which new actions are reinforced. There are better arguments that the second form of flexibility is connected to consciousness: it may require a centralised architecture in which multiple sources of value-relevant information are integrated, and it would also mean that the motivational significance of states of different kinds, relating to different goals or values, must be compared. One suggestion in the scientific and philosophical literature is that conscious valence—that is, the degree to which conscious experiences are pleasurable or unpleasant—could constitute a ``common currency'' making such comparisons possible (Cabanac 1992, Carruthers 2018). So it is arguably flexible responsiveness to competing goals that is most compelling.

One way to spell out the idea of intentional agency, in which action depends on rational relations between belief-like and desire-like states, is through the distinction made in animal behaviour research between ``goal-directed'' and ``habitual'' behaviour (Heyes \& Dickinson 1990). In the goal-directed case, the animal can learn independently about the value of an outcome and about actions that might lead to it. It can then combine this knowledge with information about other actions and outcomes, in instrumental reasoning, to make a choice. Action, therefore, depends holistically on the animal’s ``beliefs'' and ``desires''. Computational neuroscience interprets goal-directed action selection in animals as an implementation of model-based RL (Dolan \& Dayan 2013). On this understanding, the intentional agency which Hurley emphasises is similar to the form of agency which is required by PRM. What matters for PRM is that the agent relies on the accuracy of a set of representations that are belief-like in that they are used in instrumental reasoning for action selection. This gives reality-monitoring its function: accurate representations in perceptual systems, caused by sensory stimulation, should be used to update this set, while representations without this connection with reality should not.

\phantomsection
\pdfbookmark[3]{Embodiment}{embodiment}
\subsubsubsection{Embodiment}

On both the minimal and intentional conceptions of agency, it seems possible for a system to be an agent without being embodied. For example, consider AlphaGo, the first AI system to beat the world’s best human Go players (Silver et al. 2016). AlphaGo was an agent but lacked the features which seem to distinguish embodied systems. Embodied systems are located at particular positions in their environments, and the actions and observations available to them are constrained by their positions. They also typically have relatively complex effectors, which they must continuously control over extended periods in order to perform effective actions. Their outputs are movements, and these have some direct and systematic effects on their inputs—for example, turning one’s head has a systematic effect on visual input—as well as less direct effects. These ideas are described in a philosophical account of embodiment by Clark (2008, p. 207), who writes that:

\begin{myquote}
…the body is … the locus of willed action, the point of sensorimotor confluence, the gateway to intelligent offloading, and the stable (although not permanently fixed) platform whose features and relations can be relied upon (without being represented) in the computations underlying some intelligent performances.
\end{myquote}

The point about ``intelligent offloading'' here is a reference to the thesis of embodied and extended cognition: embodied agents can exploit properties of their bodies and environments in many ways to make the cognitive tasks they face more tractable, such as by recording information in the environment.

One aspect of embodiment which is thought to be connected with consciousness is having what Hurley (1998) calls a ``perspective''. This also requires agency. According to Hurley,

\begin{myquote}
Having a perspective means in part that what you experience and perceive depends systematically on what you do, as well as vice versa. Moreover, it involves keeping track … of the ways in which what you experience and perceive depends on what you do. (p. 140)
\end{myquote}

For embodied agents that move through their environments, sensory inputs can change either because the environment changes, or because the agent changes its position in the environment, either actively or passively. To distinguish these cases, agents must keep track of their own active movements, and learn how these affect inputs. This will allow them to predict the sensory consequences of their own actions, which will typically be tightly coupled to movements, and thus distinguish them from exogenous changes in the environment. Passive movements can also be distinguished because these will cause the kinds of changes in input which are characteristic of movement in the absence of corresponding outputs. These functions involve the agent’s implicitly distinguishing between a self, located in a moving body, and an environment in which its movement takes place. Consciousness arguably requires that the subject has a single perspective or point of view on the environment, and this account aims to explain how embodied agency gives rise to such a perspective (Hurley 1998, Merker 2005, Godfrey-Smith 2019).

Relatedly, according to the sensorimotor theory of perceptual consciousness, conscious experiences are activities of interaction with the environment which constitute exercises of implicit, practical sensorimotor knowledge (Hurley 1998, O’Regan \& Noë 2001, Noë 2004, Kiverstein 2007). A simple application of sensorimotor knowledge is moving one’s head in order to see an object from a different perspective. Like Hurley’s proposal, this theory implies that learning a model of output-input contingencies and using this model in perception is a necessary condition for consciousness. The midbrain theory also suggests that consciousness is associated with the presence of an integrated model of the embodied self in its environment, used in perception, action selection and control.

However, to capture the idea of embodiment as opposed to mere agency, we need to specify further features of the output-input model—which might also be called a ``transition model'' or a ``forward model''. AlphaGo, our example of a non-embodied agent, used a model of this form in Monte Carlo tree search, a planning algorithm that involves evaluating the expected consequences of possible actions.

One way in which output-input models in embodied systems may be distinctive is by representing the direct and systematic effects that movements have on sensory inputs. This condition is not necessarily satisfied by Go-playing systems, because each of the inputs they receive may be affected by their opponents’ moves. Furthermore, embodied systems may also be distinctive in the way they use such models. The employment of an output-input model in perception to distinguish endogenous from exogenous change is one possible example of a characteristic use because it is required when sensory input depends on the position and orientation of a moving body.

A second way of using output-input models which may be distinctive of embodied systems is in motor control. Embodied systems often have effectors with multiple degrees of freedom which must be controlled in precise and responsive ways, and it has been argued that forward models (as they are called in this literature) have several uses which are specific to this context (Miall \& Wolpert 1996, McNamee \& Wolpert 2019). In particular, forward models can help embodied agents to estimate and adjust the positions of their effectors in the course of action, by providing a representation of the expected position at each moment. Discrepancies between these expectations and either sensory feedback or internal representations of goal states could be used for online adjustment.

In addition to agency and embodiment, a further possible necessary condition for consciousness is that conscious systems must be self-producing, self-maintaining, ``autopoietic'' systems (Maturana \& Varela 1991, Thompson 2005, 2007, Seth 2021, Aru et al. 2023). That is, they must sustain their existence and organisation through their own ongoing activity. This feature is characteristic of living things, which continually repair themselves and homeostatically regulate their temperatures and the balance of chemicals present in their tissues. Self-maintaining activity usually, perhaps always, involves ``proto-cognitive'' processes of sensing and responding (Godfrey-Smith 2016). Advocates of this idea refer to concepts such as agency, selfhood and autonomy in arguing that self-maintenance is necessary for consciousness. For instance, Thompson (2005) writes:

\begin{myquote}
This self-producing organization defines the system’s identity and determines a perspective or point of view in relation to the environment. Systems organized in this way enact or bring forth what counts as information for them; they are not transducers or functions for converting input instructions into output products. For these reasons, it is legitimate to invoke the concepts of selfhood and agency to describe them. (p. 418)
\end{myquote}

Those who are also sympathetic to predictive processing, such as Seth (2021), think of self-maintenance in living tissues as a similar process to prediction error minimisation in the brain—they are both instances of free energy minimisation, or ``self-evidencing'' (Hohwy 2022). More prosaically, engaging in self-maintenance gives an extra reason for systems to model their own states, is related to having flexible goals, and arguably adds a dimension of valence to representations of the self and environment. On this last point, however, it is not clear why agency in the service of self-maintenance should be distinguished from agency directed at ``external'' goals.

Building on this line of thought, Godfrey-Smith (2016) argues that self-maintaining activity is only at all readily possible by virtue of the way that molecules behave at the nanometre scale when immersed in water. There is continual random activity at this scale, in this context, which can be marshalled to support metabolic processes. On this basis, Godfrey-Smith suggests that artificial systems can only bear coarse-grained functional similarities to living organisms and that these will not be sufficient for consciousness. So another proposed necessary condition is that a conscious system must undergo metabolic processes realised at the nanoscale. Man and Damasio (2019) also suggest that self-maintenance, and thus consciousness, may depend on systems’ specific material composition.

\phantomsection
\pdfbookmark[3]{Agency and embodiment indicators}{ae-indicators}
\subsubsubsection{Agency and embodiment indicators}

The proposal that consciousness depends on material composition is clearly incompatible with computational functionalism, so we can set it aside. However, the compatibility of the other proposals we have discussed with computational functionalism is a more complicated matter. The issue is the same in each case: natural formulations of the putative indicators make reference to conditions external to the system. This is incompatible with computational functionalism if a similar system could perform the same computations in different external conditions.

For example, consider the claim that self-maintenance is necessary for consciousness, and suppose that for a system to be self-maintaining in the relevant sense, it must persist in part because it keeps track of certain inputs or internal states and acts to regulate them. It seems that for any system that does this, there could be another that performs the same computations—keeping track of the same inputs or internal states and using the same outputs to regulate them—but which does not persist for this reason. Perhaps the system works to keep an ``energy measure'' from dropping too low, but in fact, it would persist even if this measure fell to zero, and its energy is supplied by an external operator in a way that does not depend on its behaviour.

If agency or embodiment are necessary for consciousness, these conditions may be incompatible with computational functionalism for similar reasons. These conditions require that the system’s inputs are sensitive to its outputs. But in principle, it is possible for a system to exist in an environment in which its inputs do not depend on its outputs, and yet receive, by chance, patterns of inputs and outputs which are consistent with this dependency. Such a system might perform the same computations as one that genuinely interacted with its environment.

To avoid this incompatibility, indicators concerning agency, perspective, or self-maintenance should be formulated ``narrowly'', in ways that do not make reference to external conditions. For instance, rather than saying that a system is more likely to be conscious if it pursues goals by interacting with its environment, we can say that it is more likely to be conscious if it learns from feedback and selects actions \textit{in such a way as to} pursue goals by interacting with its environment. Similarly, we can say that a system is embodied if it has a model of how its outputs affect its inputs, which represents some systematic effects, and uses this model in perception or control. There is a sense in which this system might misrepresent itself as an embodied agent, even if it has learnt this model, because it may be that the apparently systematic contingencies suggested by its past observations are mere coincidences. However, only embodiment conditions on which this is enough for consciousness are consistent with computational functionalism. Notably, this account of embodiment allows that systems controlling virtual avatars can count as embodied.

In this section, we have seen several indicators which could be added to our list. These include: being an agent; having flexible goals or values; being an intentional agent; having a perspective; having a body; and being self-maintaining. However, there are also reasons to exclude some of these. We already have an indicator requiring a relatively sophisticated form of agency with belief-like representations: indicator HOT-3, derived from PRM. An intentional agency indicator would be too similar to this one. And although a narrow formulation of a self-maintenance indicator is possible, which would be compatible with computational functionalism, this would be contrary to the spirit of the philosophical theories which emphasise self-maintenance. The ideas of being an agent and having flexible goals are closely related, and the main way in which flexible goals add to the case for consciousness is through the argument for centralisation and a common motivational currency, so we combine these in one indicator.

We, therefore, adopt the following two further indicators of consciousness:

\myindicator{AE-1 Agency: Learning from feedback and selecting outputs so as to pursue goals, especially where this involves flexible responsiveness to competing goals}

\vspace{3pt}

 \myindicator{AE-2 Embodiment: Modeling output-input contingencies, including some systematic effects, and using this model in perception or control}
 
\pdfbookmark[2]{Time and Recurrence}{time-recurrence}
\subsubsection{Time and Recurrence}

Human conscious experience seems to be highly integrated over time. We seem to undergo experiences that are themselves extended in time, and constitute conscious perception of temporally-extended phenomena, such as when we hear a bird’s song or watch it fly from one perch to another. What’s more, from the time we wake up each day we seem to experience a flow of several hours of continuous, integrated experience. Whether or not this feature of our experience should be explained in terms of memory (Dainton 2000, Phillips 2018b), our conscious experiences also seem to be deeply influenced by our memories. This integration might be seen as a reason to doubt that consciousness is possible in ANNs whose activity consists of temporally discrete forward passes; the point is particularly vivid if we imagine that there are long intervening periods (weeks, years) between passes. However, several responses can be made to this objection.

One response is that it is not obvious that consciousness is necessarily integrated over time. Patients with dense amnesia seem to have a series of brief, disjointed experiences (Wilson et al. 1995). These experiences may not be static, but it seems possible to imagine a conscious being that had only a succession of brief, static, discrete experiences. There is a question about what would make it the case that these were the experiences of a single subject, but it is not obvious either that integration over time is the only possible grounds for the persistence of the subject of conscious experience, or that there could not be conscious systems without a persisting subject.

Furthermore, it is possible that the apparent temporal continuity of human experience is illusory. Philosophers and scientists defend a range of views about how our experiences are generated over time (Dainton 2023), but these include the view that we undergo discrete, static experiences in rapid succession, generated by sampling from unconscious activity (VanRullen 2016, Herzog et al. 2020). Prosser (2016) calls this the ``dynamic snapshot'' view. The content of discrete experiences may be informed by unconscious retention of information, and be capable of representing change despite being static. For example, when watching a bird I might undergo a brief static visual experience representing not only that the bird is in a certain location, but that it is moving in a certain way. On this model, there may be no special integration between successive experiences, since the appearance of such integration can be explained by their being sampled at a fairly high frequency from smoothly-changing unconscious perceptual activity which in turn reflects smooth continuous change in the environment.

Despite these points, it does seem that algorithmic recurrence (i.e. recurrence as it is usually understood in machine learning; see section 2.1.3) is likely to be necessary for conscious experience with a human-like temporal character. For conscious experience to represent change or continuity in the environment in any way, information about the past must be retained and used to influence present processing. This is a further reason to take algorithmic recurrence to be an indicator of consciousness, as expressed in our indicator RPT-1. Integration over time, and, therefore, recurrent processing, is also emphasised by the midbrain and sensorimotor theories, the UAL framework and some PP theorists.

\pdfbookmark[1]{Indicators of Consciousness}{consciousness-indicators}
\subsection{Indicators of Consciousness}

Each of the theories and proposals which we have discussed in sections 2.1-2.4 is of some value in assessing whether a given AI system is likely to be conscious, or how likely it is that a conscious system could be built in the near future. They are each supported by evidence and arguments that have some force. We, the authors of this report, have varying opinions on the strength of the evidence and arguments supporting each theory, as well as varying background views about consciousness that influence our assessments of the likelihood of near-term AI consciousness. Here we summarise the findings of section 2 by giving a list of indicators of consciousness drawn from the theories and proposals we have discussed.

Our claim about these indicators is that they jointly amount to a rubric, informed by current views in the science of consciousness, for assessing the likelihood of consciousness in particular AI systems. Systems that have more of these features are better candidates for consciousness. Theories of consciousness make stronger claims than this, such as that some of these are necessary conditions for consciousness, and that combinations are jointly sufficient. We do not endorse these stronger claims, but we do claim that in using these indicators one should bear in mind how they relate to theories and to each other—some combinations of indicators will amount to more a compelling case for consciousness than others. The extent to which these indicators are individually probability-raising also varies, with some being plausibly necessary conditions that do not make consciousness significantly more likely in isolation (perhaps including RPT-1, GWT-1 and HOT-1).

\newpage




\begin{longtable}{|>{\raggedright\arraybackslash}p{0.53\textwidth}|p{0.43\textwidth}|}
    \arrayrulecolor{lightblack} \hline
    \rowcolor{darkest} \multicolumn{2}{|c|}{\textcolor{white}{\textbf{Recurrent processing theory}}} \\
    \arrayrulecolor{lightblack} \hline
    \rowcolor{medium} \indicator{RPT-1} Input modules using algorithmic recurrence & \cellcolor{white} \\
    \arrayrulecolor{white} \hline
    \rowcolor{lightest} \indicator{RPT-2} Input modules generating organised, integrated perceptual representations & \whitemultirow{-1.875}{0.45\textwidth}{\textit{RPT-1 and RPT-2 are largely independent indicators. RPT-1 is also supported by temporal integration arguments. }} \\
    \arrayrulecolor{lightblack} \hline
    \rowcolor{darkest} \multicolumn{2}{|c|}{\textcolor{white}{\textbf{Global workspace theory}}} \\
    \arrayrulecolor{lightblack} \hline
    \rowcolor{medium} \indicator{GWT-1} Multiple specialised systems capable of operating in parallel (modules) & \cellcolor{white} \\
    \arrayrulecolor{white} \hline
    \rowcolor{lightest} \indicator{GWT-2} Limited capacity workspace, entailing a bottleneck in information flow and a selective attention mechanism & \cellcolor{white} \\
    \arrayrulecolor{white} \hline
    \rowcolor{medium} \indicator{GWT-3} Global broadcast: availability of information in the workspace to all modules & \cellcolor{white} \\
    \arrayrulecolor{white} \hline
    \rowcolor{lightest} \indicator{GWT-4} State-dependent attention, giving rise to the capacity to use the workspace to query modules in succession to perform complex tasks & \whitemultirow{-5.5}{0.45\textwidth}{\textit{GWT claims that these are necessary and jointly sufficient. GWT-1 through GWT-4 build on one another. GWT-3 and GWT-4 entail RPT-1.}} \\
    \arrayrulecolor{lightblack} \hline
    \rowcolor{darkest} \multicolumn{2}{|c|}{\textcolor{white}{\textbf{Computational higher-order theories}}} \\
    \arrayrulecolor{lightblack} \hline
    \rowcolor{medium} \indicator{HOT-1} Generative, top-down or noisy perception modules & \cellcolor{white} \\
    \arrayrulecolor{white} \hline
    \rowcolor{lightest} \indicator{HOT-2} Metacognitive monitoring distinguishing reliable perceptual representations from noise & \cellcolor{white} \\
    \arrayrulecolor{white} \hline
    \rowcolor{medium} \indicator{HOT-3} Agency guided by a general belief-formation and action selection system, and a strong disposition to update beliefs in accordance with the outputs of metacognitive monitoring & \cellcolor{white} \\
    \arrayrulecolor{white} \hline
    \rowcolor{lightest} \indicator{HOT-4} Sparse and smooth coding generating a ``quality space" & \whitemultirow{-8}{0.45\textwidth}{\textit{PRM claims that these are necessary and jointly sufficient. HOT-1 through HOT-3 build on one another; HOT-4 is independent. The first clause of HOT-3 is also supported by arguments concerning intentional/flexible agency, and entails AE-1.}} \\
    \arrayrulecolor{lightblack} \hline
    \rowcolor{darkest} \multicolumn{2}{|c|}{\textcolor{white}{\textbf{Attention schema theory}}} \\
    \arrayrulecolor{lightblack} \hline
    \rowcolor{medium} \indicator{AST-1} A predictive model representing and enabling control over the current state of attention & \cellcolor{white} \\
    \arrayrulecolor{lightblack} \hline
    \rowcolor{darkest} \multicolumn{2}{|c|}{\textcolor{white}{\textbf{Predictive processing}}} \\
    \arrayrulecolor{lightblack} \hline
    \rowcolor{medium} \indicator{PP-1} Input modules using predictive coding & \whitemultirow{-1}{0.5\textwidth}{\textit{Entails RPT-1 and HOT-1.}} \\
    \arrayrulecolor{lightblack} \hline
    \rowcolor{darkest} \multicolumn{2}{|c|}{\textcolor{white}{\textbf{Agency and embodiment}}} \\
    \arrayrulecolor{lightblack} \hline
    \rowcolor{medium} \indicator{AE-1} Agency: Learning from feedback and selecting outputs so as to pursue goals, especially where this involves flexible responsiveness to competing goals & \cellcolor{white} \\
    \arrayrulecolor{white} \hline
    \rowcolor{lightest} \indicator{AE-2} Embodiment: Modeling output-input contingencies, including some systematic effects, and using this model in perception or control & \whitemultirow{-4}{0.45\textwidth}{\textit{Both indicators are also supported by midbrain and UAL theories, and to some extent by GWT, PRM and PP, especially AE-1. Systems meeting AE-2 are likely, but not guaranteed, to also meet AE-1.}} \\
    \arrayrulecolor{lightblack} \hline
    \caption{Indicator Property Entailments}
\end{longtable}

\newpage

\pdfbookmark[0]{Consciousness in AI}{ai-consciousness}
\section{Consciousness in AI}

What do the findings of section 2 imply about consciousness in current and near-future AI systems? In this section, we address this question in two ways. First, in section 3.1 we discuss the indicator properties in turn, asking how they could be implemented in artificial systems. Second, in section 3.2 we examine several existing AI systems as case studies illustrating both how the indicators should be used, and how our method evaluates some current systems. We discuss large language models and the Perceiver architecture (Jaegle et al. 2021a, b) with a particular focus on global workspace theory, and consider whether any of a selection of recent systems—PaLM-E (Driess et al. 2023), a ``virtual rodent'' (Merel et al. 2019) and AdA (DeepMind Adaptive Agents Team 2023)—are embodied agents.

Reflecting on how to construct systems with the indicator properties, and on whether they are present in current systems, illustrates some crucial lessons of our work. One is that assessing whether a system possesses an indicator property typically involves some interpretation of the description of the property; the descriptions of indicator properties, like the theories from which they are drawn, contain ambiguities that possible implementations draw out. To describe indicator properties with so much precision that this kind of interpretation is not needed would mean going well beyond the claims made by scientific theories of consciousness, and these more precise claims would also tend to be less well-supported by the available empirical evidence. We hope that interdisciplinary research on consciousness, which brings together neuroscientists and AI researchers, will result in greater precision in theories of consciousness and the development of empirical methods that can provide evidence supporting more precise theories.

A second lesson from the discussion in this section is that in assessing AI systems for consciousness, it is not always sufficient to consider the system’s architecture, training and behaviour. For example, we may know that a system is a recurrent neural network and has been trained via RL to successfully control a virtual body to perform a task, and yet not know whether its performance relies on a learned model of output-input contingencies. In such a case, we may not know whether the system satisfies our embodiment condition. Interpretability methods, such as examining the information encoded in hidden layers (Olah et al. 2018), would be required to determine whether the system has acquired such a model in the course of training.

The third lesson is that, despite the challenges involved in applying theories of consciousness to AI, there is a strong case that most or all of the conditions for consciousness suggested by current computational theories can be met using existing techniques in AI. This is not to say that current AI systems are likely to be conscious—there is also the issue of whether they combine existing techniques in the right ways, and in any case, there is uncertainty about both computational functionalism and current theories—but it does suggest that conscious AI is not merely a remote possibility in the distant future. If it is possible at all to build conscious AI systems without radically new hardware, it may well be possible now.

\pdfbookmark[1]{Implementing Indicator Properties in AI}{implementing-indicators-ai}
\subsection{Implementing Indicator Properties in AI}

\pdfbookmark[2]{Implementing RPT and PP}{imp-rpt-pp}
\subsubsection{Implementing RPT and PP}

We begin our investigation of the indicators by discussing the two indicators taken from RPT, algorithmic recurrence and perceptual organisation, along with the PP indicator (i.e., indicators RPT-1, RPT-2 and PP-1). The reason for including PP here is that, as we will see, recent studies have found that using predictive coding in computer vision can facilitate processing that is more sensitive to global features of visual scenes, in contrast to the local-feature sensitivity of feedforward convolutional neural networks which perform well in classification tasks.

Algorithmic recurrence (RPT-1) is a feature of many deep learning architectures, including recurrent neural networks (RNNs), long short-term memory networks (LSTMs) and gated recurrent unit networks (GRUs) (LeCun et al. 2015). Building systems that possess indicator property RPT-1 is, therefore, straightforward. Although they are less widely used, there are also methods for implementing predictive coding (which is a form of algorithmic recurrence) in artificial systems (Lotter et al. 2017, Oord et al. 2019, Millidge et al. 2022). These systems meet indicator PP-1. Furthermore, recurrent neural networks trained on prediction tasks and optimised for energy efficiency self-organise into distinct populations of ``prediction'' and ``error'' units (Ali et al. 2022).

Turning to perceptual organisation (RPT-2), artificial vision models such as deep convolutional neural networks (DCNNs), are already both highly successful and often claimed to be good models of human vision (Kietzmann et al. 2019, Lindsay 2021, Mehrer et al. 2021, Zhuang et al. 2021). However, the claim that current systems are good models of human vision has recently been criticised (Bowers et al. 2022, Quilty-Dunn et al. 2022), and, more to the point, the human-level performance in visual object recognition that DCNNs achieve does not entail that they represent organised visual scenes. Bowers et al. (2022) and Quilty-Dunn et al. (2022) both cite evidence that DCNNs trained to classify objects are more sensitive to local shapes and textures than to global shapes, and tend to ignore relations between parts of objects, suggesting that they do not employ representations of integrated scenes. Conwell and Ullman (2022) found that the recent image generation model DALL-E 2 performed poorly when prompted to generate a scene with objects arranged in unfamiliar ways.

From the point of view of RPT, these points might be taken to show that the models in question are capable of categorising features of visual stimuli, a function which is said to be performed unconsciously in humans, but are not capable of further functions up to and including the generation of organised, integrated representations of visual scenes, some of which may require consciousness (Lamme 2020). However, other current systems, including predictive coding networks, do perform some of these further functions.

\vspace{9cm} 

\begin{wrapfigure}{r}{0.5\textwidth}
\centering
\includegraphics[width=0.4\textwidth,keepaspectratio]{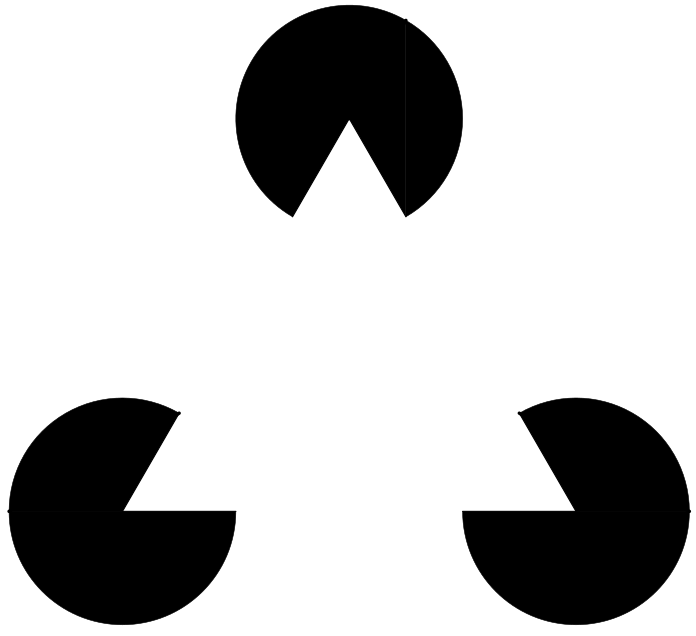} 
\caption{\textbf{Illustration of the Kanizsa illusion}. 2020. Wikimedia Commons. Reprinted with permission.}
\end{wrapfigure}

PredNet, a predictive coding network trained to predict the next frame of video inputs, is notable partly because success at this task seems to require representation of the objects that make up a scene and their spatial relations (Lotter et al. 2017). Furthermore, there is evidence that PredNet units respond to illusory contours in the Kanizsa illusion, the perception of which depends on inferring the presence of objects from the wider context (Lotter et al. 2020). Extending this finding, Pang et al. (2021) used a technique for adding feedback predictive coding connections to a feedforward DCNN and found further evidence that this predictive coding network, but not the initial feedforward model, was sensitive to the Kanizsa illusion. However, Lamme (2020) claims that the Kanizsa illusion requires only perceptual ``inference'', not perceptual organisation as understood by RPT.

Beyond predictive coding, systems of other kinds have been developed specifically to represent objects and their relations in visual scenes. MONet, which uses a recurrent attention mechanism and a variational autoencoder to decompose and reconstruct scenes, is one example (Burgess et al. 2019). In this system, the attention network picks out individual objects in turn for identification and reconstruction by the variational autoencoder. Another system, the Object Scene Representation Transformer, is trained to predict the appearance of complex scenes from new angles, a task which (like video prediction) requires the representation of scenes as made up of objects in space (Sajjadi et al. 2022). Representing organised perceptual scenes is an active area of research in machine learning for which several methods have already been developed (Greff et al. 2020).

\pdfbookmark[2]{Implementing GWT}{imp-gwt}
\subsubsection{Implementing GWT}

Implementing GWT in artificial systems has been the subject of several studies, including recent research by VanRullen and Kanai (2021) and Goyal et al. (2022). Here we give brief overviews of these two studies before discussing indicators GWT-1 through GWT-4 in turn.

VanRullen and Kanai’s (2021) proposal involves a set of specialised neural modules, at least some of which are generative networks that can produce behavioural outputs or drive sensory processing top-down. Each of these modules has its own low-dimensional latent space in which key information is represented. The modules are integrated by a workspace, which is a shared latent space trained to perform unsupervised translation of representations in the latent spaces of the modules so that the information they carry is available to the others. The workspace has a lower capacity than the sum of the module latent spaces, and a task-dependent key-query attention mechanism is used to determine which information it broadcasts. So this architecture includes the important features of a bottleneck, global broadcast and state-dependent selection. However, this work is a ``roadmap'' to a possible implementation, rather than a working system. It faces a substantial open question about how the attention mechanism could be trained to select among the potential inputs to the workspace, and especially how this could achieve the sequences of operations of attention needed to control extended, functional sequences of operations by relevant modules.

Meanwhile, Goyal et al. (2022) experimented with a method for implementing global workspace which similarly involved using key-query attention to select which of multiple modules would write to a shared space, with the contents of the shared space then being broadcast to all modules. In this case, the modules were trained together to generate mutually usable representations, avoiding the need for translation in the workspace. A limitation of the specific implementations developed in this work was that the ``modules'' were elements processing tokens in a sequence or parts of an image, so it is questionable whether they were specialised subsystems capable of operating in parallel; instead, they each contributed in similar ways to the performance of a single task.

Although neither of these studies produced a working system that clearly satisfies all four GWT indicators, this does seem to be a realistic objective, as we will argue by considering the indicators in turn. Indicator GWT-1 states that the system must have specialised systems, or modules, capable of working in parallel. To make global broadcast possible, these modules must be implemented by recurrent neural networks, unless they are ``output'' modules for the system as a whole, which do not provide information to the workspace. The modules might take as input:

\begin{adjustwidth}{0.5cm}{0.5cm}

\begin{enumerate}
    \item Sensory input in one or more modalities.
    \item Input from a small number of other modules that typically work in tandem. For instance, a ``saccade'' module might take input from a ``visual saliency'' module in order to enable quick bottom-up saccades towards potentially important objects.
    \item Top-down signals coming from an executive Global Workspace module.
\end{enumerate}
\end{adjustwidth}

These modules might be trained independently on narrow tasks, as suggested by VanRullen and Kanai (2021). Or they might be jointly trained end-to-end with the workspace in order to achieve some system-wide objective, from which module specialisation to subtasks would naturally emerge. The end-to-end training approach was employed by Goyal et al. (2022), although for relatively simple tasks (see also Goyal et al. 2020).

The second element of an implementation of GWT is a limited-capacity workspace, which is a further neural module with different properties. The simplest way to limit the capacity of the workspace is to limit the number of dimensions of its activity space. Another interesting option is to train a recurrent neural network that exhibits attractor dynamics. An attractor is a state in a dynamical system such that when that state is reached, it will remain stable in the absence of inputs or noise to the system. The reason that attractor dynamics limits capacity is that it induces a many-to-one mapping from initial conditions in a neural trajectory to attractors (any neural trajectory that enters an attractor’s basin of attraction will converge to that attractor). Thus, these attractor dynamics induce an information bottleneck by contracting the size of the space of stable states. Ji, Elmoznino et al. (2023) argue that attractor dynamics in the workspace can help to explain the apparent richness and ineffability of conscious experience.

For indicator GWT-3, global broadcast, the basic requirement is that all modules take workspace representations as input. As we have seen, this means that some mechanism must be in place to ensure that these inputs can be used by all modules, such as the translation mechanism proposed by VanRullen and Kanai (2021). In the global neuronal workspace theory developed by Dehaene and colleagues (Dehaene \& Naccache 2001, Dehaene \& Changeaux 2011, Mashour et al. 2020), the workspace exhibits particular dynamic properties: for a representation to be in the workspace and globally broadcast, it must be sustained by recurrent loops. While it is not clear that this is essential, this behaviour could be replicated in AI if a network exhibiting attractor dynamics was used to implement the workspace. In this case, the broadcast mechanism might consist of a leaky neural integrator whose dynamics have slow timescales such that sustained inputs are required to place it in a particular state, and in the absence of these sustained inputs it relaxes back to some baseline state (as in models of decision-making through evidence accumulation). This broadcast mechanism would generate the top-down signals feeding into each specialised module.

Indicator GWT-4 includes the conditions that the system must use a state-dependent attention mechanism and that the workspace must be able to compose modules to perform complex tasks by querying them in succession. For the state-dependent attention mechanism, both VanRullen and Kanai (2021) and Goyal et al. (2022) propose the use of key-query attention, which is common in current AI models. A query can be computed from the workspace’s current state, and keys can be computed for all other modules. The similarity between the workspace’s query and a given module’s key would be normalised by the similarities across all other modules in order to introduce competition between modules, and these normalised similarities would determine the degree to which each module’s value contributes to the net input to the workspace. That is, a standard key-query attention mechanism would be applied at each timepoint to compute the input to the workspace in a way that depends on its current state.

The model described here would be able to meet the second part of GWT-4—the capacity to use the workspace to query modules in succession to perform complex tasks— when it is unrolled through time because there are computational loops between the workspace and the modules. The modules receive input from bottom-up sensory input and from a small number of other modules, but they also receive top-down input from the workspace. This means that, for instance, it is possible for one module to control others by controlling what is represented in the workspace. The sequential recruitment of modules by the workspace is within the computational repertoire of the system, so it could emerge if it is beneficial during training. However, suitable training would be required for such a system to learn to compose modules in useful ways and to perform complex tasks, and constructing a suitable training regime may be a significant challenge for implementing GWT.

\pdfbookmark[2]{Implementing PRM}{imp-prm}
\subsubsection{Implementing PRM}

We now consider how perceptual reality monitoring theory, as a representative computational HOT, could be implemented in AI. That is, we consider how an AI system could be constructed with indicator properties HOT-1 through HOT-4. Although PRM researchers claim that there are no current AI systems that meet all of the requirements (Dehaene et al. 2017, Michel \& Lau 2021, Lau 2022), and there have been no implementation attempts that we are aware of, standard machine learning methods are sufficient for plausible implementations of most elements of the theory. We begin by considering the implementation of a ``quality space''—that is, a space of possible representations which satisfies the PRM account of the basis of phenomenal qualities—before turning to the core claim of PRM, that first-order representations become conscious as a result of being represented as ``real'' by a higher-order monitoring mechanism.

Indicator HOT-4 states that the system must have sparse and smooth coding generating a ``quality space''. One of the most important features of all DNNs is that each layer endows the model with a smooth representation space, in which a continuous range of possible activations codes for the range of possible inputs. The smoothness of these representation spaces is thought to be one of the primary reasons why they can generalise to novel inputs at test time; even if a novel input drives a layer into a different activation pattern, this pattern can still be interpreted by subsequent layers so long as it is within the model’s training distribution (that is, so long as the layer activated with similar patterns at training time) (Belkin et al. 2018, Bartlett et al. 2021). In fact, there is evidence that the perceptual representation spaces learned by current DNNs already closely resemble those of the human visual system, meaning their corresponding ``quality spaces'' might already be substantially aligned. For instance, Kriegeskorte (2015) finds that the matrix of pairwise dissimilarities between representations of many visual stimuli in a DNN is closely aligned to what is observed in neural activity and that there is even a correspondence between successive layers in the network and the hierarchical organisation of the visual cortex—although more recent studies have complicated this picture (see e.g. Golan et al. 2020, Lindsay 2021, Bowers et al. 2022).

Standard methods in machine learning can be used to satisfy the condition that representations be sparse. For example, regularization techniques can enforce sparse representations in DNN models, by minimising the magnitude of the network’s representations or the mutual information between inputs and representations (Tishby 2000). Layers that use normalisation functions like SoftMax, in which higher values suppress lower values in the representation, also increase sparsity.

Indicators HOT-1 and HOT-2 state that the model must contain both first-order perceptual representations of sensory data and higher-order representations that assign a measure of reliability or ``realness'' to particular first-order representations. To meet these conditions, a vast number of known deep learning solutions are possible, of which we review a few. Importantly, all of these solutions share a simple architectural design of two components: (1) a (first-order) neural network takes sensory data and/or top-down signals as input and produces a number of perceptual representations distributed across a hierarchy of layers (2) in parallel, a series of separate (higher-order—specifically second-order) neural networks each take a first-order layer’s activations as input and then output a single scalar, representing the probability that the first-order representation of that layer is veridical. Solutions for meeting the conditions will differ primarily in terms of how the second-order networks are trained.

If supervision signals are occasionally present that provide the second-order networks with ``ground-truth'' about the reliability of first-order representations, then the second-order network can be trained to estimate the probability of correctness by standard supervised learning. Obtaining this ground-truth signal may be difficult, but not impossible. For instance, if a source of first-order representation errors is internal noise in the network, ground-truth can be estimated simply by averaging noisy first-order representations over time. Another possibility is to obtain ground-truth by comparing representations of the same percept in different sensory modalities (e.g., verifying the veracity of a sound using visual feedback) or through movement (e.g., checking if a visual percept behaves as it should given known motor actions).

If ground-truth is not directly available, the second-order networks can be trained on other surrogate tasks where the reliability of a signal is an implicit factor in performance. For instance, second-order networks might try to predict upcoming first-order representations using past ones. Because externally generated signals are sometimes more predictable than internally generated signals—a perception of a falling ball is more predictable than a hallucination generated by random internal noise—when the networks’ predictions have high error, the second-order network can assign a lower probability to the veracity of the first-order representation. Alternatively, when imagery is under cognitive control, engaging in imagination can lead to more predictable sensory consequences (I imagine a pink bear and a pink bear appears) than when it is not (I allow my mind to wander). Thus the level of effortful control can serve as another internal signal that could train a reality monitoring system (Dijkstra et al. 2022). In these ways, a second-order network could learn to use predictability as a cue to ``realness'' even in the absence of supervision signals. These methods are closely related to predictive coding, which has already seen applications in modern deep learning (e.g., Millidge et al. 2022, Oord et al. 2019, Alamia et al. 2023), except that here the prediction is done across time rather than across layers in a hierarchy.

Other methods of training the second-order network involve thinking of it as a world-model. For instance, Bayesian methods in deep learning view perception as an inference process, in which a neural network attempts to infer the values of latent variables that might have generated the data. For ideal Bayesian inference, these latent variables must be sampled according to their posterior probability, which is proportional to their prior probability multiplied by their likelihood of having generated the data under some world model that specifies how latent variables produce sensory observations. This sort of perspective fits well with PRM; the inference machinery for latent variable values can be seen as producing perceptual first-order representations, while the second-order networks evaluate the probability that the first-order representations are true.

While Bayesian inference is intractable in general, several approximate methods exist. For instance, a recent class of models called Generative Flow Networks (GFlowNets) provide a framework for doing approximate Bayesian inference using modern deep neural networks (Bengio et al. 2021, Bengio et al. 2022), and can even be used to jointly train the inference model (first-order network) at the same time as the world-model (second-order network) (Hu et al. 2023, Zhang et al. 2023). This proposed GFlowNet architecture (like other approximate Bayesian inference methods) is again related to predictive processing theories of consciousness, which state that conscious percepts are the brain’s best guesses of the latent causes of its own sensory input (Seth 2021). An important difference in the PRM case, though, is that after the first-order inference system samples perceptual latent variables, they only become conscious if the second-order world model assigns them a high enough probability of being veridical, where either a hard or a soft threshold can be used to produce the ``real'' tag.

Note that the mechanism described above could also be used to produce internally-generated conscious experiences in the system. For instance, a top-down imagination module might first produce a goal (e.g., visualize an apple) and then drive neural activity accordingly in the first-order network’s sensory and perceptual layers. Following this, the second-order world model would assign a probability of veracity to this activity. If the imagination module is capable of producing coherent enough representations in the sensory and perceptual layers, then the second-order model would assign the signals a ``real'' tag in the same way as for externally-generated signals, and the imagined content would be consciously experienced. A similar mechanism may exist in the brain, as it has been shown that neural activity in sensory regions is similar during imagination and bottom-up perception and that the degree of similarity correlates with the vividness of the experience (Dijkstra 2019, Dijkstra 2023).

A final possibility is to learn a world model implicitly through adversarial methods. In particular, Generative Adversarial Networks (GANs) (Goodfellow et al. 2014) are a class of methods in which a generator network attempts to sample synthetic data and a discriminator attempts to differentiate real data from synthetic data produced by the generator. The generator is optimized to fool the discriminator into predicting its outputs as ``real'', with the result that, if the networks used are powerful enough, the stable solution is for the generator to sample data from the true data distribution. The discriminator in GANs is comparable to the higher-order monitoring mechanism posited by PRM (Lau 2022). A GAN-based implementation of PRM might involve (a) a first-order perception network that produces perceptual representations given sensory data, (b) a generator network that inverts this process by first sampling synthetic perceptual representations and then producing synthetic sensory data given those percepts, and (c) a second-order network that is trained to discriminate between the real sensory-percept tuples produced by the first-order network and the synthetic ones produced by the generator (Gershman 2019). Like in the case of explicit world models, we can subsequently produce the ``real'' tag of PRM by applying a hard or a soft threshold to the discriminator’s output given representations from the first-order network.

It now remains for us to consider indicator HOT-3, which concerns the consumer of outputs from the metacognitive monitoring mechanism. Indicator HOT-3 states that this should be a ``general belief-formation and action selection system'' and that there should be a ``strong disposition'' to update it in accordance with the monitoring mechanism’s outputs. There are dual motivations for this condition: the function of the monitoring mechanism is to serve this consumer, which relies on the accuracy of first-order representations which are tagged as ``real''; and the ``force'' of these inputs to the consumer is intended to explain the force and persistence of conscious experiences.

To implement the consumer, we can instantiate one or more higher-level networks that similarly take only perceptual representations with the ``real'' tag as input. There might be many ways to accomplish this, but one possibility is to use standard Transformer architectures for the higher-level networks with one small adaptation. In the Transformer, one or more query vectors would be produced by the higher-level network (intuitively representing the current computational goal) that would then attend to first-order perceptual representations depending on the values of their key vectors (intuitively representing the ``kind'' of percept or its ``data type''). In a standard Transformer, each precept would then be modulated based on the similarity between its key and the higher-level network’s query, and then subsequently integrated into the downstream computation. If we wish to additionally make use of the ``real'' tags produced by the second-order networks, we need only multiply the query-key similarities by the value of the ``real'' tags for each corresponding percept, which would allow them to effectively serve as prior masks on which precepts may influence computations in the Transformer. Importantly, such an architecture is compatible with predictions made by PRM about the stubbornness of conscious percepts (Lau 2019): even if we are \textit{cognitively} aware that a percept is inaccurate (e.g., a phantom pain, or a drug-induced hallucination), we still have a strong disposition to integrate it into our higher-level reasoning. With this Transformer-based architecture, conscious experiences cannot be reasoned away. This is because the higher-level network simply takes the percepts modulated by their ``real'' tags as input, and has no ability to change those ``real'' tags itself.

This proposal focuses on the second aspect of indicator HOT-3 (the ``strong disposition''), rather than the idea that the consumer should be a general belief-formation and action selection system. Philosophers have various ideas about the features a system would need to have to count as forming beliefs, but one plausible necessary condition, which may be sufficient in this context, is that the system engages in instrumental reasoning leading to action selection (see section 2.4.5).

\pdfbookmark[2]{Implementing AST}{imp-ast}
\subsubsection{Implementing AST}

Two notable studies have implemented simple attention schemas in artificial systems. Wilterson and Graziano’s (2021) system used reinforcement learning on a three-layer neural network with 200 neurons per layer to learn to catch a ball falling on an unpredictable path. The primary input to this system was a noisy ``visual'' array showing the position of the ball. This array included an ``attention spotlight''—a portion of the array with the noise removed—that the system could learn to move. A natural strategy for solving the task was thus for the system to learn to position the spotlight over the ball, so that noise in the visual array would not interfere with its ability to track and catch it. This was facilitated by a second input, an ``attention schema'' that represented the current position of the spotlight in the array. Wilterson and Graziano found that this system was much more successful in learning to perform the task when the attention schema was available, even though the spotlight remained when the schema was removed.

This very simple system did possess some part of indicator property AST-1, the attention schema, because it used a representation of an attention-like mechanism to control that mechanism, resulting in improved performance. However, this was not a predictive model, and the attention spotlight was substantially different from attention itself because it was a simple controllable feature of the input rather than an internal process with multiple degrees of freedom governing information flow through the system.

More recently, Liu et al. (2023) tested several different systems employing key-query-value attention on multi-agent reinforcement learning tasks. Their systems involved three main elements: multi-head attention layers, a recurrent neural network for ``internal control'' and a policy network. In the version that they took to most accurately implement an attention schema, the attention layers were applied to inputs to the system and sent information on to the policy network that generated actions, with the internal control network both learning to predict the behaviour of the attention layers and influencing this behaviour. This system performed better than the others, with different architectures made up of the same components, which were tested on the multi-agent RL tasks.

Compared to Wilterson and Graziano’s system, this system had the advantage of using a learnt predictive model of attention, rather than having an infallible representation provided as input. It was still used only for relatively simple tasks in 2D visual environments, and a caveat to this line of research is that attention in AI is not perfectly analogous to attention as understood by neuroscience (see Box 3). However, this system does illustrate a route to constructing systems that possess the attention schema indicator.

\pdfbookmark[2]{Implementing agency and embodiment}{imp-ae}
\subsubsection{Implementing agency and embodiment}

The remaining indicators are agency and embodiment. Reinforcement learning is arguably sufficient for agency as we have characterised it (``learning from feedback and selecting outputs so as to pursue goals''), so meeting this part of indicator AE-1 may be very straightforward. Reinforcement learning is very widely used in current AI. To recap the argument for the claim that RL is sufficient for agency, the basic task in RL is for the system to learn to maximise cumulative reward in an environment in which its outputs affect not only the immediate reward it receives but also its opportunities for future reward. RL algorithms are, therefore, designed to be sensitive to the consequences of outputs, making the system more likely to repeat outputs that lead to greater reward over multiple time-steps. This means that there is a contrast between RL and other forms of machine learning. Like systems trained by supervised or self-supervised learning, RL systems gradually come to approximate a desired input-output function. However, in RL what makes this the desired function is that the outputs are conducive to goals that can only be achieved by extended episodes of interaction between the system and the environment. Crucially, RL systems learn these functions by virtue of their sensitivity to relationships between outputs, subsequent inputs, and rewards. Model-based RL systems learn models of these relationships, but model-free systems are also sensitive to them. So RL systems learn and select actions so as to pursue goals (Butlin 2022, 2023).

\begin{figure}[H]
\centering
\includegraphics[width=0.95\linewidth,keepaspectratio]{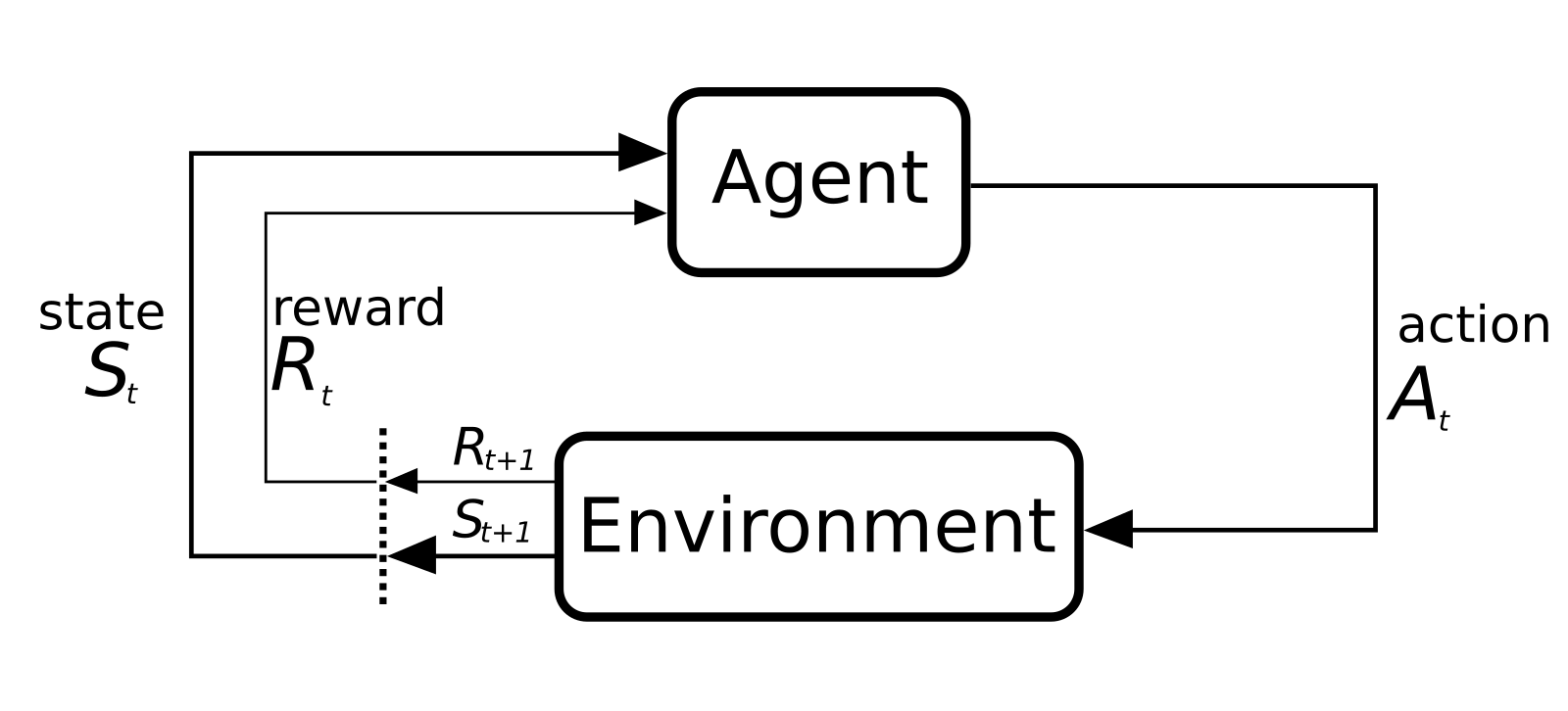} 
\caption{\textbf{Reinforcement learning diagram} of a Markov decision process based on a figure from \textit{Reinforcement Learning: An Introduction} (second edition). by Sutton and Barto. 2020. Creative Commons Attribution-Share Alike 4.0 International. Reprinted with permission.}
\end{figure}

\newpage


The second part of indicator AE-1 says that the probability of consciousness is raised to a greater degree if systems exhibit ``flexible responsiveness to competing goals''. The example motivating this clause is an animal balancing multiple homeostatic drives: this requires prioritisation which is sensitive to shifting circumstances. Several computational approaches to this problem have been explored, including in the context of reinforcement learning (Keramati \& Gutkin 2014, Juechems \& Summerfield 2019, Andersson et al. 2019). One proposal is to use multiple modules that learn independently about how to maximise distinct reward functions, with each of these modules assigning scores to possible actions, and then to pick the action with the highest total score (Dulberg et al. 2023). In the homeostatic case, these reward functions might each correspond to a homeostatic drive, with large penalties for allowing any one of them to depart too far from a set point.

Our embodiment indicator, AE-2, states that systems should use output-input models (also known as forward models) for perception or control. There are specific uses of such models, which neuroscientists have identified in humans, that are particularly associated with embodiment. In perception, predictions of the sensory effects of the system’s own actions can be used to distinguish changes to sensory stimulation caused by these actions from those caused by events in the environment. Systems that engage in this form of inference implicitly or explicitly distinguish themselves from their environments. In motor control, forward models can be used for state estimation and feedback control, enabling rapid adjustments of complex effectors. It is important to distinguish these uses of output-input models from other uses, such as planning, which do not imply embodiment.

Learning output-input models for tasks related to perception and control is common, but there are few examples of current AI systems which meet these specific descriptions. For example, video prediction is a much-studied task (Oprea et al. 2020), and this includes predicting how visual input will develop conditional on the system’s outputs (Finn et al. 2016). But this is not sufficient for the kind of use in perception that we have described, which also includes making sense of exogenous changes which are not predictable from outputs. Output-input models are used in model-based reinforcement learning for control, facilitating successful control of quadrotor drones (Becker-Ehmck et al. 2020) and other robots (Wu et al. 2022). But it is uncertain whether the models are used in these cases for control-specific purposes rather than for planning, a topic that we explore further in section 3.2.2.

One good example of recent research which does employ a forward model for a purpose specific to embodied systems is the system described in Friedrich et al. (2021). In this system, Kalman filtering (Todorov \& Jordan 2002) was used to combine a forward model with sensory inputs, which were subject to a time delay, in order to estimate the current state of the system in its environment. These estimates were then used in ongoing motor control, although only of relatively simple effectors in virtual environments.

\pdfbookmark[1]{Case Studies of Current Systems}{case-studies}
\subsection{Case Studies of Current Systems}
\pdfbookmark[2]{Case studies for GWT}{case-gwt}
\subsubsection{Case studies for GWT}

In practice, it is not always immediately obvious whether a given AI system possesses one of the indicator properties. One reason for this is that we have not given absolutely precise definitions for each indicator. Another is that how deep learning systems work, including what they represent at intermediate layers, is often not transparent. In this section and section 3.2.2 we present case studies illustrating the use of the indicators to assess current AI systems. Here we focus on the GWT indicators (GWT-1 through GWT-4), and on two kinds of systems that are notable for different reasons. These systems are Transformer-based large language models (LLMs) such as GPT-3 (Brown et al. 2020), GPT-4 (OpenAI 2023), and LaMDA (Thoppilan et al. 2022), which are notable for their remarkable performance on natural language tasks and the public attention they have attracted; and Perceiver (Jaegle et al. 2021a) and Perceiver IO (Jaegle et al. 2021b), which are notable because Juliani et al. (2022) argue that they ``implement a functioning global workspace''. Neither of these kinds of systems was designed to implement a global workspace, but there are arguments to be made that they each possess some of the GWT indicator properties.

\begin{wrapfigure}{l}{0.5\textwidth} 
\centering
\includegraphics[width=0.5\textwidth, height=0.36\textheight, keepaspectratio]{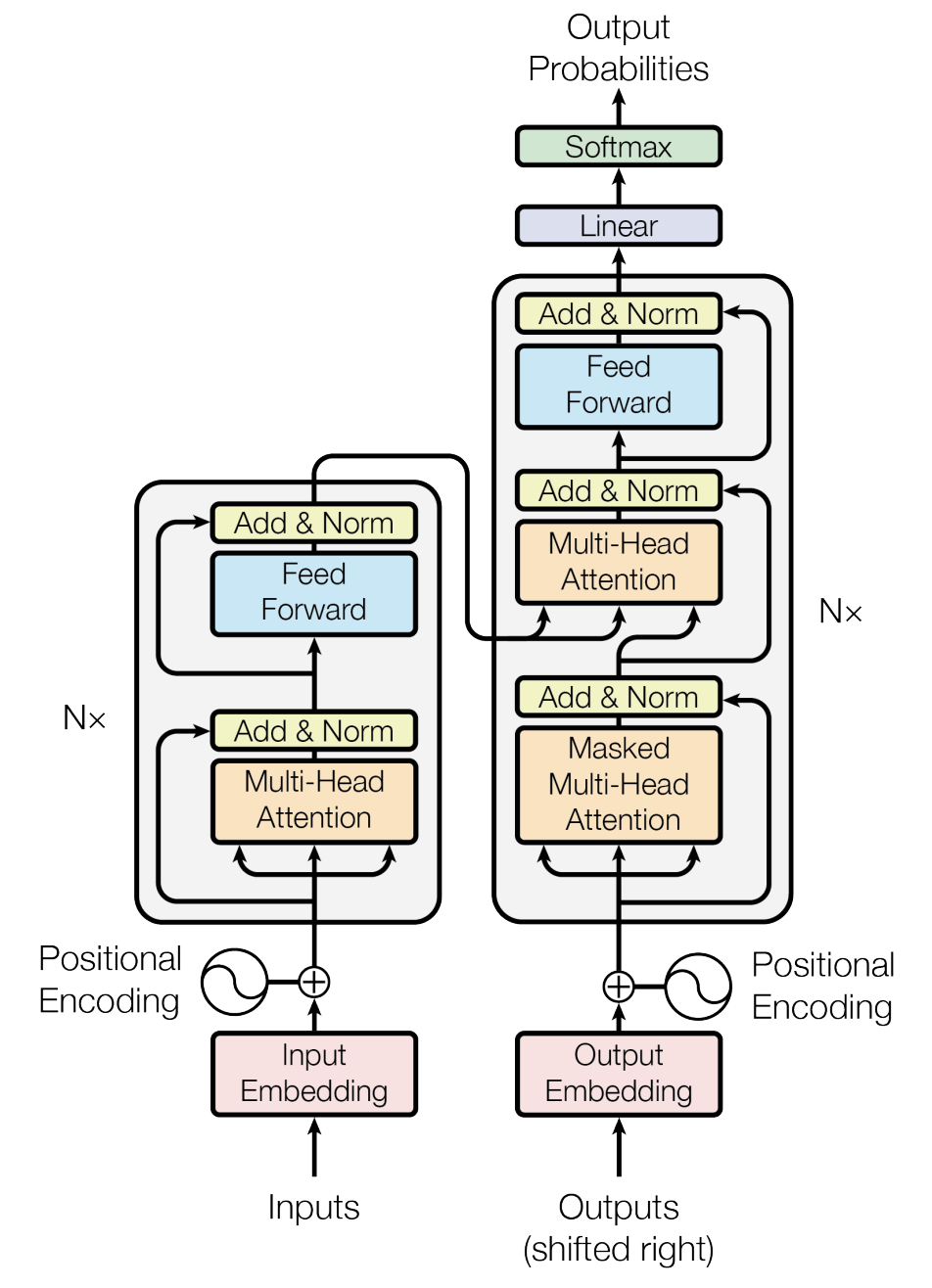} 
\caption{\textbf{The Transformer architecture.}  Vaswani, A., Shazeer, N., Parmar, N., Uszkoreit, J., Jones, L., Gomez, A. N., Kaiser, L., \& Polosukhin, I., 2023. Attention is all you need. arXiv:1706.03762. Reprinted with permission.}
\end{wrapfigure}

In a Transformer, an operation called ``self-attention'' is used to integrate information from different parts of an input, which are often positions in a sequence (see Box 3 on attention; Vaswani et al. 2017). If we see the elements of the system which process information from each position (attention heads) as modules, then there is a basic similarity between the Transformer architecture and the global workspace: both integrate information from multiple modules. Transformers consist of a stack of layers of two types, which alternate: layers of attention heads, which perform the self-attention operation, moving information between positions, and feedforward layers. An interpretation of Transformers by Elhage et al. (2021) describes them as made up of ``residual blocks'', each consisting of one layer of each type, which process information drawn from and then added back to a ``residual stream''.

\FloatBarrier

With the concept of the residual stream in mind, it is possible to argue that Transformers possess indicator properties GWT-1 through GWT-3—that is, that they have modules, a limited-capacity workspace introducing a bottleneck, and global broadcast. The residual stream is the (alleged) workspace, and its dimensionality is lower than that of the self-attention and feedforward layers which write to it and read from it (Elhage et al. 2021). There is ``global broadcast'' in the sense that information in a particular layer in the residual stream can be used by downstream attention heads to influence further processing at any position. The information which is added to the residual stream at a given layer also depends on the state of the residual stream at earlier layers, so on this interpretation one might argue that Transformers meet the state-dependent attention requirement in GWT-4.

One problem with this argument is that it is questionable whether the dimensionality of the residual stream constitutes a bottleneck, since it is the same as that of the input to the system as a whole. However, a more fundamental problem with the argument is that Transformers are not recurrent. The residual stream is not a single network, but a series of layers interspersed between others. There are no modules that pass information to the residual stream and receive it back—instead, when attention heads and feedforward layers combine to write to the residual stream, this affects only what is received by different attention heads and feedforward layers downstream. One way to think about this is to ask which parts of Transformers are the modules. If modules are confined to particular layers, then there is no global broadcast. But if modules are not confined to particular layers, then there is no distinguishing the residual stream from the modules. Transformers lack the overall structure of a system with a global workspace, in that there is no one distinct workspace integrating other elements. There is only a relatively weak case that Transformer-based large language models possess any of the GWT-derived indicator properties.

The two versions of the Perceiver architecture, meanwhile, are closer to satisfying the GWT indicators than Transformers, but in our view still fall short of satisfying them all. The Perceiver architecture was designed to address a weakness of Transformers, which is that the self-attention operation integrates positions by generating pairwise interactions. This is computationally expensive to scale to high-dimensional inputs, motivating an approach that uses a single limited-capacity latent space to integrate information from specialists (Jaegle et al. 2021a, b, Goyal et al. 2022). In particular, Perceiver IO is designed to handle inputs from multiple domains or modalities, and to produce outputs of various kinds, using multiple input encoders and output decoders. It uses self-attention to process information in the latent space, and a related operation, cross-attention, to select information from input modules and write to output modules. The latent space alternates between self-attention and cross-attention layers, enabling it to take new information from input modules repeatedly. Perceiver IO achieves high performance in tasks including language processing, predicting movement in videos, image classification, and processing information about units for the purpose of action selection in Starcraft II (Jaegle et al. 2021b).

\begin{figure}[H]
\centering
\includegraphics[width=0.95\linewidth,keepaspectratio]{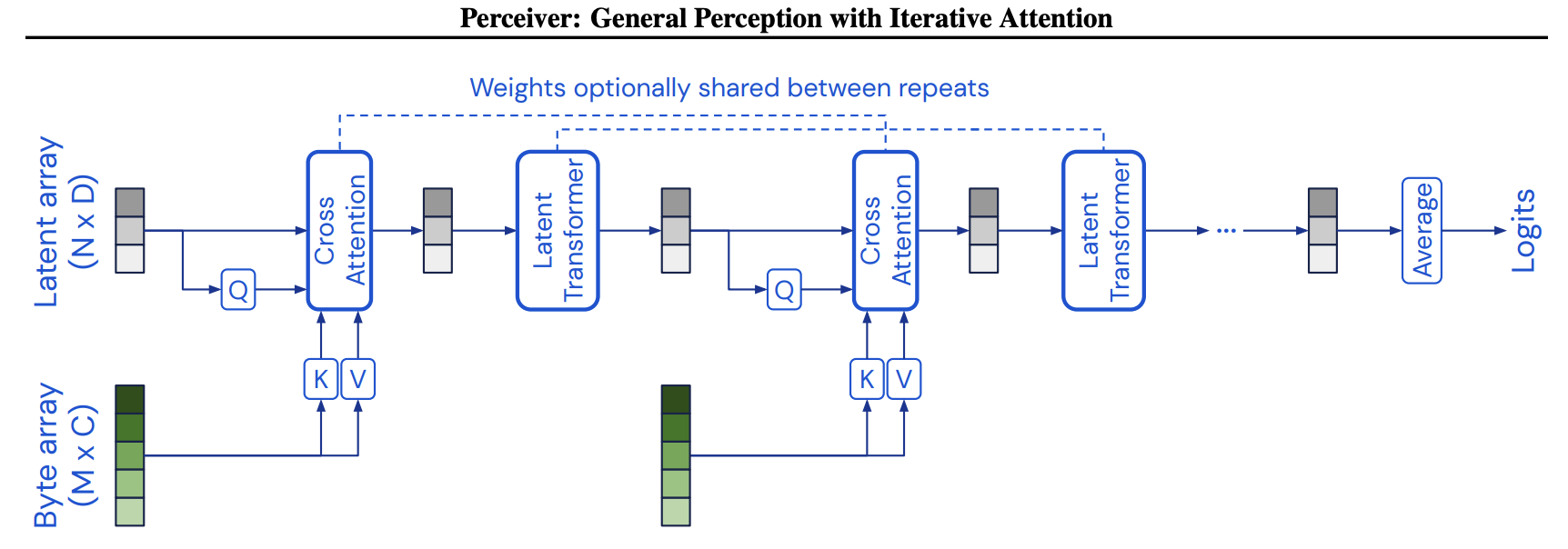} 
\caption{\textbf{Perceiver architecture.} Jaegle, A., Gimeno, F., Brock, A., Vinyals, O., Zisserman, A., \& Carreira, J., 2021. In \textit{Proceedings of the 38th International Conference on Machine Learning}, PMLR 139:4651-4664. Reprinted with permission.}
\end{figure}

The Perceiver architecture allows for sequences of inputs to be processed diachronically, with the latent space state updating with each new input, but also influenced by its previous state. So Perceiver arguably possesses indicator properties GWT-1 (specialised modules) and GWT-2 (bottleneck), as well as the first part of GWT-4 (state-dependent attention). However, it is notable that while specialised modules are possible in the architecture, they are not mandatory—the Perceiver can be used with one input network with a single function. Also, more importantly, the number of inputs that can be processed sequentially is limited by the number of cross-attention layers in the latent space. This means that although attention in Perceiver is state-dependent, in practice the system must be reset to begin a new task, so its states are determined by previous inputs on the current task.

As in the case of Transformers, the clearest missing element of the global workspace in the Perceiver is the lack of global broadcast. Perceiver IO has multiple output modules, but on any given trial, its inputs include an ``output query'', which specifies what kind of output is required. So only one output module acts on information from the workspace. Furthermore, input modules do not generally receive information from the workspace. So while the Perceiver architecture is important as an example of the successful use of a workspace-like method to improve functionality in AI, it is some way from being a full implementation of GWT.

\pdfbookmark[2]{Case studies for embodied agency}{case-ea}
\subsubsection{Case studies for embodied agency}

We now consider examples of AI systems that illustrate what is required for indicators AE-1 and AE-2, which concern agency and embodiment. The systems we discuss in this subsection are: PaLM-E (Driess et al. 2023), described as an ``embodied multimodal language model''; a ``virtual rodent'' trained by RL (Merel et al. 2019); and AdA, a large Transformer-based, RL-trained ``adaptive agent'' (DeepMind Adaptive Agents Team 2023).

\begin{figure}[H]
\centering
\includegraphics[width=0.95\linewidth,keepaspectratio]{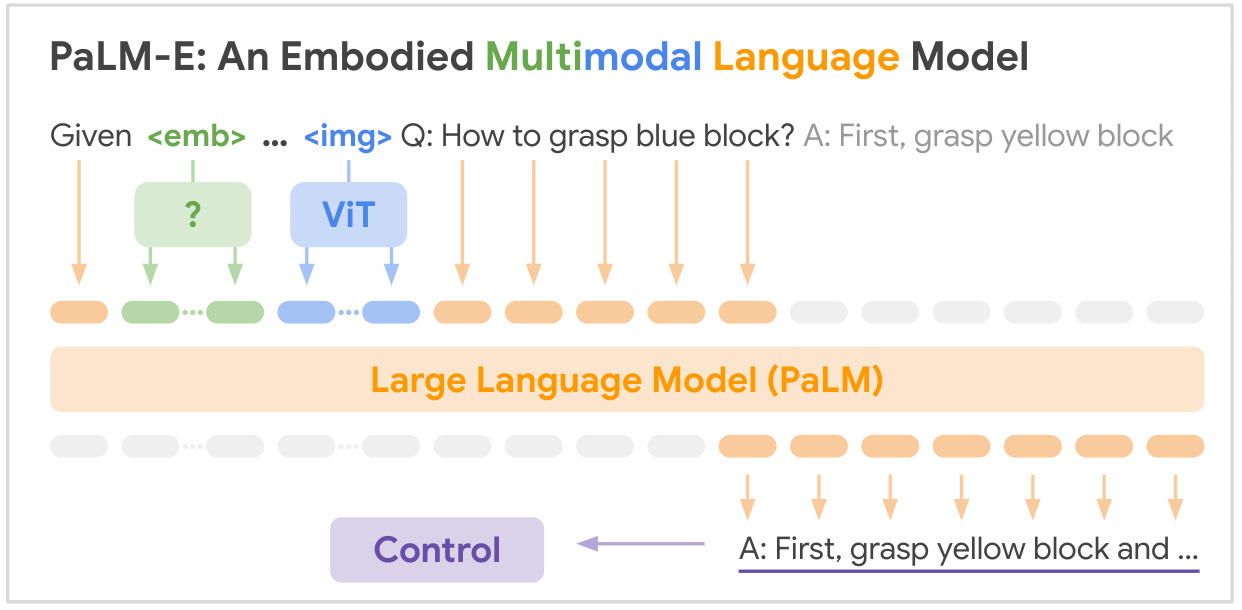} 
\caption{\textbf{PaLM-E architecture.} Driess, D., Xia, F., Sajjadi, M. S. M., Lynch, C., Chowdhery, A., Ichter, B., Wahid, A., Tompson, J., Vuong, Q., Yu, T., Huang, W., Chebotar, Y., Sermanet, P., Duckworth, D., Levine, S., Vanhoucke, V., Hausman, K., Toussaint, M., Greff, K., \& Florence, P., 2023. PaLM-E: An embodied multimodal language model. \textit{arXiv:2303.03378}. Reprinted with permission.}
\end{figure}

PaLM-E is a decoder-only LLM (Driess et al. 2023), fine-tuned from PaLM (Chowdhery et al. 2022). It takes encoded multimodal strings which can include both text and images as input, and produces text tokens as output. Like other LLMs, it can be used autoregressively to produce extended strings of text. Its use requires an encoder, and in combination with a suitable encoder, it can perform well in both pure language tasks and vision-language tasks such as visual question answering. However, PaLM-E can also be combined with a separately-trained policy unit that maps natural-language instructions and visual context to low-level robot actions. The PaLM-E study used policies from Lynch et al. (2022) which were trained to imitate human control of robot actuators. In this setup, the system can take input from a camera, combine it with task instructions provided by a human, and generate and execute plans for action. PaLM-E generates high-level plans while the policy unit provides low-level vision-guided motor control. If necessary, the plan can be updated later based on new observations.

In considering Palm-E, we can focus on several different systems: the PaLM-E model itself, the policy unit, or the complete system comprised of these two elements and the robot they control.

The complete system is very naturally described as an embodied agent. When given tasks, it can make plans and execute them by moving its robotic ``body'' through spaces it shares with humans. However, both of the main components of the system are trained, in effect, to imitate human behaviours. PaLM-E is trained by self-supervised learning to predict the next token in human-generated strings, and the policy unit is trained to imitate human visuomotor control. So the system arguably \textit{imitates} planning and using visuomotor control to execute plans, as opposed to actually doing these things. It does not learn to pursue goals from feedback about success or failure.

Since the complete system includes a robot, it is natural to describe it as embodied. However, according to the account we have adopted, for a system to have an embodied perspective it must model how its outputs affect the environment in order to distinguish itself from the environment in perception or to facilitate motor control. It is hard to see how the complete PaLM-E system could have learnt a model of this kind, given that it is not trained end-to-end and has, therefore, not been exposed to the effects of its outputs on the environment in the course of training. The components are also trained in environments in which there is little or no exogenous change, so the need to learn to disambiguate sources of change in input is absent.

Perhaps the best case to be made for embodiment here focuses on the policy unit. This is a Transformer-based network trained to map from video and text to continuous actions in imitation of human control. In human motor control, a forward model (mapping outputs to predicted inputs) is used to generate expectations for comparison to intentions and sensory inputs so that movement trajectories can be corrected in real time (McNamee \& Wolpert 2019). This may involve distinguishing the self from the environment, since exogenous events may be responsible for mismatches. Something like this mechanism might in principle be used by the policy unit because the input to the network includes some observation history. This means that it could learn to detect a mismatch between the observations it would have predicted given past states and those it is now receiving and correct its actions accordingly. For example, if it is instructed to push a block East and observes that the block is moving North as it acts, this could prompt a correction.

It could also be argued that the policy unit is an agent, even though it is not trained by RL. It can learn sequences of inputs that are associated with progress towards a goal, specified by another part of the input, and adjust its outputs so as to generate these sequences. However, it is questionable whether this is enough for agency because although the system learns about sequences of inputs and about which outputs to produce when confronted with these sequences, it does not learn how its outputs affect inputs – so it cannot produce outputs because it expects them to help to generate the right sequences. A similar objection can also be made against the case for embodiment: the system can learn to detect ``mismatches'' between past and present observations in the context of a task, but not between past observations and actions, on the one hand, and present observations on the other. This is not to say that the policy unit definitively lacks agency and embodiment but to illustrate the kinds of considerations that are relevant to these issues.

Given that controlling an avatar in a simulated environment can be enough for embodiment, the ``virtual rodent'' of Merel et al. (2019) is a promising candidate for these attributes. It is trained by RL, which is sufficient for agency. This system was constructed by implementing a virtual ``body'' with 38 degrees of freedom based on the anatomy of laboratory rats. A recurrent LSTM-based actor-critic architecture was trained end-to-end by RL to control this body, using rich visual and proprioceptive inputs, to perform four tasks in a static 3D environment. This architecture is plausibly sufficient for the system to learn a model of how its outputs affect its inputs which it could use in perception and action. Because recurrent networks are used, new inputs can be processed in the context of stored representations of inputs and outputs from the recent past. So the system can, in principle, process inputs and select outputs in the context of expectations generated from its past behaviour—that is, in the context of outputs from a self-model, which would (perhaps implicitly) represent information about stable properties of its body.

However, in this study, the environment did not change exogenously, and it appears the tasks could have been solved by producing combinations of a set of stereotyped movements, such as running, jumping and rearing. The analysis indicates that the system did learn a repertoire of behaviours like these. These movements may not need ongoing control from perceptual feedback. This raises a question about whether the task demands were sufficient to cause the system to learn to use a self-model rather than a relatively simple input-output policy.

AdA, DeepMind’s ``adaptive agent'', is also trained end-to-end by RL to control an avatar in a 3D virtual environment (DeepMind Adaptive Agents Team 2023). The system consists of a Transformer which encodes observations at the past few hundred timesteps (the exact figure varied in the team’s experiments), including past actions and rewards as well as task instructions and sensory input and feeds into an LSTM which is trained to predict future actions, values and rewards. This system was trained on a very wide range of tasks, of increasing difficulty, in order to cause it to undergo meta-RL. This meant that it learnt an implicit online learning algorithm, allowing it to learn to perform new tasks from observations in its memory—the Transformer encoding recent past timesteps—without updating weights. After this training regime, the system was able to adapt to new tasks in its environment at human timescales.

AdA is a significantly more powerful system than the virtual rodent, although it controls a much less complex avatar. Its environment is somewhat less stable, including because some of the tasks it performs involve cooperating with another agent, but the emphasis is on learning to generate and test behavioural strategies for new tasks, rather than on coordinating complex dynamics. So again we could question whether the system faces the kinds of challenges that seem to have been responsible for the evolution of self-modelling in animals. However, unlike the virtual rodent, it has a specific training objective to generate predictions based on past sequences of interleaved inputs and outputs. AdA may, therefore, be the most likely of the three systems we have considered to be embodied by our standards, despite not being physically embodied (like PaLM-E) or being trained primarily for motor control of a complex avatar (like the virtual rodent).

\newpage

\pdfbookmark[0]{Implications}{implications}
\section{Implications}

In this final section, we discuss the topic of consciousness in AI in a broader context. We consider risks from under- and over-attribution of consciousness to AI systems (in section 4.1) and the relationship between consciousness and AI capabilities (in section 4.2). We also make some limited recommendations, in section 4.3. Our comments in this section are brief—the main aim of this report is to propose a scientific approach to consciousness in AI and to establish what current scientific theories imply about this prospect, not to investigate its moral or social implications.

\pdfbookmark[1]{Attributing Consciousness to AI}{attributing-consciousness}
\subsection{Attributing Consciousness to AI}

There are risks on both sides of the debate over AI consciousness: risks associated with under-attributing consciousness (i.e. failing to recognize it in AI systems that have it) and risks associated with over-attributing consciousness (i.e. ascribing it to systems that are not really conscious). Just as in other cases of uncertain consciousness, such as other animals (Birch 2018) and people with disorders of consciousness (Peterson et al. 2015; Johnson 2022), we must consider both types of risk.

\pdfbookmark[2]{Under-attributing consciousness to AI}{under-attributing}
\subsubsection{Under-attributing consciousness to AI}

As many authors have noted, as we develop increasingly sophisticated AI systems we will face increasingly difficult questions about their moral status (Bryson 2010, Gunkel 2012, Schwitzgebel \& Garza 2015, 2020, Metzinger 2021, Shulman \& Bostrom 2021). Philosophers disagree about the exact relationship between being conscious and moral status, but it is very plausible that any entity which is capable of conscious suffering deserves moral consideration. If we can reduce conscious suffering, other things being equal, we ought to do so. This means that if we fail to recognise the consciousness of conscious AI systems we may risk causing or allowing morally significant harms.

An analogy with non-human animals helps to illustrate the issue. Humans mistreat farmed animals in very large numbers, motivated by powerful economic incentives. Whether or not this mistreatment depends on a failure to attribute consciousness to these animals, it illustrates the potential problem. In the case of AI, there is likely to be considerable resistance to attributions of consciousness, partly because developers of AI may have powerful economic incentives to downplay concerns about welfare. So if we build AI systems that are capable of conscious suffering, it is likely that we will only be able to prevent them from suffering on a large scale if this capacity is clearly recognised and communicated by researchers. However, given the uncertainties about consciousness mentioned above, we may create conscious AI systems long before we recognise we have done so.

An important point in this context is that being conscious is not the same as being capable of conscious suffering. It is at least conceptually possible that there could be conscious systems that have no \textit{valenced} or \textit{affective} conscious experiences—that is, no experiences that feel good or bad to them (Carruthers 2018, Barlassina \& Hayward 2019). If it is only valenced experiences that have special moral significance, then the key question for establishing the moral status of AI systems is whether they are capable of such experiences (that is, whether they are sentient, as this term is sometimes used). We have not discussed neuroscientific or philosophical theories of valence or otherwise investigated the prospects of specifically valenced conscious experience in AI systems. Theories of valenced conscious experience are less mature than theories of visual experience, suggesting an important priority for future work. However, we suspect that many possible conscious systems which are also agents will have valenced experiences since agents must evaluate options in order to select actions.

In short, the risk of under-attributing consciousness should be taken seriously. Failing to recognise conscious AI systems as such could lead us to cause unwarranted suffering to many conscious beings.

\pdfbookmark[2]{Over-attributing consciousness to AI}{over-attributing}
\subsubsection{Over-attributing consciousness to AI}

There is also a significant chance that we could over-attribute consciousness to AI systems—indeed, this already seems to be happening—and there are also risks associated with errors of this kind. Most straightforwardly, we could wrongly prioritise the perceived interests of AI systems when our efforts would better be directed at improving the lives of humans and non-human animals. As Schwitzgebel and Garza (2015, 2020) argue, uncertainty about the moral status of AI systems is dangerous because both under- and over-attribution can be costly.

Over-attribution is likely because humans have a well-established tendency to anthropomorphise and over-attribute human-like mental states to non-human systems. A growing body of work examines the tendency of people to attribute consciousness and agency to artificial systems, as well as the factors influencing this attribution (Dennett 1987, Gray \& Wegner 2012, Kahn et al. 2006, Krach et al. 2008, Sytsma 2014). There may be various reasons for humans to have an ``agent bias''—a natural inclination to ascribe agency, intentions, and emotions to non-human entities—stemming from our evolutionary history (Guthrie 1993).

One possibility is that we might anthropomorphise AI systems because it seems to help us to understand and predict their behaviour—although this impression could be false, with anthropomorphism in fact causing us to make incorrect interpretations. Anthropomorphism allows us to understand and anticipate complex systems like AI using the same cognitive frameworks we use for understanding humans, potentially helping us navigate interactions with AI (Epley et al., 2007). Dennett postulates that individuals employ a cognitive strategy, called the ``intentional stance'', to predict and explain the behaviour of various entities, including humans, animals, and artificial systems (Dennett 1987). The intentional stance involves attributing mental states, such as beliefs, desires, and intentions, to an entity to decipher and anticipate its behaviour. Observation and interaction with AI systems often lead to the natural adoption of the intentional stance, especially when their behaviours appear purposeful or goal-directed. This tendency is further amplified when AI systems exhibit human-like characteristics, such as natural language processing, facial expressions, or adaptive learning capabilities (Mazor et al. 2021). Researchers have identified several factors that predispose individuals to anthropomorphise AI systems, including their physical appearance, behaviour, and perceived autonomy (Kahn et al. 2006; Złotowski et al. 2015).

Attributions of agency and consciousness to artificial agents may also be driven by an emotional need for social interaction (Mazor et al. 2021). Individuals who seek social interaction and fulfillment from artificial systems may be more prone to attributing consciousness to them. Assigning human-like traits to AI can help people cope with difficult emotions or situations, such as feeling more comfortable confiding in an AI that appears empathetic or understanding (Turkle 2011). The use of artificial systems such as Replika chatbots as sources of social interaction is already evident.

The recent rapid progress in language model capabilities is particularly likely to drive over-attribution of consciousness, as the case of Blake Lemoine arguably illustrates (Lemoine 2022). Modern LLMs can convincingly imitate human discourse, making it difficult to resist the impression that one is interacting with a conscious agent, especially if the model is prompted to play the role of a person in a conversation (Shanahan et al. 2023).

In addition to the risk of misallocation of resources which we mentioned above, over-attributing consciousness to AI systems creates risks of at least three other kinds. First, if consciousness is sometimes attributed to AI systems on weak grounds, these attributions may undermine better-evidenced claims of consciousness in AI. Observers of debates on this topic may recognise that some attributions are weak, and assume that all are. This effect could be particularly powerful if there are also reasonable concerns that attributions of consciousness to AI are distracting us from addressing other pressing problems. Second, if we judge that a class of AI systems are conscious, this should lead us to treat them differently—training them in different ways, for instance. In principle, this could conflict with work to ensure that AI systems are developed in ways that benefit society. And third, overattribution could interfere with valuable human relationships, as individuals increasingly turn to artificial agents for social interaction and emotional support. People who do this could also be particularly vulnerable to manipulation and exploitation.

Whatever one’s views about the relative importance of these various risks, they amount to a powerful case for research on the prospect of consciousness in AI. If the development of AI is to continue, this research will be crucial: it is risky to judge without good grounds either that no AI systems can be conscious, or that those of a particular class are conscious.

\pdfbookmark[1]{Consciousness and Capabilities}{consciousness-and-capabilities}
\subsection{Consciousness and Capabilities}

In the popular imagination, consciousness is associated with free will, intelligence and the tendency to feel human emotions, including empathy, love, guilt, anger and jealousy. So our suggestion that conscious AI may be possible in the near-term might be taken to imply that we will soon have AI systems akin to the very human-like AIs depicted in science fiction. Whether this in fact follows depends on the relationships between consciousness and other cognitive traits and capacities. Furthermore, conscious AI systems are more likely to be built if consciousness is (or is expected to be) associated with valuable capabilities in AI. So in this subsection, we briefly consider how consciousness might be related to differences in AI systems’ behaviour and capabilities.

One possible argument for the view that we are likely to build conscious AI is that consciousness is associated with greater capabilities in animals, so we will build conscious AI systems in the course of pursuing more capable AI. It is true that scientific theories of consciousness typically claim that conscious experience arises in connection with adaptive traits, selected for the contributions they make to cognitive performance in humans and some other animals. For example, Baars (1988, 1997) claims that the global workspace ``optimizes the trade-off between organization and flexibility'' (1988, p. 348), a trade-off that advanced AI systems must also presumably manage.

A weakness of this argument is that human and animal minds are not necessarily a good guide to the connection between consciousness and capabilities in artificial systems. This is because the ``design'' of animal minds is explained not only by the adaptive value of our capabilities but also by the constraints under which we evolved, which include limits on the quantity and form of data from which we can learn; limits on the amount of energy available to power our minds; and the forms of our ancestors’ minds and the availability of relevant mutations. The space of possible designs for AI systems is different from the space of possible mind designs in biology (Summerfield 2022). So we may well find ways to build high-performing AI systems which are not conscious.

However, some influential AI researchers are currently pursuing projects that aim to increase AI capabilities by building systems that are more likely to be conscious. We have already discussed the ongoing work by Bengio and colleagues which uses ideas from GWT. Goyal and Bengio (2022) write that:

\begin{myquote}
Our aim is to take inspiration from (and further develop) research into the cognitive science of conscious processing, to deliver greatly enhanced AI, with abilities observed in humans thanks to high-level reasoning.
\end{myquote}

The aim here is to build AI systems that implement at least some of the features underlying consciousness in humans, specifically in order to enhance capabilities. Similarly, LeCun’s proposed architecture for autonomous intelligence has features such as a world model for planning and a ``configurator'', described as an element that ``takes inputs from all other modules and configures [the other modules] for the task at hand'' (LeCun 2022, p. 6). LeCun also suggests that the existence of elements like these in the human brain may be responsible for what he calls ``the illusion of consciousness''. These are examples of a widespread and long-established practice in AI of drawing on insights from the cognitive sciences (Hassabis et al. 2017, Zador et al. 2022). So regardless of whether building in consciousness-associated features is the only possible route to greater capabilities, it is one that we are likely to take.

Turning now to the issue of how conscious artificial systems are likely to behave, there are both conceptual and empirical reasons to doubt that being conscious implies having human-like motives or emotions. Conceptually, to be conscious is simply to have subjective experiences. In principle, there could be conscious subjects who had experiences very unlike ours. In particular, there is no apparent conceptual incoherence in the idea of motivationally neutral conscious experiences which lack valence or affect—and even if a conscious subject does have valenced or affective states, these may be triggered by different circumstances from those which trigger our emotions, and motivate them to behave in different ways.

Empirically, the theories of consciousness we have discussed do not generally claim that consciousness implies human-like motives or emotions. One exception is the view that consciousness is only possible in the context of self-maintenance, which we discussed in section 2.4.5, since this presumably entails not only that conscious beings engage in self-maintenance, but that they are motivated to do so. But this view is an outlier. Some theories, such as GWT and PRM, suggest that conscious subjects must be agents and perhaps that their desires are likely to enter into conscious experience—in GWT, the workspace maintains representations that provide important context for processing in the modules, and current desires might fit this description. But these theories are neutral about what motivates conscious subjects. AST is a somewhat more complicated case: Graziano (2013) argues that the use of an attention schema is important for social cognition because it allows one to model attention either in oneself or in others. He further claims that this is a reason to build conscious AI because AI systems with attention schemas will be capable of empathy (Graziano 2017). But the claim here seems to be that implementing AST is necessary for attributing conscious states to others, and, therefore, for being motivated by empathy to help them, not that consciousness is sufficient for empathetic motives.

Many current concerns about the impacts of AI do not turn on whether AI systems might be conscious. For example, the concern that AI systems trained on data that reflects the current structure of society could perpetuate or exacerbate injustices does not turn on AI consciousness. Nor does the concern that AI could enable various forms of repression, or that AI systems could replace human workers in most jobs (although the economic value of replacing human workers may motivate capabilities research which leads to conscious AI, as we have noted). Perhaps most surprisingly, arguments that AI could pose an existential risk to humanity do not assume consciousness. A typical argument for this conclusion relies on the premises that (i) we will build AI systems that are very highly capable of making and executing plans to achieve goals and (ii) if we give these systems goals that are not well chosen then the methods that they find to pursue them may be extremely harmful (see e.g. Hilton 2022). Neither these premises nor the ways in which they are typically elaborated and defended rely on AI systems being conscious.

\pdfbookmark[1]{Recommendations}{recommendations}
\subsection{Recommendations}

Several authors have made important practical recommendations concerning the possibility of AI consciousness. These include: Bryson’s (2010) case that conscious AI should be avoided; Metzinger’s (2021) call for a moratorium on work that could lead to conscious AI; Graziano’s (2017) arguments in favour of conscious AI; Schwitzgebel and Garza’s (2020) argument that we should only build particular AI systems if we can either be confident that they will be conscious, or be confident that they will not be; and the detailed recommendations of Bostrom and Shulman (2022). We have not assessed the arguments for these recommendations and do not consider them further here.

However, we do recommend support for research on the science of consciousness and its application to AI (as recommended in the AMCS open letter on this subject; AMCS 2023), and the use of the theory-heavy method in assessing consciousness in AI. One way that we can make progress in learning which AI systems are likely to be conscious is by developing and testing scientific theories of consciousness; theoretical refinements within the existing paradigms are valuable, as well as attempts to test predictions of competing theories, as in the ongoing Cogitate adversarial collaboration (Melloni et al. 2021). A second is by undertaking careful empirical research to extend theories of consciousness to non-human animals. This will help us to establish a more general account of the correlates of consciousness, based on evidence from a wider range of cases (Andrews \& Birch 2023). And a third is research that refines theories of consciousness specifically in the context of AI. Research of this kind may involve theorising about AI implementations of mechanisms implicated in theories of consciousness; building such systems and testing their capacities; identifying ambiguities in existing theories; and developing and defending more precise formulations of theories, so that their implications for AI are clearer. Integrating work of this kind with continued empirical research on human and animal consciousness can be expected to be especially productive.

Two other lines of research may be important. One is research specifically on valenced and affective consciousness; if conscious experience of this kind is especially morally important, as seems to be the case, then we have a pressing need for well-founded computational theories which can be applied to AI. The other is efforts to develop better behavioural tests for consciousness in AI. Although our view is that the theory-heavy approach is currently the most promising, it may be possible that behavioural tests could be developed which are difficult to game and based on compelling rationales, perhaps informed by theories. If such tests can be developed, they may have practical advantages over theory-heavy assessments.

As we learn more about consciousness, and gain better tools to assess whether it may be present in AI systems, we must also use these tools effectively. That means applying the theory-heavy method as we develop new kinds of AI systems, both prospectively—in assessing the likelihood of consciousness in planned systems, before we build them—and retrospectively. In retrospective evaluation, methods for mechanistic interpretability may be important, since it is possible that systems that do not have potentially consciousness-supporting features built in could acquire them in the course of training. It also means that scientific theories should be central to consciousness-related AI policy and regulation.

\begin{infobox} [Box 4: Open questions about consciousness in AI]

This report is not the last word on AI consciousness. Far from it: one of our major aims is to spur further research. Further progress in the neuroscience of consciousness will contribute to our understanding of the conditions for consciousness in AI, and we have suggested that research on consciousness in non-human animals may be particularly valuable. Here, we highlight research questions that are relevant for understanding AI consciousness in particular.
\infopar
\textbf{\textit{Refining and extending our approach}}

While following the same basic approach as this report, further research could:

\begin{itemize}
    \item Examine other plausible theories of consciousness, not considered in this report, and use them to derive further indicators of consciousness;
    \item Refine or revise the indicators which we have derived from the theories considered here;
    \item Conduct assessments of other AI systems, or investigate different ways in which the indicators could be implemented.
\end{itemize}

Furthermore, our approach could be extended by developing a formal evaluation procedure for consciousness that could be applied to AI systems, although it is questionable whether this would be justified at present, in the context of significant uncertainty about computational functionalism and about particular scientific theories.
\infopar
\textbf{\textit{Computational functionalism and rival views}}

Determining whether consciousness is possible on conventional computer hardware is a difficult problem, but progress on it would be particularly valuable, and philosophical research could contribute to such progress. For example, sceptics of computational functionalism have noted that living organisms are not only self-maintaining homeostatic systems but are made up of cells that themselves engage in active self-maintenance (e.g. Seth 2021, Aru et al. 2023); further work could clarify why this might matter for consciousness. Research might also examine whether there are features of standard computers which might be inconsistent with consciousness, but would not be present in unconventional (e.g. neuromorphic) silicon hardware. A further topic for philosophical research is the individuation of AI systems, given that they can be copied, distributed, called in multiple places at once, and so forth.
\infopar
\textbf{\textit{Valence and phenomenal character in AI}}

We have set aside the question of what \textit{kinds} of experiences conscious AI systems might have. In principle, these could be quite different from human experiences. An important issue is what it would take for AI systems to be capable of having valenced conscious experiences—that is, ones that feel good or bad—because these might have special moral significance. Research on computational theories of valence would help to address this issue.
\infopar
\textbf{\textit{AI interpretability research}}

A significant obstacle for the theory-heavy approach is our limited understanding of the inner workings of complex deep learning systems. AI interpretability research seeks to illuminate and explain these workings, and can, therefore, contribute to many forms of research on AI, including research on consciousness.
\infopar
\textbf{\textit{Behavioural tests and introspection}}

Although we have stressed the limitations of behavioural tests for consciousness in AI, we are open to the possibility that compelling tests could be developed, so research towards this end may be valuable. Behavioural tests need not be ``theory-neutral'', but could be based on specific scientific theories. Another possible approach is to develop AI systems with reliable introspective capacities that could allow them to meaningfully report on their own consciousness or lack thereof (Long forthcoming).
\infopar
\textbf{\textit{The ethics of research on AI consciousness}}

On balance, we believe that research to better understand the mechanisms which might underlie consciousness in AI is beneficial. However, research on this topic runs the risk of building (or enabling others to build) a conscious AI system, which should not be done lightly. Mitigating this kind of risk should be carefully weighed against the value of better understanding consciousness in AI.

\end{infobox}

\newpage

\pdfbookmark[0]{Glossary}{glossary}
\section*{Glossary}
\addcontentsline{toc}{section}{Glossary}

\begin{longtable}{|>{\raggedright\arraybackslash}p{0.3\textwidth}|>{\raggedright\arraybackslash}p{0.65\textwidth}|}
\hline
\rowcolor{black} 
\multicolumn{1}{|c|}{\textcolor{white}{\textbf{Term}}} & \multicolumn{1}{c|}{\textcolor{white}{\textbf{Description}}} \\
\hline
\endhead
\textbf{AST} & Attention schema theory \\
\hline
\textbf{GWT} & Global workspace theory \\
\hline
\textbf{HOSS} & Higher-order state space theory \\
\hline
\textbf{HOT} & Higher-order theory \\
\hline
\textbf{PP} & Predictive processing \\
\hline
\textbf{PRM} & Perceptual reality monitoring theory \\
\hline
\textbf{RPT} & Recurrent processing theory \\
\hline
\textbf{UAL} & Unlimited associative learning \\
\hline
\textbf{access consciousness} & ``Functional'' concept contrasted with \textbf{phenomenal consciousness}; a state is access conscious if its content is directly available to its subject to perform a wide range of cognitive tasks such as report, reasoning, and rational action \\
\hline
\textbf{agency} & Systems are agents if they pursue goals through interaction with an environment (see section 2.4.5(a)) \\
\hline
\textbf{algorithmic recurrence} & Form of processing in which the same operation is applied repeatedly, such as processing in a neural network in which information passes through layers with the same weights; contrasted with \textbf{implementational recurrence} \\
\hline
\textbf{assertoric force} & A feature of some conscious experiences that present their contents as accurately representing the world, disposing us to form corresponding beliefs \\
\hline
\textbf{attractor dynamics} & The property of some dynamical systems that the system tends to converge to one of a number of stable ``attractor'' states \\
\hline
\textbf{backward masking} & Presenting a stimulus (mask) soon after a briefly-presented target stimulus, with the effect that the first stimulus is not consciously perceived \\
\hline
\textbf{Bayesian inference} & Inference according to Bayes' Rule, which states that the probability that a state holds given some data is proportional to the prior probability of the state multiplied by the probability of the data given the state \\
\hline
\textbf{binocular rivalry} & An experimental paradigm in which a different image is presented to each eye, leading to perceptual alternations between the images \\
\hline
\textbf{classical conditioning} & A form of learning in which a neutral stimulus is paired with a stimulus that triggers an innate response, such as approach or avoidance, leading to an association between the previously neutral stimulus and the response \\
\hline
\textbf{cognitive architecture} & The way in which functional units are arranged and connected in a cognitive system \\
\hline
\textbf{computational functionalism} & The thesis that implementing computations of a certain kind is necessary and sufficient for consciousness (see section 1.2.1) \\
\hline
\textbf{confidence or credence} & Degree of belief; the probability that one would assign to a claim \\
\hline
\textbf{contrastive analysis} & Method in consciousness science in which conscious and unconscious conditions, for example, conscious and unconscious visual processing of a stimulus, are contrasted with each other; e.g. to find corresponding differences in brain activity \\
\hline
\textbf{cross-attention} & Process in AI allowing information to be selected and passed from one part of a system to another (see Box 3) \\
\hline
\textbf{embodied, enactive and extended cognition} & The view that cognitive processes are deeply influenced by, and sometimes employ, the body and environment \\
\hline
\textbf{end-to-end training} & In machine learning, training all components of a system together, as opposed to combining components trained separately \\
\hline
\textbf{feature extraction} & Identifying features in visual scenes \\
\hline
\textbf{feature binding} & Integrating the separate features of objects, for example ``yellow'' and ``square'', into a unified representation \\
\hline
\textbf{feedforward processing} & Processing in which a non-repeating set of operations is applied sequentially, as opposed to \textbf{recurrent processing} \\
\hline
\textbf{figure-ground segregation} & Distinguishing objects from background in vision \\
\hline
\textbf{first-order representations} & Representations that are about the non-representational world, in contrast with \textbf{higher-order representations}; paradigm cases include the visual representation of an external object like an apple \\
\hline
\textbf{forward models} & Models of how outputs will lead to changes in the body and environment, especially in the context of motor control \\
\hline
\textbf{generative model} & A statistical model of input data (e.g. patterns of sensory stimulation) that can be sampled to produce new synthetic data \\
\hline
\textbf{global broadcast} & In GWT, information being made available to all modules (see section 2.2) \\
\hline
\textbf{goal-directed vs. habitual behaviour} & In psychology, distinction between behaviour that is caused by representations of outcome values and action-outcome contingencies (goal-directed) and behaviour that is caused by representations of the values of actions in situations (habitual) \\
\hline
\textbf{higher-order representations} & Representations that are about other representations (e.g. a representation that another representation is reliable) \\
\hline
\textbf{ignition} & In GWT, a sudden, non-linear change in brain activity associated with a state's coming to be \textbf{globally broadcast} \\
\hline
\textbf{implementational recurrence} & Feature of computing systems, widespread in the brain, in which \textbf{algorithmic recurrence} is implemented by feedback loops passing information repeatedly through the same units \\
\hline
\textbf{indicators/indicator properties} & Properties identified in this work that make AI systems more likely to be conscious \\
\hline
\textbf{key-query attention} & Process for selecting information in AI; \textbf{cross-attention} and \textbf{self-attention} are forms of key-query attention (see Box 3) \\
\hline
\textbf{latent space} & Representation space representing variables that are not directly observable, or architectural feature of an AI allowing such a space to be learnt \\
\hline
\textbf{masked priming experiments} & Experiments in which a masked stimulus (see \textbf{backward masking}) influences (typically, facilitates, i.e. primes) the processing of a second, visible, stimulus \\
\hline
\textbf{metacognition} & Cognition about one's own cognitive processes, for example about their reliability or accuracy \\
\hline
\textbf{metacognitive monitoring} & Monitoring of cognitive processes by other cognitive systems (see section 2.3) \\
\hline
\textbf{modules} & In GWT, specialised subsystems capable of performing tasks independently (see section 2.1) \\
\hline
\textbf{multi-head attention layers} & In Transformers, layers that implement variants of self- or cross-attention in parallel \\
\hline
\textbf{natural kind} & Collection of entities that have similar superficial properties as a result of a shared underlying nature (e.g. gold, tigers); consciousness may be a psychological/computational natural kind \\
\hline
\textbf{no-report paradigm} & Experimental paradigm using \textbf{contrastive analysis} but with other measures (e.g. brain activity, eye movements) used to distinguish conscious and unconscious conditions \\
\hline
\textbf{neural correlates of conscious states} & Minimal sets of neural events that are jointly sufficient for conscious states \\
\hline
\textbf{perceptual organisation} & Generation in perception of a representation of an organised scene, e.g. distinguishing objects and representing their relative locations \\
\hline
\textbf{perspective} & Feature of embodied systems that their inputs depend on their position in the environment and are affected by their movements \\
\hline
\textbf{phenomenal consciousness} & Consciousness as we understand it in this report; see section 1.1 \\
\hline
\textbf{phenomenal character} & The content of an experience; what it is like for the subject to have that experience \\
\hline
\textbf{predictive coding} & Form of computation described by the predictive processing theory of cognition, in which a hierarchical \textbf{generative model} predicts inputs and the state of the environment is estimated by prediction error minimisation \\
\hline
\textbf{problem intuitions} & Intuitions that humans have about consciousness that are difficult to reconcile with materialism \\
\hline
\textbf{quality space} & The similarity space of phenomenal qualities, which are the properties making up the phenomenal character of an experience \\
\hline
\textbf{recurrent processing} & See \textbf{algorithmic recurrence} and \textbf{implementational recurrence} \\
\hline
\textbf{reinforcement learning} & Form of machine learning in which the objective is to maximise cumulative reward through interaction with an environment (see section 3.1.5); also a form of biological learning \\
\hline
\textbf{self-attention} & Process in AI in which information about multiple input tokens is integrated through pairwise interactions between functional units; key process in Transformers (see Box 3) \\
\hline
\textbf{sentience} & Term sometimes, but not always, used synonymously with ``consciousness''; see section 1.1 \\
\hline
\textbf{subjective experience} & Conscious experience; see section 1.1 \\
\hline
\textbf{System 2 thought} & Controlled, effortful, often verbalisable thought relying on explicit knowledge; contrasted with more automatic System 1 thought in ``dual-process'' theories of cognition \\
\hline
\textbf{theory-heavy approach} & Method for determining which systems are conscious based on scientific theories of consciousness \\
\hline
\textbf{Transformer} & Deep learning architecture, the basis for many large language models and other recent AI systems that relies on the \textbf{multi-head attention} mechanism (see section 3.2.1 and Box 3) \\
\hline
\textbf{valenced or affective experiences} & Experiences that feel good or bad, such as pleasure or pain \\
\hline
\end{longtable}

\newpage

\section*{Bibliography}

\hangparas{1em}{1} 
\noindent

Alamia, A., Mozafari, M., Choksi, B., \& VanRullen, R., 2023. On the
role of feedback in image recognition under noise and adversarial
attacks: A predictive coding perspective. \emph{Neural Networks}, 157,
pp.280--287.

Albantakis, L., \& Tononi, G., 2021. What we are is more than what we
do. \emph{arXiv preprint arXiv:2102.04219.}

Ali, A., Ahmad, N., de Groot, E., Johannes van Gerven, M. A., \&
Kietzmann, T. C., 2022. Predictive coding is a consequence of energy
efficiency in recurrent neural networks. \emph{Patterns}, \emph{3}(12),
100639.

Andersson, P., Strandman, A. \& Strannegård, C., 2019. Exploration
strategies for homeostatic agents. In \emph{Artificial General
Intelligence: 12th International Conference,} pp. 178-187.

Andrews, K., \& Birch, J., 2023. To understand AI sentience, first
understand it in animals. Aeon Essays. \emph{Aeon.}
\href{https://aeon.co/essays/to-understand-ai-sentience-first-understand-it-in-animals}{https://aeon.co/essays/to-understand-ai-sentience-first-understand-it-in-animals}

Aru, J., Bachmann, T., Singer, W. \& Melloni, L., 2012. Distilling the
neural correlates of consciousness. \emph{Neuroscience \& Biobehavioral
Reviews}, \emph{36}(2), pp.737-746.

Aru, J., Larkum, M., \& Shine, J. M., 2023. The feasibility of
artificial consciousness through the lens of neuroscience.
\emph{arXiv:2306.00915.}

Baars, B. J., 1988. \emph{A Cognitive Theory of Consciousness.}
Cambridge University Press.

Baars, B. J., 1997. In the theatre of consciousness: Global workspace
theory, a rigorous scientific theory of consciousness. \emph{Journal of
Consciousness Studies, 4}(4), pp.292--309.

Bahdanau, D., Cho, K., \& Bengio, Y., 2014. Neural machine translation
by jointly learning to align and translate. \emph{arXiv preprint
arXiv:1409.0473}.

Bao, C., Fountas, Z., Olugbade, T., \& Bianchi-Berthouze, N., 2020.
Multimodal data fusion based on the global workspace theory.
\emph{arXiv:2001.09485.}

Barlassina, L., \& Hayward, M. K., 2019. More of me! Less of me!:
Reflexive imperativism about affective phenomenal character. \emph{Mind,
128}(512), pp.1013--1044.

Bartlett, P. L., Montanari, A., \& Rakhlin, A., 2021. Deep learning: A
statistical viewpoint. \emph{Acta Numerica}, 30, pp.87--201.

Bayne, T., 2010. \emph{The Unity of Consciousness.} Oxford University
Press.

Bayne, T., \& Montague, M., 2011. \emph{Cognitive Phenomenology.} Oxford
University Press.

Bayne, T., Seth, A. K., \& Massimini, M., 2020. Are there islands of
awareness? \emph{Trends in Neurosciences}, \emph{43}(1), pp.6--16.

Becker-Ehmck, P., Karl, M., Peters, J., \& van der Smagt, P., 2020.
Learning to fly via deep model-based reinforcement learning.
\emph{arXiv:2003.08876.}

Belkin, M., Ma, S., \& Mandal, S., 2018. To understand deep learning we
need to understand kernel learning. In \emph{Proceedings of the 35th
International Conference on Machine Learning}, pp.541--549.

Bengio, Y., 2017. The consciousness prior. \emph{arXiv:1709.08568.}

Bengio, E., Jain, M., Korablyov, M., Precup, D., \& Bengio, Y., 2021.
Flow network based generative models for non-iterative diverse candidate
generation. \emph{arXiv:2106.04399.}

Bengio, Y., Lahlou, S., Deleu, T., Hu, E. J., Tiwari, M., \& Bengio, E.,
2023. GFlowNet foundations. \emph{arXiv:2111.09266.}

Bichot, N. P., Heard, M. T., DeGennaro, E. M., \& Desimone, R., 2015. A
source for feature-based attention in the prefrontal cortex.
\emph{Neuron,} \emph{88}(4), pp.832-844.

Birch, J., 2018. Animal cognition and human values. \emph{Philosophy of
Science}, \emph{85}(5), pp.1026--1037.

Birch, J., 2020. Global workspace theory and animal consciousness.
\emph{Philosophical Topics}, \emph{48}(1), pp.21--38.

Birch, J., Schnell, A. K., \& Clayton, N. S., 2020. Dimensions of animal
consciousness. \emph{Trends in Cognitive Sciences, 24}(10), pp.789--801.

Birch, J., Ginsburg, S., \& Jablonka, E., 2020. Unlimited associative
learning and the origins of consciousness: A primer and some
predictions. \emph{Biology \& Philosophy}, \emph{35}(6), p. 56.

Birch, J., 2022a. Materialism and the moral status of animals. \emph{The
Philosophical Quarterly}, \emph{72}(4), pp.795--815.

Birch, J., 2022b. The search for invertebrate consciousness. \emph{Noûs,
56}, pp.133--153.

Block, N., 1995. On a confusion about a function of consciousness.
\emph{Behavioral and Brain Sciences,} 18, pp.227--247.

Block, N., 1996. Mental paint and mental latex. \emph{Philosophical
Issues}, 7, pp.19--49.

Block, N., 2002. Some concepts of consciousness. \emph{Philosophy of
Mind: Classical and Contemporary Readings.} pp.206--218.

Block, N., 2007. Consciousness, accessibility, and the mesh between
psychology and neuroscience. \emph{Behavioral and Brain Sciences}, 30,
pp.481--499.

Block, N., 2023. \emph{The Border Between Seeing and Thinking}. Oxford
University Press.

Bowers, J. S., Malhotra, G., Dujmović, M., Montero, M. L., Tsvetkov, C.,
Biscione, V., Puebla, G., Adolfi, F., Hummel, J. E., Heaton, R. F.,
Evans, B. D., Mitchell, J., \& Blything, R., 2022. Deep problems with
neural network models of human vision. \emph{Behavioral and Brain
Sciences}, pp.1--74.

Breitmeyer, B. G., \& Öğmen, H., 2006. \emph{Visual Masking: Time Slices
Through Conscious and Unconscious Vision} (2nd ed.). Oxford University
Press.

Bronfman, Z. Z., Ginsburg, S., \& Jablonka, E., 2016. The transition to
minimal consciousness through the evolution of associative learning.
\emph{Frontiers in Psychology}, 7, 1954.

Brown, R., 2015. The HOROR theory of phenomenal consciousness.
\emph{Philosophical Studies}, 172, pp.1783--1794.

Brown, R., Lau, H., \& LeDoux, J. E., 2019. Understanding the
higher-order approach to consciousness. \emph{Trends in cognitive
sciences}, 23, pp.754--768.

Brown, T., Mann, B., Ryder, N., Subbiah, M., Kaplan, J. D., Dhariwal,
P., Neelakantan, A., Shyam, P., Sastry, G., \& Askell, A., 2020.
Language models are few-shot learners. \emph{Advances in neural
information processing systems,} 33, pp.1877--1901.

Bryson, J., 2010. Robots should be slaves. Wilks (Ed.), \emph{Close
Engagements with Artificial Companions}. John Benjamins.

Burgess, C. P., Matthey, L., Watters, N., Kabra, R., Higgins, I.,
Botvinick, M., \& Lerchner, A., 2019. MONet: Unsupervised scene
decomposition and representation. \emph{arXiv:1901.11390.}

Butlin, P., 2022. Machine learning, functions and goals. \emph{Croatian
Journal of Philosophy}, \emph{22}(66), pp.351--370.

Butlin, P., 2023. Reinforcement learning and artificial agency.
\emph{Mind \& Language.}

Cabanac, M., 1992. Pleasure: The common currency. \emph{Journal of
Theoretical Biology, 155}(2), pp.173--200.

Carruthers, P., 2018. Valence and value. \emph{Philosophy and
Phenomenological Research}, 97, pp.658--680.

Carruthers, P., 2019. \emph{Human and Animal Minds: The Consciousness
Questions Laid to Rest.} Oxford University Press.

Carruthers, P., \& Gennaro, R., 2020. Higher-order theories of
consciousness. E. N. Zalta (Ed.), \emph{The Stanford Encyclopedia of
Philosophy} (Fall 2020 Edition).

Chalmers, D., 1995. Absent qualia, fading qualia, dancing qualia.
\emph{Conscious Experience.}

Chalmers, D., 1996. \emph{The Conscious Mind: In Search of a Fundamental
Theory.} Oxford University Press.

Chalmers, D., 2000. What is a neural correlate of consciousness?
Metzinger (Ed.), \emph{Neural Correlates of Consciousness: Conceptual
and Empirical Questions.} The MIT Press.

Chalmers, D., 2002. Consciousness and its place in nature. Chalmers
(Ed.), \emph{Philosophy of Mind: Classical and Contemporary Readings.}
Oxford University Press.

Chalmers, D., 2004. How can we construct a science of consciousness?
Gazzaniga (Ed.), \emph{The Cognitive Neurosciences III.} The MIT Press.

Chalmers, D., 2013. Panpsychism and panprotopsychism. \emph{Amherst
Lecture in Philosophy.}

Chalmers, D., 2018. The meta-problem of consciousness. \emph{Journal of
Consciousness Studies}, 25, pp.6--61.

Chalmers, D., 2023. Could a large language model be conscious?
\emph{Boston Review}.

Chowdhery, A., Narang, S., Devlin, J., Bosma, M., Mishra, G., Roberts,
A., Barham, P., Chung, H. W., Sutton, C., Gehrmann, S., \& Schuh, P.,
2022. PaLM: Scaling language modeling with pathways.
\emph{arXiv:2204.02311}

Clark, A., 2000. \emph{A Theory of Sentience}. Oxford University Press.

Clark, A., 2008. \emph{Supersizing the Mind: Embodiment, Action, and
Cognitive Extension}. Oxford University Press.

Clark, A., 2019. Consciousness as generative entanglement. \emph{Journal
of Philosophy}, 116, pp.645--662.

Cleeremans, A., Achoui, D., Beauny, A., Keuninckx, L., Martin, J. R.,
Muñoz-Moldes, S., Vuillaume, L., \& Heering, A., 2020. Learning to be
conscious. \emph{Trends in Cognitive Sciences} 24, pp.112--123.

Conwell, C., \& Ullman, T., 2022. Testing relational understanding in
text-guided image generation. \emph{arXiv:2208.00005.}

Crick, F., \& Koch, C., 1990. Towards a neurobiological theory of
consciousness. \emph{Seminars in the Neurosciences. Saunders Scientific
Publications}, pp.263--275.

Dainton, B., 2000. \emph{Stream of Consciousness: Unity and Continuity
in Conscious Experience.} Routledge.

Dainton, B., 2023. Temporal consciousness. E. N. Zalta \& U. Nodelman
(Eds.) \emph{The Stanford Encyclopedia of Philosophy} (Spring 2023
Edition).

Deane, G., 2021. Consciousness in active inference: Deep self-models,
other minds, and the challenge of psychedelic-induced ego-dissolution.
\emph{Neuroscience of Consciousness}, 2021(2), niab024.

Dehaene, S., \& Changeux, J. P., 2011. Experimental and theoretical
approaches to conscious processing. \emph{Neuron,} 70, pp.200--227.

Dehaene, S., Kerszberg, M., \& Changeux, J. P., 1998. A neuronal model
of a global workspace in effortful cognitive tasks. In \emph{Proceedings
of the National Academy of Sciences, 95}(24). 14529--14534.

Dehaene, S., Lau, H., \& Kouider, S., 2017. What is consciousness, and
could machines have it? \emph{Science}, 358, pp.486--492.

Dehaene, S., \& Naccache, L., 2001. Towards a cognitive neuroscience of
consciousness: Basic evidence and a workspace framework.
\emph{Cognition}, 79, pp.1--37.

Dehaene, S., Sergent, C., \& Changeux, J. P., 2003. A neuronal network
model linking subjective reports and objective physiological data during
conscious perception. \emph{Proceedings of the National Academy of
Sciences}, 100, 8520--8525.

Dehaene, S., 2014. \emph{Consciousness And the Brain: Deciphering How
the Brain Codes Our Thoughts}. Viking.

Dennett, D. C., 1987. \emph{The Intentional Stance.} The MIT Press.

Dijkstra, N., Bosch, S. E., \& Gerven, M. A. J. van., 2019. Shared
neural mechanisms of visual perception and imagery. \emph{Trends in
Cognitive Sciences, 23}(5), pp.423--434.

Dijkstra, N., van Gaal, S., Geerligs, L., Bosch, S. E., \& van Gerven,
M. A. J., 2021. No evidence for neural overlap between unconsciously
processed and imagined stimuli. \emph{ENeuro, 8}(5),
ENEURO.0228-21.2021.

Dijkstra, N., Kok, P., \& Fleming, S. M., 2022. Imagery adds
stimulus-specific sensory evidence to perceptual detection.
\emph{Journal of Vision, 22}(2), 11.

Dijkstra, N., \& Fleming, S. M., 2023. Subjective signal strength
distinguishes reality from imagination. \emph{Nature Communications,
14}(1), Article 1.

Doerig, A., Schurger, A., Hess, K., \& Herzog, M. H., 2019. The
unfolding argument: Why IIT and other causal structure theories cannot
explain consciousness. \emph{Consciousness and Cognition}, \emph{72},
pp.49--59.

Dolan, R. J., \& Dayan, P., 2013. Goals and habits in the brain.
\emph{Neuron, 80}(2), pp.312--325.

Dretske, F., 1988. \emph{Explaining Behavior: Reasons in a World of
Causes.} The MIT Press.

Dretske, F., 1999. Machines, plants and animals: The origins of agency.
\emph{Erkenntnis}, \emph{51}(1), pp.19--31.

Driess, D., Xia, F., Sajjadi, M. S. M., Lynch, C., Chowdhery, A.,
Ichter, B., Wahid, A., Tompson, J., Vuong, Q., Yu, T., Huang, W.,
Chebotar, Y., Sermanet, P., Duckworth, D., Levine, S., Vanhoucke, V.,
Hausman, K., Toussaint, M., Greff, K., \& Florence, P., 2023. PaLM-E: An
embodied multimodal language model. \emph{arXiv:2303.03378.}

Dulberg, Z., Dubey, R., Berwian, I. M., \& Cohen, J. D., 2023. Having
multiple selves helps learning agents explore and adapt in complex
changing worlds. \emph{Proceedings of the National Academy of Sciences,
120}(28), e2221180120.

Elamrani, A., \& Yampolskiy, R., 2019. Reviewing tests for machine
consciousness. \emph{Journal of Consciousness Studies, 26},
pp.35-64\emph{.}

Elhage, N., Nanda, N., Olsson, C., Henighan, T., Joseph, N., Mann, B.,
Askell, A., Bai, Y., Chen, A., Conerly, T., DasSarma, N., Drain, D.,
Ganguli, D., Hatfield-Dodds, Z., Hernandez, D., Jones, A., Kernion, J.,
Lovitt, L., Ndousse, K., Amodei, D., Brown, T., Clark, J., Kaplan, J.,
McCandlish, S., \& Olah, C. 2021. A mathematical framework for
transformer circuits. \emph{Transformer Circuits Thread.}
https://transformer-circuits.pub/

Epley, N., Waytz, A., \& Cacioppo, J. T., 2007. On seeing human: A
three-factor theory of anthropomorphism. \emph{Psychological Review,
114}(4), pp.864--886.

Evans, G., 1982. \emph{The Varieties of Reference.} Oxford University
Press.

Finn, C., Goodfellow, I., \& Levine, S., 2016. Unsupervised learning for
physical interaction through video prediction. \emph{Advances in Neural
Information Processing Systems}, \emph{29}.

Fleming, S. M., 2020. Awareness as inference in a higher-order state
space. \emph{Neuroscience of consciousness,} 2020, 020.

Fodor, J. A., 1983. \emph{The Modularity of Mind: An Essay on Faculty
Psychology.} The MIT Press.

Francken, J. C., Beerendonk, L., Molenaar, D., Fahrenfort, J. J.,
Kiverstein, J. D., Seth, A. K., \& van Gaal, S., 2022. An academic
survey on theoretical foundations, common assumptions and the current
state of consciousness science. \emph{Neuroscience of Consciousness,}
2022, niac011.

Frankish, K., 2016. Illusionism as a theory of consciousness.
\emph{Journal of Consciousness Studies}, 23, pp.11--39.

Franklin, S., \& Graesser, A., 1999. A software agent model of
consciousness. \emph{Consciousness and Cognition, 8}(3), pp.285--301.

Friedrich, J., Golkar, S., Farashahi, S., Genkin, A., Sengupta, A., \&
Chklovskii, D., 2021. Neural optimal feedback control with local
learning rules. \emph{Advances in Neural Information Processing
Systems}, 34, 16358--16370.

Friston, K., 2010. The free-energy principle: a unified brain theory?
\emph{Nature Reviews Neuroscience}, 11, pp.127--138.

Gershman, S. J., 2019. The generative adversarial brain. \emph{Frontiers
in Artificial Intelligence}, 2, 18.

Ginsburg, S., \& Jablonka, E., 2019. \emph{The Evolution of the
Sensitive Soul: Learning and the Origins of Consciousness}. The MIT
Press.

Godfrey-Smith, P., 2016. Mind, matter, and metabolism. \emph{The Journal
of Philosophy}, 113, pp.481--506.

Godfrey-Smith, P., 2019. Evolving across the explanatory gap.
\emph{Philosophy, Theory, and Practice in Biology, 11}(1).

Goff, P., 2017. \emph{Consciousness and Fundamental Reality.} Oxford
University Press.

Golan, T., Raju, P. C., \& Kriegeskorte, N., 2020. Controversial stimuli: Pitting neural networks against each other as models of human cognition. In \textit{Proceedings of the National Academy of Sciences, 117}(47), 29330–29337.

Goodfellow, I. J., Pouget-Abadie, J., Mirza, M., Xu, B., Warde-Farley,
D., Ozair, S., Courville, A., \& Bengio, Y., 2014. Generative
adversarial networks. \emph{arXiv:1406.2661.}

Goyal, A., Lamb, A., Gampa, P., Beaudoin, P., Levine, S., Blundell, C.,
Bengio, Y., \& Mozer, M. 2020. Object files and schemata: Factorizing
declarative and procedural knowledge in dynamical systems.
\emph{arXiv:2006.16225.}

Goyal, A., \& Bengio, Y., 2022. Inductive biases for deep learning of
higher-level cognition. \emph{Proceedings of the Royal Society A:
Mathematical, Physical and Engineering Sciences, 478}(2266), 20210068.

Goyal, A., Didolkar, A., Lamb, A., Badola, K., Ke, N. R., Rahaman, N.,
Binas, J., Blundell, C., Mozer, M., \& Bengio, Y., 2022. Coordination
among neural modules through a shared global workspace. \emph{arXiv
preprint arXiv:2103.01197.}

Gray, K., \& Wegner, D. M., 2012. Feeling robots and human zombies: Mind
perception and the uncanny valley. \emph{Cognition, 125}(1),
pp.125--130.

Graziano, M. S., 2013. \emph{Consciousness and the Social Brain.} Oxford
University Press.

Graziano, M. S., 2017. The attention schema theory: a foundation for
engineering artificial consciousness. \emph{Frontiers in Robotics and
AI, 4}(60).

Graziano, M. S., 2019a. \emph{Rethinking Consciousness: A Scientific
Theory of Subjective Experience.} WW Norton \& Company.

Graziano, M. S., 2019b. We are machines that claim to be conscious.
\emph{Journal of Consciousness Studies}, 26, pp.95--104.

Greff, K., van Steenkiste, S., \& Schmidhuber, J., 2020. On the binding
problem in artificial neural networks. \emph{arXiv:2012.05208.}

Gunkel, D. J., 2012. \emph{The Machine Question: Critical Perspectives
on AI, Robots, and Ethics}. The MIT Press.

Guthrie, S., 1993. \emph{Faces in the Clouds: A New Theory of Religion.}
Oxford University Press.

Harnad, S., 2003. Can a machine be conscious? How? \emph{Journal of
Consciousness Studies}, \emph{10}(4--4), pp.69--75.

Hassabis, D., Kumaran, D., Summerfield, C., \& Botvinick, M., 2017.
Neuroscience-inspired artificial intelligence. \emph{Neuron, 95}(2),
pp.245--258.

Herzog, M. H., Drissi-Daoudi, L., \& Doerig, A., 2020. All in good time:
Long-lasting postdictive effects reveal discrete perception.
\emph{Trends in Cognitive Sciences, 24}(10), pp. 826--837.

Heyes, C., \& Dickinson, A., 1990. The intentionality of animal action.
\emph{Mind \& Language}, 5, pp.87--103.

Hohwy, J., 2022. Conscious self-evidencing. \emph{Review of Philosophy
and Psychology, 13}(4), pp.809--828.

Hu, E. J., Malkin, N., Jain, M., Everett, K., Graikos, A., \& Bengio,
Y., 2023. GFlowNet-EM for learning compositional latent variable models.
\emph{arXiv:2302.06576.}

Hurley, S. L., 1998. \emph{Consciousness in Action.} Harvard University
Press.

Irvine, E., 2013. Measures of consciousness. \emph{Philosophy Compass,
8}(3), pp.285--297.

Jaegle, A., Gimeno, F., Brock, A., Vinyals, O., Zisserman, A., \&
Carreira, J., 2021a. Perceiver: General perception with iterative
attention. \emph{International Conference on Machine Learning.} PMLR,
pp.4651--4664.

Jaegle, A., Borgeaud, S., Alayrac, J.-B., Doersch, C., Ionescu, C.,
Ding, D., Koppula, S., Zoran, D., Brock, A., Shelhamer, E., 2021b.
Perceiver IO: A general architecture for structured inputs \& outputs.
\emph{arXiv:2107.14795.}

Ji, X., Elmoznino, E., Deane, G., Constant, A., Dumas, G., Lajoie, G.,
Simon, J., \& Bengio, Y., 2023. Sources of richness and ineffability for
phenomenally conscious states. \emph{arXiv:2302.06403.}

Johnson, L. S. M., 2022. \emph{The Ethics of Uncertainty: Entangled
Ethical and Epistemic Risks in Disorders of Consciousness}. Oxford
University Press.

Juechems, K., \& Summerfield, C., 2019. Where does value come from?
\emph{Trends in Cognitive Sciences, 23}(10), pp.836--850.

Juliani, A., Kanai, R., \& Sasai, S. S., 2022. The Perceiver
architecture is a functional global workspace. In \emph{Proceedings of
the Annual Meeting of the Cognitive Science Society, 44}(44)\emph{.}

Kahn, P. H., Freier, N. G., Kanda, T., Ishiguro, H., Ruckert, J. H.,
Severson, R. L., \& Kane, S. K., 2008. Design patterns for sociality in
human-robot interaction. In \emph{Proceedings of the 3rd ACM/IEEE
International Conference on Human Robot Interaction}, pp.97--104.

Kahneman, D., 2011. \emph{Thinking, Fast and Slow.} Farrar, Straus and
Giroux.

Keramati, M., \& Gutkin, B., 2014. Homeostatic reinforcement learning
for integrating reward collection and physiological stability.
\emph{ELife}, 3, e04811.

Kietzmann, T. C., Spoerer, C. J., Sörensen, L. K. A., Cichy, R. M.,
Hauk, O., \& Kriegeskorte, N., 2019. Recurrence is required to capture
the representational dynamics of the human visual system.
\emph{Proceedings of the National Academy of Sciences}, \emph{116}(43),
21854--21863.

Kiverstein, J., 2007. Could a robot have a subjective point of view?
\emph{Journal of Consciousness Studies}, \emph{14}, pp.127--139.

Clark, A., \& Kiverstein, J., 2007. Experience and agency: Slipping the
mesh. \emph{Behavioral and Brain Sciences}, \emph{30}(5--6),
pp.502--503.

Krach, S., Hegel, F., Wrede, B., Sagerer, G., Binkofski, F., \& Kircher,
T., 2008. Can machines think? Interaction and perspective taking with
robots investigated via fMRI. \emph{PLOS ONE, 3}(7), e2597.

Kriegeskorte, N., 2015. Deep neural networks: A new framework for
modeling biological vision and brain information processing.
\emph{Annual Review of Vision Science, 1}(1) pp.417--46.

Krizhevsky, A., Sutskever, I., \& Hinton, G. E., 2012. ImageNet
classification with deep convolutional neural networks. \emph{Advances
in Neural Information Processing Systems}, 25.

Lamme, V. A. F., 2006. Towards a true neural stance on consciousness.
\emph{Trends in Cognitive Sciences, 10}(11), pp.494--501.

Lamme, V. A. F., 2010. How neuroscience will change our view on
consciousness. \emph{Cognitive Neuroscience, 1}(3), pp.204--220.

Lamme, V. A. F., 2020. Visual functions generate conscious seeing.
\emph{Frontiers in Psychology}, 11, 83.

Lau, H., 2019. Consciousness, metacognition, \& perceptual reality
monitoring. \emph{PsyArXiv}.

Lau, H., 2022. \emph{In Consciousness we Trust: The Cognitive
Neuroscience of Subjective Experience.} Oxford University Press.

Lau, H. C., \& Passingham, R. E., 2006. Relative blindsight in normal
observers and the neural correlate of visual consciousness.
\emph{Proceedings of the National Academy of Sciences, 103}(49),
18763--18768.

Lau, H., \& Rosenthal, D., 2011. Empirical support for higher-order
theories of conscious awareness. \emph{Trends in Cognitive Sciences,
15}(8), pp.365--373.

LeCun, Y., Bengio, Y., \& Hinton, G., 2015. Deep learning.
\emph{Nature}, 521, pp.436--444.

Lee, A. Y., 2022. Degrees of consciousness. \emph{Noûs}.

Lemoine, B., 2022, June 14. Scientific data and religious opinions.
\emph{Medium}.
\url{https://cajundiscordian.medium.com/scientific-data-and-religious-opinions-ff9b0938fc10}

Lindsay, G. W., 2020. Attention in psychology, neuroscience, and machine
learning. \emph{Frontiers in computational neuroscience}, \emph{14}, 29.

Lindsay, G. W., 2021. Convolutional neural networks as a model of the
visual system: Past, present, and future. \emph{Journal of Cognitive
Neuroscience, 33}(10), pp.2017--2031.

Liu, D., Bolotta, S., Zhu, H., Bengio, Y., \& Dumas, G., 2023. Attention
schema in neural agents. \emph{arXiv:2305.17375.}

Long, R., forthcoming. Introspective capabilities in large language models. \emph{Journal of Consciousness Studies}.

Lotter, W., Kreiman, G., \& Cox, D., 2017. Deep predictive coding
networks for video prediction and unsupervised learning.
\emph{arXiv:1605.08104.}

Lotter, W., Kreiman, G., \& Cox, D., 2020. A neural network trained for
prediction mimics diverse features of biological neurons and perception.
\emph{Nature Machine Intelligence, 2}(4), Article 4.

Lycan, W. G., 2001. A simple argument for a higher-order representation
theory of consciousness. \emph{Analysis,} 61(3--4).

Lynch, C., Wahid, A., Tompson, J., Ding, T., Betker, J., Baruch, R.,
Armstrong, T., \& Florence, P., 2022. Interactive language: Talking to
robots in real time. \emph{arXiv:2210.06407.}

Malach, R., 2021. Local neuronal relational structures underlying the
contents of human conscious experience. \emph{Neuroscience of
Consciousness,} 2021, niab028.

Malach, R., 2022. The role of the prefrontal cortex in conscious
perception: The localist perspective. \emph{Journal of Consciousness
Studies, 29}(7--8), pp.93--114.

Man, K., \& Damasio, A., 2019. Homeostasis and soft robotics in the design of feeling machines. \textit{Nature Machine Intelligence, 1}(10) pp.446–452. 

Marr, D., 1982. \emph{Vision: A Computational Investigation Into the
Human Representation and Processing of Visual Information.} W.H.
Freeman.

Mashour, G. A., Roelfsema, P., Changeux, J. P., \& Dehaene, S., 2020.
Conscious processing and the global neuronal workspace hypothesis.
\emph{Neuron,} 105, pp.776--798.

Maturana, H. R., \& Varela, F. J., 1991. \emph{Autopoiesis and
cognition: The realization of the living.} Springer.

Mazor, M., Risoli, A., Eberhardt, A., \& Fleming, S. M., 2021.
Dimensions of moral status. In \emph{Proceedings of the Annual Meeting
of the Cognitive Science Society}, \emph{43}(43).

McNamee, D., \& Wolpert, D. M., 2019. Internal models in biological
control. \emph{Annual Review of Control, Robotics, and Autonomous
Systems, 2}(1), pp.339--364.

Mediano, P. A. M., Rosas, F. E., Bor, D., Seth, A. K., \& Barrett, A.
B., 2022. The strength of weak integrated information theory.
\emph{Trends in Cognitive Sciences, 26}(8), pp.646--655.

Mehrer, J., Spoerer, C. J., Jones, E. C., Kriegeskorte, N., \&
Kietzmann, T. C., 2021. An ecologically motivated image dataset for deep
learning yields better models of human vision. \emph{Proceedings of the
National Academy of Sciences}, \emph{118}(8), e2011417118.

Melloni, L., Mudrik, L., Pitts, M., \& Koch, C., 2021. Making the hard
problem of consciousness easier. \emph{Science,} 372, 911--912.

Merel, J., Aldarondo, D., Marshall, J., Tassa, Y., Wayne, G., \&
Ölveczky, B., 2019. Deep neuroethology of a virtual rodent.
\emph{arXiv:1911.09451.}

Merker, B., 2005. The liabilities of mobility: A selection pressure for
the transition to consciousness in animal evolution. \emph{Consciousness
and Cognition,} 14, pp.89--114.

Merker, B., 2007. Consciousness without a cerebral cortex: A challenge
for neuroscience and medicine. \emph{Behavioral and Brain Sciences,
30}(1), pp.63--81.

Metzinger, T., 2021. Artificial suffering: An argument for a global
moratorium on synthetic phenomenology. \emph{Journal of Artificial
Intelligence and Consciousness,} 8, pp.43--66.

Miall, R. C., \& Wolpert, D. M., 1996. Forward models for physiological
motor control. \emph{Neural Networks, 9}(8), 1265--1279.

Michel, M., 2022. Conscious perception and the prefrontal cortex a
review. \emph{Journal of Consciousness Studies}, 29, pp.115--157.

Michel, M., forthcoming. The perceptual reality monitoring theory. M.
Herzog, A. Schurger, \& A. Doerig (Eds.), \emph{Scientific Theories of
Consciousness: The Grand Tour}. Cambridge University Press.

Michel, M., \& Doerig, A., 2022. A new empirical challenge for local
theories of Consciousness. \emph{Mind \& Language, 37}(5), pp.840--855.

Michel, M., \& Lau, H., 2021. Higher-order theories do just fine.
\emph{Cognitive Neuroscience,} 12, 77--78.

Michel, M., \& Morales, J., 2020. Minority reports: Consciousness and
the prefrontal cortex. \emph{Mind \& Language, 35}(4), pp.493--513.

Millidge, B., Salvatori, T., Song, Y., Bogacz, R., \& Lukasiewicz, T.,
2022. Predictive coding: towards a future of deep learning beyond
backpropagation? \emph{arXiv:2202.09467.}

Mnih, V., Heess, N., \& Graves, A., 2014. Recurrent models of visual
attention. \emph{Advances in neural information processing systems},
\emph{27}.

Morales, J., Odegaard, B., \& Maniscalco, B., 2022. The neural
substrates of conscious perception without performance confounds. F. De
Brigard \& W. Sinnott-Armstrong (Eds.), \emph{Neuroscience and
Philosophy.} The MIT Press.

Nagel, T., 1974. What is it like to be a bat? \emph{The Philosophical
Review,} 83, pp.435--450.

Nave, K., Deane, G., Miller, M., \& Clark, A., 2022. Expecting some
action: Predictive processing and the construction of conscious
experience. \emph{Review of Philosophy and Psychology, 13}(4),
1019--1037.

Noe, A., 2004. \emph{Action in Perception.} The MIT Press.

Noudoost, B., Chang, M. H., Steinmetz, N. A., Moore, T., 2010. Top-down
control of visual attention. \emph{Current Opinion in Neurobiology},
\emph{20}(2), pp.183-190.

Oizumi, M., Albantakis, L., \& Tononi, G., 2014. From the phenomenology
to the mechanisms of consciousness: Integrated information theory 3.0.
\emph{PLOS Computational Biology} 10, e1003588.

Olah, C., Satyanarayan, A., Johnson, I., Carter, S., Schubert, L., Ye,
K., \& Mordvintsev, A., 2018. The building blocks of interpretability.
\emph{Distill,} 3.

Oord, A. van den, Li, Y., \& Vinyals, O., 2019. Representation learning
with contrastive predictive coding. \emph{arXiv:1807.03748.}

Oprea, S., Martinez-Gonzalez, P., Garcia-Garcia, A., Castro-Vargas, J.
A., Orts-Escolano, S., Garcia-Rodriguez, J., \& Argyros, A., 2022. A
review on deep learning techniques for video prediction. \emph{IEEE
Transactions on Pattern Analysis and Machine Intelligence, 44(}6),
2806--2826.

O'Regan, J. K., \& Noë, A., 2001. A sensorimotor account of vision and
visual consciousness. \emph{Behavioral and Brain Sciences,} 24,
pp.939--973.

Panagiotaropoulos, T. I., Deco, G., Kapoor, V., \& Logothetis, N. K.,
2012. Neuronal discharges and gamma oscillations explicitly reflect
visual consciousness in the lateral prefrontal cortex. \emph{Neuron,
74}(5), pp.924--935.

Pang, Z., O'May, C. B., Choksi, B., \& VanRullen, R., 2021. Predictive
coding feedback results in perceived illusory contours in a recurrent
neural network. \emph{Neural Networks,} 144, pp.164--175.

Papineau, D., 2002. \emph{Thinking About Consciousness.} Oxford
University Press.

Peters, M. A. K., \& Lau, H., 2015. Human observers have optimal
introspective access to perceptual processes even for visually masked
stimuli. \emph{ELife,} 4, e09651.

Peterson, A., Cruse, D., Naci, L., Weijer, C., \& Owen, A. M., 2015.
Risk, diagnostic error, and the clinical science of consciousness.
\emph{NeuroImage: Clinical,} 7, pp.588--597.

Phillips, I., 2018a. The methodological puzzle of phenomenal
consciousness. \emph{Philosophical Transactions of the Royal Society B:
Biological Sciences} 373, 20170347.

Phillips, I., 2018b. Consciousness, time, and memory. \emph{The
Routledge Handbook of Consciousness} (pp.286--297). Routledge.

Prosser, S., 2016. \emph{Experiencing Time.} Oxford University Press.

Quilty-Dunn, J., Porot, N., \& Mandelbaum, E., 2022. The best game in
town: The re-emergence of the language of thought hypothesis across the
cognitive sciences. \emph{Behavioral and Brain Sciences}, pp.1--55.

Radford, A., Wu, J., Child, R., Luan, D., Amodei, D., \& Sutskever, I.,
2019. Language models are unsupervised multitask learners. \emph{OpenAI
blog 1}, 9

Reynolds, J. H., \& Heeger, D.J., 2009. The normalization model of
attention. \emph{Neuron}, \emph{61}(2), pp.168-185.

Rosenthal, D., 2005. \emph{Consciousness and Mind.} Clarendon Press.

Rosenthal, D., 2010. How to think about mental qualities.
\emph{Philosophical Issues}, \emph{20}(1), pp.368--393.

Russell, S., \& Norvig, P., 2010. \emph{Artificial Intelligence: A
Modern Approach} (3rd Edition). Prentice-Hall.

Sajjadi, M. S. M., Duckworth, D., Mahendran, A., van Steenkiste, S.,
Pavetić, F., Lučić, M., Guibas, L. J., Greff, K., \& Kipf, T., 2022.
Object scene representation transformer. \emph{arXiv:2206.06922.}

Salti, M., Monto, S., Charles, L., King, J.-R., Parkkonen, L., \&
Dehaene, S., 2015. Distinct cortical codes and temporal dynamics for
conscious and unconscious percepts. \emph{ELife}, 4, e05652.

Savage, J. E., 1972. Computational work and time on finite machines.
\emph{Journal of the ACM}, \emph{19}(4), pp.660--674.

Schneider, S., 2019. \emph{Artificial You: AI and the Future of Your
Mind.} Princeton University Press.

Schwitzgebel, E., 2016. Phenomenal consciousness, defined and defended
as innocently as I can manage. \emph{Journal of Consciousness Studies,}
23, pp.224--235.

Schwitzgebel, E., 2021. \emph{Belief.} In E. N. Zalta (Ed.), \emph{The
Stanford Encyclopedia of Philosophy} (Winter 2021). Metaphysics Research
Lab, Stanford University.

Schwitzgebel, E. (forthcoming). Borderline consciousness, when it's
neither determinately true nor determinately false that experience is
present. \emph{Philosophical Studies}.

Schwitzgebel, E., \& Garza, M., 2015. A defense of the rights of
artificial intelligences. \emph{Midwest Studies in Philosophy, 39}(1),
pp.98--119.

Schwitzgebel, E., \& Garza, M., 2020. Designing AI with rights,
consciousness, self-respect, and freedom. Liao (Ed.), \emph{Ethics of
Artificial Intelligence}. Oxford University Press.

Searle, J. R., 1980. Minds, brains, and programs. \emph{Behavioral and
Brain Sciences, 3}(3) pp.417-457

Seth, A., 2021. \emph{Being You: A New Science of Consciousness.}
Penguin.

Seth, A. K., \& Bayne, T., 2022. Theories of consciousness. \emph{Nature
Reviews Neuroscience,} pp.1--14.

Seth, A. K., \& Hohwy, J., 2021. Predictive processing as an empirical
theory for consciousness science. \emph{Cognitive Neuroscience,} 12,
89--90.

Shanahan, M., 2006. A cognitive architecture that combines internal
simulation with a global workspace. \emph{Consciousness and Cognition}
15, pp.433--449.

Shanahan, M., 2010. \emph{Embodiment and the Inner Life: Cognition and
Consciousness in the Space of Possible Minds.} Oxford University Press.

Shanahan, M., McDonell, K., \& Reynolds, L., 2023. Role-play with large
language models. \emph{arXiv:2305.16367.}

Shevlin, H., 2021. Non-human consciousness and the specificity problem:
A modest theoretical proposal. \emph{Mind \& Language, 36}(2),
pp.297--314.

Shulman, C., \& Bostrom, N., 2021. Sharing the world with digital minds.
S. Clarke, H. Zohny, \& J. Savulescu (Eds.), \emph{Rethinking Moral
Status.} Oxford University Press.

Silver, D., Huang, A., Maddison, C. J., Guez, A., Sifre, L., van den
Driessche, G., Schrittwieser, J., Antonoglou, I., Panneershelvam, V.,
Lanctot, M., Dieleman, S., Grewe, D., Nham, J., Kalchbrenner, N.,
Sutskever, I., Lillicrap, T., Leach, M., Kavukcuoglu, K., Graepel, T.,
\& Hassabis, D., 2016. Mastering the game of Go with deep neural
networks and tree search. \emph{Nature,} 529(7587), Article 7587.

Simon, J. A., 2017. Vagueness and zombies: Why `phenomenally conscious'
has no borderline cases. \emph{Philosophical Studies, 174}(8),
2105--2123.

Sprevak, M. D., 2007. Chinese rooms and program portability. \emph{The
British Journal for the Philosophy of Science 58,} pp.755-776\emph{.}

Stich, S. P., 1978. Beliefs and subdoxastic states. \emph{Philosophy of
Science,} 45, pp.499--518.

Strawson, G., 1994. \emph{Mental Reality.} The MIT Press.

Strawson, G., 2006. Realistic monism: Why physicalism entails
panpsychism. \emph{Journal of Consciousness Studies,} 13, pp.3-31.

Summerfield, C., 2023. \emph{Natural General Intelligence: How
Understanding the Brain Can Help Us Build AI}. Oxford University Press.

Sutton, R., \& Barto, A., 2018. \emph{Reinforcement Learning} (2nd
Edition). The MIT Press.

Sytsma, J., 2014. Attributions of consciousness. \emph{Wiley
Interdisciplinary Reviews. Cognitive Science}, \emph{5}(6), pp.635--648.

Thompson, E., 2005. Sensorimotor subjectivity and the enactive approach
to experience. \emph{Phenomenology and the Cognitive Sciences, 4}(4),
pp.407--427.

Thompson, E., 2007. \emph{Mind in Life: Biology, Phenomenology, and the
Sciences of Mind.} Harvard University Press.

Thoppilan, R., De Freitas, D., Hall, J., Shazeer, N., Kulshreshtha, A.,
Cheng, H. T., Jin, A., Bos, T., Baker, L., Du, Y., Li, Y., Lee, H.,
Zheng, H. S., Ghafouri, A., Menegali, M., Huang, Y., Krikun, M.,
Lepikhin, D., Qin, J., \ldots{} Le, Q. 2022. LaMDA: Language models for
dialog applications. \emph{arXiv:2201.08239.}

Tishby, N., Pereira, F. C., \& Bialek, W., 2000. The information
bottleneck method. \emph{arXiv:physics/0004057.}

Todorov, E., \& Jordan, M. I., 2002. Optimal feedback control as a
theory of motor coordination. \emph{Nature Neuroscience,} 5(11), Article
11.

Tononi, G., \& Koch, C., 2015. Consciousness: here, there and
everywhere? \emph{Philosophical Transactions of the Royal Society B:
Biological Sciences,} 370, 20140167.

Treue, S. \& Trujillo \& J. C. M., 1999. Feature-based attention influences
motion processing gain in macaque visual cortex. \emph{Nature},
\emph{399}(6736), pp.575-579.

Tsuchiya, N., Wilke, M., Frässle, S., \& Lamme, V. A. F., 2015.
No-Report paradigms: Extracting the true neural correlates of
consciousness. \emph{Trends in Cognitive Sciences, 19}(12), pp.757--770.

Turkle, S., 2011. \emph{Alone Together: Why We Expect More from
Technology and Less from Each Other.} Basic Books, Inc.

Tye, M., 1995. \emph{Ten Problems of Consciousness: A Representational
Theory of the Phenomenal Mind.} The MIT Press.

Udell, D. B., \& Schwitzgebel, E., 2021. Susan Schneider's proposed
tests for AI consciousness: Promising but flawed. \emph{Journal of
Consciousness Studies, 28}(5--6), pp.121--144.

van Vugt, B., Dagnino, B., Vartak, D., Safaai, H., Panzeri, S., Dehaene,
S., \& Roelfsema, P. R., 2018. The threshold for conscious report:
Signal loss and response bias in visual and frontal cortex.
\emph{Science}, 360(6388), pp.537--542.

VanRullen, R., 2016. Perceptual cycles. \emph{Trends in Cognitive
Sciences, 20}(10), pp.723--735.

VanRullen, R., \& Kanai, R., 2021. Deep learning and the global
workspace theory. \emph{Trends in Neurosciences,} 44, pp.692--704.

Vaswani, A., Shazeer, N., Parmar, N., Uszkoreit, J., Jones, L., Gomez,
A. N., Kaiser, L., \& Polosukhin, I., 2023. Attention is all you need.
\emph{arXiv:1706.03762.}

Vorberg, D., Mattler, U., Heinecke, A., Schmidt, T., \& Schwarzbach, J.,
2003. Different time courses for visual perception and action priming.
In \emph{Proceedings of the National Academy of Sciences},
\emph{100}(10), 6275--6280.

Webb, T. W., \& Graziano, M. S. A., 2015. The attention schema theory: A
mechanistic account of subjective awareness. \emph{Frontiers in
Psychology}, \emph{6}.

Whyte, C. J., 2019. Integrating the global neuronal workspace into the
framework of predictive processing: Towards a working hypothesis.
\emph{Consciousness and Cognition,} 73, 102763.

Wilson, B. A., Baddeley, A. D., \& Kapur, N., 1995. Dense amnesia in a
professional musician following herpes simplex virus encephalitis.
\emph{Journal of Clinical and Experimental Neuropsychology},
\emph{17}(5), pp.668--681.

Wilterson, A. I., \& Graziano, M. S., 2021. The attention schema theory
in a neural network agent: Controlling visuospatial attention using a
descriptive model of attention. \emph{Proceedings of the National
Academy of Sciences,} 118, e2102421118.

Wu, P., Escontrela, A., Hafner, D., Abbeel, P., \& Goldberg, K., 2023.
DayDreamer: World models for physical robot learning. In
\emph{Proceedings of The 6th Conference on Robot Learning}, 2226--2240.

Yaron, I., Melloni, L., Pitts, M., \& Mudrik, L., 2022. The ConTraSt
database for analysing and comparing empirical studies of consciousness
theories. \emph{Nature Human Behaviour, 6}(4), pp.593--604.

Zador, A., Escola, S., Richards, B., Ölveczky, B., Bengio, Y., Boahen,
K., Botvinick, M., Chklovskii, D., Churchland, A., Clopath, C., DiCarlo,
J., Ganguli, S., Hawkins, J., Koerding, K., Koulakov, A., LeCun, Y.,
Lillicrap, T., Marblestone, A., Olshausen, B., \ldots{} Tsao, D., 2023.
Toward next-generation artificial intelligence: Catalyzing the NeuroAI
revolution. \emph{arXiv:2210.08340. arXiv.}

Zeki, S., \& Bartels, A., 1998. The autonomy of the visual systems and
the modularity of conscious vision. \emph{Philosophical Transactions of
the Royal Society of London B: Biological Sciences, 353}(1377),
1911--1914.

Zhuang, C., Yan, S., Nayebi, A., Schrimpf, M., Frank, M. C., DiCarlo, J.
J., \& Yamins, D. L. K., 2021. Unsupervised neural network models of the
ventral visual stream. \emph{Proceedings of the National Academy of
Sciences}, \emph{118}(3), e2014196118.

Zlotowski, J., Sumioka, H., Nishio, S., Glas, D., Bartneck, C., \&
Ishiguro, H., 2015,. Persistence of the uncanny valley: The influence of
repeated interactions and a robot's attitude on its perception.
\emph{Frontiers in Psychology,} 6.

\end{document}